\documentclass{article} 
\usepackage{iclr2026_conference,times}

\usepackage{amsmath,amsfonts,bm}

\def\eqref#1{equation~\ref{#1}}

\def\1{\bm{1}}

\DeclareMathAlphabet{\mathsfit}{\encodingdefault}{\sfdefault}{m}{sl}
\SetMathAlphabet{\mathsfit}{bold}{\encodingdefault}{\sfdefault}{bx}{n}

\newcommand{\R}{\mathbb{R}}

\DeclareMathOperator*{\argmin}{arg\,min}

\usepackage{wrapfig} 

\usepackage[utf8]{inputenc} 
\usepackage[T1]{fontenc}    
\usepackage{hyperref}       
\usepackage{url}            
\usepackage{booktabs}       
\usepackage{amsfonts}       
\usepackage{nicefrac}       
\usepackage{microtype}      
\usepackage{xcolor}         
\usepackage{array}          
\usepackage{caption}        
\usepackage{lipsum}
\usepackage[pdftex]{graphicx}
\usepackage{float}
\usepackage{amsmath}
\usepackage{amssymb}
\usepackage{amsthm}
\usepackage[ruled, noline]{algorithm2e}
\usepackage{cleveref}
\usepackage[normalem]{ulem}
\usepackage{algorithmic}
\usepackage{tikz}

\theoremstyle{definition}

\theoremstyle{remark}

\newcommand*{\N}{\mathbb{N}} 
\newcommand*{\tran}{^{\mkern-1.5mu\mathsf{T}}} 

\newcommand{\update}[1]{{\color{blue}#1}}

\title{Rapid training of Hamiltonian Graph \\ Networks using random features}

\author{
  \hspace{-0.7em}Atamert Rahma\(^{1,2}\)\(\thanks{Corresponding author, atamert.rahma@tum.de}\)\(\,\,,\) Chinmay Datar\(^{1,2,3}\)\(,\) Ana \v Cukarska\(^{1,2}\)\(,\) Felix Dietrich\(^{1,2,4}\)\\
  \hspace*{-0.4em}\(^{1}\)Technical University of Munich \quad \quad \hspace*{+0.0em} \(^{2}\)Munich Center for Machine Learning\\
  \hspace*{-0.4em}\(^{3}\)TUM Institute for Advanced Study \quad
  \(^{4}\)Munich Data Science Institute
}

\iclrfinalcopy 
\begin{document}

\maketitle

\begin{abstract}
Learning dynamical systems that respect physical symmetries and constraints remains a fundamental challenge in data-driven modeling.
Integrating physical laws with graph neural networks facilitates principled modeling of complex N-body dynamics and yields accurate and permutation-invariant models. 
However, training graph neural networks with iterative, gradient-descent-based optimization algorithms (e.g., Adam, RMSProp, LBFGS) often leads to slow training, especially for large, complex systems. 
In comparison to 15 different optimizers, we demonstrate that Hamiltonian Graph Networks (HGN) can be trained 150-600× faster--but with comparable accuracy--by replacing iterative optimization with random feature-based parameter construction. 
We show robust performance in diverse simulations, including N-body mass-spring and molecular systems in up to $3$ dimensions and 10,000 particles with different geometries, while retaining essential physical invariances with respect to permutation, rotation, and translation. Our proposed approach is benchmarked using a NeurIPS 2022 Datasets and Benchmarks Track publication to further demonstrate its versatility. We reveal that even when trained on minimal 8-node systems, the model can generalize in a zero-shot manner to systems as large as 4096 nodes without retraining.
Our work challenges the dominance of iterative gradient-descent-based optimization algorithms for training neural network models for physical systems.

\end{abstract}

\begin{figure}[h]
    \centering
       \includegraphics[width=\linewidth]{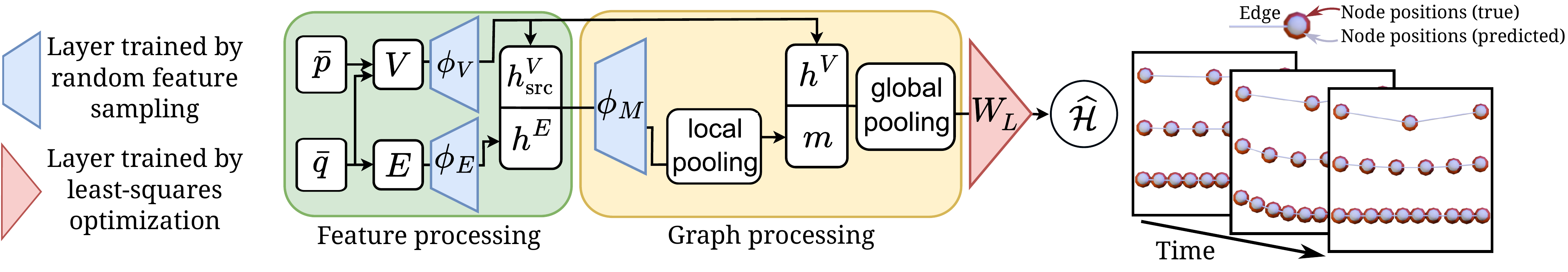}
    \caption{We propose an efficient training method for Hamiltonian graph networks using random feature sampling and linear solvers (left, also see \Cref{fig:swim-hgnn-architecture}).
    The HGN captures ground truth dynamics of physical systems (shown: chain of 10 nodes, trained on 5) and trains up to 600× faster than State-Of-The-Art (SOTA) optimizers.}
    \label{fig:main figure}
\end{figure}

\section{Introduction}\label{sections:intro}

\textbf{Learning from data} requires careful design in several key areas: the data, the model, and the training processes. 
To enable the model to generalize beyond the training set, it is important to incorporate a set of inductive biases into these processes. 
\textbf{When approximating physical systems}, it is beneficial to include physical priors to accurately capture the system's characteristics, including its dynamics and the fundamental physical laws \citep{tenenbaum-2000-physics-priors, chang-2016-physics-priors, watters-2017-physics-priors, de-2018-physics-priors, sharma-2025}. Consequently, many architectural designs are rooted in modeling physical frameworks, such as Hamiltonian mechanics \citep{bertalan-2019-hnn, greydanus-2019-hnn}, Lagrangian mechanics \citep{cranmer-2019-lagrangian-nn, lutter-2018-lagrangian-deep-nn,ober-blobaum-2023}, port-Hamiltonian systems \citep{shaan-2021-port-hnn, roth-2025-stable-port-hnn}, and GENERIC \citep{hernandez-2021-generic-spnn, lee-2021-generic-gnode, zhang-2022-generic-gfinn, gruber-2025-generic-nms}. 

\textbf{Graph networks} have useful inductive biases such as locality and permutation invariance \citep{corso-2024}, which are desirable for many interconnected, complex systems observed in nature. 
Thus, for many applications in natural sciences, a graph network model is a suitable choice. 
The key aspects of modern graph networks include neural message passing \citep{gilmer-2017} and encoding additional local information into the system \citep{corso-2024, schlichtkrull-2018-cond-edge-feature-gnn, brockschmidt-2020-cond-edge-feature-gnn}. 
In physics, graph networks have been employed to analyze data from the Large Hadron Collider \citep{dezoort-2023}, model mechanical systems \citep{zhao-2024}, and fluid dynamics \citep{xue-2022, peng-2023, li-2024}.

Efficient and robust training of graph networks on large systems for natural and life sciences is an active area of research. 
Despite the advantages of using graphs for physical N-body systems, their \textbf{training is reportedly slow due to gradient-descent-based iterative optimization} \citep{kose-2023, shukla-2022, vignac-2020, kumar-2023, zhao-2025, marino-2025}. 
These challenges become even more pronounced when a numerical integrator is incorporated into the model architecture \citep{xiong-2021-nonseparable-hard-training-with-integrator}. Furthermore, physics-informed models are often sensitive to the selection of hyperparameters \citep{shukla-2022}, which increases the challenges posed by slow iterative training.

Recently \textbf{random feature (RF) networks} have been shown to be promising for approximating physical systems \citep{fabiani-2021,datar-2024-swim-pde, rahma-2024-swim-hnn, bolager-2024-swim-rnn,fabiani-2025,galaris-2022}. However, to the best of our knowledge, random features have not been used to train graph networks for modeling physical systems.
Recent work on RF-Hamiltonian neural networks (RF-HNNs) is promising \citep{rahma-2024-swim-hnn}, where RF-HNNs are trained without using iterative algorithms. 
The authors demonstrate very low approximation errors, but only for very small systems and without leveraging the graph structure. 
In this paper, we introduce an efficient and accurate training method that utilizes random features for Hamiltonian Graph Networks (RF-HGNs, see~\Cref{fig:main figure}). 
Our main contributions are as follows.
\begin{itemize}
    \item We \textbf{introduce Random Feature Hamiltonian Graph Networks}, combining random sampling with graph-based physics-informed models for the first time, and show how one can incorporate translation, rotation, and permutation invariance as well as knowledge about the physical system (see \Cref{sections:method}).
    \item We provide \textbf{a much faster and more efficient alternative to gradient-descent-based iterative optimization algorithms} for training that avoids challenges related to slow, non-convex optimization and vanishing or exploding gradients (see \Cref{section:results}). 

    \item We perform a \textbf{comprehensive optimizer comparison} with 15 different optimizers and demonstrate that random feature-based parameter construction offers up to 600 times faster training for HGNs, without sacrificing predictive performance (see \Cref{section:results:optim-study}). The demonstrations are performed on challenging benchmark problems, including mass-spring and molecular dynamics systems.
    
    \item We \textbf{demonstrate strong zero-shot generalization}, with models trained on graphs with tens of nodes accurately predicting dynamics on graphs with thousands of nodes (see \Cref{section:results:zero-shot}).

\end{itemize}

\section{Related work}\label{sections:relwork}

\paragraph{Training graph networks:}
A graph structure allows for modeling a wide range of processes due to structural properties or underlying causal relationships. For many problems, the best way to achieve good performance is by using a large, high-quality dataset to train a large model. In such settings, training can be significantly slowed down due to the computational effort needed for backpropagation.
Improvements can be achieved with specific sampling methods for the training data \citep{nagarajan-2023, zhang-2022, zhou-2022, kaler-2022, zhang-2021, lin-2020}, graph coarsening \citep{hashemi-2024, kumar-2023, jin-2022, jin-2022a, bravo-hermsdorff-2019}, or hardware acceleration\citep{shao-2024, gupta-2024, zhu-2024, wan-2023, yang-2022, kaler-2022, wang-2021, cai-2021, lin-2020}.  Nevertheless, graph networks for physics still face challenges during training due to a need for high accuracy in the dataset, irregular memory access, load imbalance during backpropagation \citep{shukla-2022}, and hyperparameter tuning~\citep{schmidt-2021}.

\paragraph{Graph networks for physics:}
A notable advantage of graph-based models is that they are tractable for high-dimensional data, assuming that the graph connectivity remains sparse, 
which is suitable for many physical systems learned in a data-driven way.
Recent work has incorporated graph neural networks into their model architectures for approximating physical
systems \citep{pfaff-2021-learning-mesh, sanchez-gonzalez-2019-hgnn, sanchez-2020, tierz-2025-generic-gnn, varghese-2025-sympgnn, bhattoo-2022-lgn-ridig, thangamuthu-2022}. 
However, training a graph network for a very large number of nodes is challenging; one possible remedy is to partition a large graph and enable information exchange between partitions, which are trained individually \citep{nabian-2024}. {Approaches for graph networks for Hamiltonian and Lagrangian systems are used by \citet{thangamuthu-2022, bhattoo-2022-lgn-ridig, bishnoi-2023}; these models are typically trained with the Adam optimizer and applied to N-body systems.} 
Other work addresses the issue of long-range information loss in large graphs by adding physics-based connections \citep{yu-2025}, yielding better predictions but not addressing training difficulties that might arise.


\paragraph{Random features for graph networks:}
{Random features originated with the idea of using a perceptron by \citet{rosenblatt-1962} and gained traction after theoretical contributions established that they can lead to accurate approximations \citep{johnson-1984, barron-1993, rahimi-2007random-fourier-features} at low computational cost.} As the machine learning community gains a better understanding of random features~\citep{rahimi-2008-random-feature, bolager-2023-swim,fabiani-2024}, 
many new variants are being explored \citep{zozoulenko-2025, bolager-2024-swim-rnn, datar-2024-swim-pde, rahma-2024-swim-hnn}. 
An innovative approach for graph classification problems used a random features approach and demonstrated competitive accuracy on large classification datasets with a training time of only a few seconds or minutes \citep{gallicchio-2020}. Such an approach is related to echo state graph networks \citep{gallicchio-2010, wang-2023}. 
Recent work has also developed graph random features enabling kernel methods on large graphs \citep{choromanski-2023, reid-2023, reid-2023a}, leading to a notable reduction of the cubic time-complexity for kernel learning. 
\section{Method}\label{sections:method}

\textbf{Problem setup: } 
In this study, we aim to efficiently learn the Hamiltonian of a dynamical system from observed phase space trajectories, while exploiting the underlying graph structure and incorporating relevant physical invariances into the model.  
We consider a target Hamiltonian for an N-body system on \(\R^{2d \cdot N}\), the Euclidean phase-space of dimension \(2d \cdot N \in \N\), 
where $d$ is the spatial dimension. 
We denote the generalized position and momentum vectors by $q, p \in \R^{d \cdot N}$, with \(q_i, p_i \in \mathbb{R}^{d} \), denoting the $i^{\text{th}}$ particle's state.  
We denote by $\dot{x}$ the time derivatives of a trajectory $x(t):\mathbb{R}\to \R^k$ for $k \in \N$. 
The Hamiltonian is a scalar-valued function \(\mathcal{H}: \R^{2d \cdot N} \rightarrow \R\) that
describes the system dynamics in the phase-space through Hamilton's equations \citep{hamilton-1834, hamilton-1835} given by 
\begin{equation}\label{eq:hamiltons-pde}
  \begin{bmatrix}\dot{q} \\  \dot{p}\end{bmatrix} = J \nabla\mathcal{H}(q,p),
   \ \ \ J = \begin{bmatrix} 0 & I \\ -I & 0 \end{bmatrix} \in \R^{(2d \cdot N) \times (2d\cdot N)},
\end{equation}
where 
\(I \in \R^{(d\cdot N) \times (d\cdot N)}\)
is the identity matrix. 
We summarize the notation in \Cref{app_notation}. 

\textbf{Graph representation:} 
We focus on N-body systems in this work, where the graph representation is naturally available, e.g., a chain of masses connected via springs. 
Given \(d_V, d_E \in \N\), we write the system
with $N$ nodes as a graph \(G=(V,E)\) with a node feature set \(V=\{v_i \in \R^{d_{V}} \mid i = 1, \dots, N\}\) and an edge feature set \(E = \{ e_{ij} \in \R^{d_E} \mid \forall \, i, j \text{ such that } A_{ij} = 1 \}\), where \(A \in \R^{N \times N}\) is the symmetric adjacency matrix that encodes the node connectivity information. 
We parametrize the Hamiltonian function $\mathcal{H}$ with a graph neural network, and then use the trained network to simulate the physical system by integrating \Cref{eq:hamiltons-pde} with the symplectic Störmer-Verlet integrator (\citep{hairer-2003}, also see \citep{offen-2022-GP-inverse-modified-hamiltonian}). 
In contrast to previous work on Hamiltonian Neural Networks~\citep{bertalan-2019-hnn, greydanus-2019-hnn,dierkes-2023-hnn-with-symplectic-prior}, we train our networks through random feature sampling algorithms rather than iterative, gradient-descent-based optimization. 
Thus, we call our approach \textbf{``gradient-descent-free.''} 

\subsection{Encoding invariances}
The systems we consider 
are translation-, permutation-, and rotation-invariant, i.e., when the whole system is shifted, permuted, or rotated, the Hamiltonian stays constant. With the general Hamiltonian formulation, however, we cannot guarantee such invariances by design \citep{du-2022-se3}.  
To construct such invariant representations, we introduce transformed coordinates
\(\bar{q}, \bar{p} \in \R^{d \cdot N}\)
derived from the original phase-space coordinates \(q, p \in \R^{d \cdot N}\) defined in an arbitrary reference frame.  

\textbf{Translation-invariant representation:}
To make the position representation (and consequently the system representation) translation invariant, we normalize the positions by subtracting the mean $q_i \gets q_i - \frac{1}{N}\sum_{i = 1}^{N}{q_i}$.  
We do not make the generalized momenta \(p\) translation invariant, as shifting the momenta would change the total energy of the system in N-body systems, for instance, when the kinetic energy depends on the norm of \(p\).

\textbf{Permutation-invariance:} 
The graph structure and appropriate message passing algorithms inherently provide us with a system representation  that is invariant with respect to node index permutation. 

\captionsetup{skip=0pt} 
\begin{wrapfigure}{r}{0.6\textwidth}
    \centering
    \includegraphics[width=\linewidth]{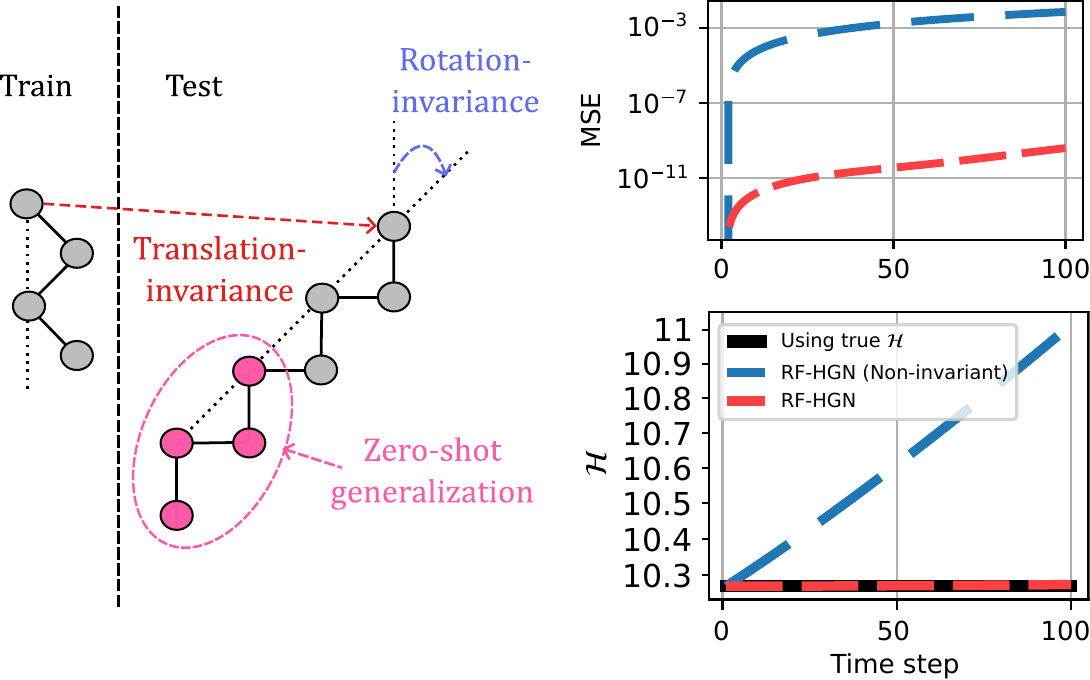}
    \vspace{2.5pt}
    \caption{Illustration of train and test N-body system positions showcasing the RF-HGN’s translation- and rotation-invariance, and its zero-shot generalization capability, validated by conserved Hamiltonian and low trajectory prediction errors for the test data (see \Cref{app_fig2_details} for details).}
    \label{fig:illustration_invariances}
    \vspace{-15pt}
\end{wrapfigure}
\textbf{Rotation-invariant representation:} 
Starting from the translation-invariant representation, we then perform another transformation to make the final representation also rotation-invariant.
Here, we explain how to encode a rotation-invariant representation for a single-body system $N = 1$ and spatial dimension $d = 2$ for brevity. 

We construct a new representation in a local orthonormal basis starting from the original coordinates $p, q \in \R^2$. 
To construct the basis, we choose the first basis vector as \(e_1 = \frac{q_1}{\lVert q_1 \rVert} \in \R^2\), where  \(\lVert \cdot \rVert\) denotes the \(l^2\) norm. 
We construct a second basis vector \(e_2 = \mathcal{R} e_1 \in \R^2\), which is orthonormal to 
\(q_1 \in \R^{2} \) and obtained by simply rotating \(q_1 \) by $90^o$ using a rotation matrix $\mathcal{R}$.
We then define an orthonormal matrix with the two basis vectors as \(\mathcal{B} = \begin{bmatrix}
    e_1 & e_2 
\end{bmatrix}\).
Finally, the rotation-invariant coordinates for any body $i \in \{1,\dots,N\}$ are 
\(\bar{q}_i = \mathcal{B}\tran q_i\). 

    Given a fixed first point, our procedure yields a rotation-invariant representation.
    One can uniquely identify the first point, independent of node ordering or orientation, as the one closest to the mean $\bar{q}$. In case of ties, we select the point with the smallest angle relative to the first coordinate axis centered at $\bar{q}$. 
The same procedure is applied to obtain rotation-invariant representations of the momenta.
In higher dimensions, when $d > 2$ and $N > 1$, we follow a similar procedure, but construct the orthogonal bases using Gram-Schmidt orthogonalization instead (see \Cref{app_rot_inv}). 
\Cref{fig:illustration_invariances} demonstrates how translating the N-body system, rotating it, and even adding new nodes without re-training (zero-shot-generalization), yields low trajectory errors while conserving the Hamiltonian.


\subsection{Model}
\begin{figure}[h]
  \begin{center}
    \includegraphics[width=.85\linewidth]{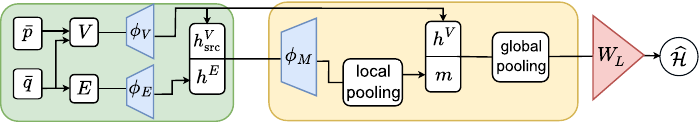}
  \end{center}
  \caption{Random-feature Hamiltonian graph neural network architecture. 
  \textbf{Left (green box):} Construction of node and edge encodings \(h_{src}^{V} \) and \(h^E\) from translation and rotation invariant position \(q\) and momenta \(p\) representations of an N-body system. \textbf{Right (orange box):} Construction of a global encoding for the graph using message passing. 
  In RF-HGN, dense layers (blue) are constructed with random features, and linear layer weights (red) are optimized by solving a linear problem. 
  }\label{fig:swim-hgnn-architecture}
\end{figure}

We now describe each component of the architecture of the Hamiltonian Graph Network (HGN) in detail (see \Cref{fig:swim-hgnn-architecture}). Please refer to \Cref{app_alg} for details on the forward pass.

\subsubsection{Node and edge encoding}\label{section:method:model:node-edge-encoding}
\textbf{Node features:} 
For an N-body system with translation- and rotation-invariant position and momenta representations, we define node features as \( v_i = \begin{bmatrix}
      \bar{q}_i\tran & \bar{p}_i\tran
  \end{bmatrix}\tran \in \R^{d_V}\), where $d_V = 2d$, for each \(i \in \{1,\dots,N\}\). 
  We define the set $V := \{ v_i \,|\, i = 1,\dots,N \}$ that collects all node encodings.

\textbf{Edge features:}
We define the edge features for each edge \((i,j)\) with \(i > j\) as 
\( 
    e_{ij} = 
    \begin{bmatrix}
        (\bar{q}_i - \bar{q}_j)\tran ; \lVert \bar{q}_i - \bar{q}_j \rVert 
    \end{bmatrix}\tran \in \R^{d_E}, \) where \( d_E = d + 1
\)
and take the absolute value \(|(\bar{q}_i - \bar{q}_j)\tran|\) in the molecular system examples. We collect all the edge feature encodings in the set $E := \{ e_{ij} \,|\, i > j \;\text{and}\; i, j \in \{1,\dots,N\} \}$. In order to reduce the memory and computation costs, we define a canonical direction by always computing edge features from higher to lower-indexed nodes, such that each edge is represented only once and set (\(e_{ji} = e_{ij}\)). 
We use the relative displacement vector \(\bar{q}_i - \bar{q}_j\) and its norm to represent the direction and distance between connected nodes in the system, in order to capture local geometric structure and pairwise interaction properties. 

\textbf{Input encoding:} The constructed node and edge features are then encoded via separate dense layers,
\begin{equation}
    h_i^{V} = \phi_{V}(v_i) = \sigma(W_V \, v_i + b_V) \in \R^{d_h} \quad \forall v_i \in V, \ \text{and}
    \label{eq:node-encoding}
\end{equation} 
\begin{equation}
    h_{ij}^{E} = \phi_{E}(e_{ij}) = \sigma(W_E \, e_{ij} + b_E) \in \R^{d_h} \quad \forall e_{ij} \in E,
    \label{eq:edge-encoding}
\end{equation}
where \(\phi_V: \R^{d_V} \rightarrow \R^{d_h}\), 
 and \(\phi_E: \R^{d_E} \rightarrow \R^{d_h}\) are
outputs of dense layers that encode the node and edge features, respectively, with weights \(W_V \in \R^{d_h \times d_V}, W_E \in \R^{d_h \times d_E}\) and biases \(b_V, b_E \in \R^{d_h}\).
We denote the activation function (here, \texttt{softplus} or \texttt{gelu}) by $\sigma$. 
Using the symmetric edge features described earlier avoids duplicate memory and computation overhead in the input encoding as well, since we only compute the encoding \(h^E_{ij}\) of each undirected edge feature \(e_{ij}\) where \(i > j\), and use the same encoding for both directions \(h^E_{ij}\) and \(h^E_{ji}\). 

\subsubsection{Message passing and final layer}\label{section:method:model:message-passing}
We perform bidirectional message passing along edges \((i,j)\), where \(A_{ij} = 1\), allowing nodes to aggregate information from their local neighborhoods. 

%
\textbf{Message construction:} Messages are constructed from the encodings of source and edge nodes via a dense layer \(\phi_M: \R^{2d_h} \rightarrow \R^{d_M}\) as 
\begin{equation}
    h^{M}_{ij} = \phi_M\left(
  \begin{bmatrix}
      h_i^V \\ h^E_{ij}
  \end{bmatrix}\right) = \sigma\left(W_M 
  \begin{bmatrix}
      h_i^V \\ h^E_{ij}
  \end{bmatrix} + b_M\right) \in \R^{d_M},
    \label{eq:message}
\end{equation}
with weights \(W_M \in \R^{d_M \times \R^{2d_h}}\) and biases \(b_M \in \R^{d_M}\).

\textbf{Message passing (local pooling):} Each node aggregates incoming messages using a permutation-invariant operation (here, summation)
\( 
    m_j = \sum_{i \in \mathcal{N}_j}{h_{ij}^M},
\)
where \(\mathcal{N}_j\) is the set of neighbors of node \(j\) (source of incoming edges to \(j\) where \(A_{ij} = 1\)). 

\textbf{Graph-level representation (global pooling):} All node embeddings and aggregated messages are pooled to form a global encoding of the network, such that
\begin{equation}
    h_G = \sum_{j = 1}^{N}{\begin{bmatrix}
        h^V_j \\ m_j
    \end{bmatrix}} \in \R^{d_L}, \,\, \text{where} \,\, d_L = d_h + d_M. 
    \label{eq:global-pooling}
\end{equation}

\textbf{Linear layer:} The graph representation is linearly mapped to a scalar value that approximates the conserved value (energy) of the system, such that 
$\widehat{\mathcal{H}} = W_L \cdot h_G + b_L,$
where \(W_L \in \R^{d_L}\) and \(b_L \in \R\)  denote weights and bias of the linear layer, respectively. 
Without loss of generality, we omit the bias term, as it acts only as an integration constant and does not affect the dynamics \(\frac{\partial \mathcal{H}}{\partial q}\) and \(\frac{\partial \mathcal{H}}{\partial p}\), which are our primary interest.
We assume this constant is known to align the model's conserved quantity with the true Hamiltonian \(\mathcal{H}\) in all examples.

\subsection{Training} \label{sec_training}
We now describe the central idea of this paper -- our training algorithm, where we use random feature sampling techniques instead of gradient-descent-based, iterative optimization algorithms. 
In particular, we discuss how to compute dense and linear layer parameters of the network.

\textbf{Dense layer parameters: }
We compute the weights and biases \(W_V, W_E, W_M, b_V, b_E, b_M\) of all the dense layers (\( \phi_V, \phi_E, \phi_M\)) using random sampling algorithms.
Specifically, we use two sampling approaches here: Extreme Learning Machines (ELM) \citep{schmidt-1992-schmidt-neural-network, pao-1992-random-vector-functional-link, huang-2004-elm, huang-2006-universal-approx-elm, rahimi-2008-random-feature, zhang-2012-augmented-elm, leung-2019-elm} and the ``Sample Where It Matters'' (SWIM) algorithm (see \citep{bolager-2023-swim}) for unsupervised learning problems (see \citep{rahma-2024-swim-hnn, datar-2024-swim-pde}). 
As an illustrative example, we describe how to compute the parameters of the dense layer $\phi_V$ and use the notation from \Cref{eq:node-encoding}. 
The parameters of the other dense layers are sampled analogously. 

The \textbf{(ELM) RF-HGN} approach is data-agnostic. The weights \(W_V\) are sampled from the standard normal distribution, and biases \(b_V\) from the standard uniform distribution. 
The \textbf{(SWIM) RF-HGN} approach is data-driven. The network parameters are computed from pairs selected uniformly at random from input data points $x_i$, where the point coordinates correspond to  dense layer inputs.
The weight and the bias of the $i^{\text{th}}$ neuron in the dense layer are constructed using the input data pair \( (x_{i}^{(1)}, x_{i}^{(2)}) \) chosen uniformly at random from all possible pairs, so that \(w_{i} = s_1 ({x_{i}^{(2)} - x_{i}^{(1)}}){ {\lVert x_{i}^{(2)} - x_{i}^{(1)} \rVert}^{-2} }\) and \(b_{i} = - \langle w_{i}, x_{i}^{(1)} \rangle - s_2 \). Here, \( (w_{i}, b_{i}) \) are the weight and bias of the $i^{\text{th}}$ neuron, and \( (s_1, s_2) \) are constants depending on the activation function used in the dense layer (see \citep{bolager-2023-swim} for an analysis).

\textbf{Linear layer parameters: }  
After sampling all the dense layer parameters, we compute the optimal parameters for the linear output layer of the network by computing the least squares solution (see \citep{rahma-2024-swim-hnn}, but also related work~\citep{bertalan-2019-hnn}). 
For an N-body system, we denote a single input to the RF-HGN by \(y \in \R^{2d \cdot N}\) and the output of the global pooling layer by $\Phi (y) \in \R^{d_L}$, and the total number of input data points by $M$. 
The linear system that approximately satisfies Hamilton's equations is then
\begin{equation}\label{eq:fully-linear-system}
  \underbrace{
    \begin{bmatrix}
      \nabla \Phi(y_1)& \cdots  & \nabla \Phi(y_M) & \Phi(y_0) \\
      0 & \cdots & 0 & 1\\
    \end{bmatrix}\tran
  }_{Z \in \R^{(2d \cdot N \cdot M + 1) \times (d_L+1)}}
  \cdot
  \underbrace{
    \begin{bmatrix}
      W_{L}\tran \\
      b_{L}
    \end{bmatrix}
  }_{\theta_L \in \R^{d_L+1}}
  \overset{!}{=}
  \underbrace{
    \begin{bmatrix}
      J^{-1} \dot{y}_1 &
      \cdots & J^{-1} \dot{y}_M & \mathcal{H}(y_0)
    \end{bmatrix}\tran
  }_{u \in \R^{2d \cdot N \cdot M + 1}}.
\end{equation}
\Cref{eq:fully-linear-system} is solved for the linear layer parameters $W_L$ and $b_L$ {using \(l^2\) regularization. The regularization constant in most examples was chosen very small (see \Cref{sec:training_setup}).}
We assume the true Hamiltonian value \( \mathcal{H}(y_0) \) to be known for a single data point to fix the integration constant \(b_L \). 
We assume there is no external force acting on the system during training, such that the total energy is conserved. 
However, we can easily add an external force while evaluating the trajectory during inference.
In the computational experiments, we mostly train with explicitly given time derivatives \(\dot{x}\). We demonstrate training the model purely from time series data as part of the benchmark experiments (\Cref{sec:benchmarking}).
\Cref{eq:fully-linear-system} results in a convex optimization problem, 
\(  [W_{L+1}\tran  b_{L+1}]\tran =\argmin_{\theta_L} { \lVert Z \theta_L - u \rVert }^2\), which can be solved using efficient least-squares algorithms~\citep{meng-2014-leastsquares}.


\textbf{Runtime and memory complexity: }
During training, the sampling of dense layer parameters and gradient computation are fast, with the primary run-time bottleneck being the least squares solve (\Cref{eq:fully-linear-system}).
Assuming $d_L \ll K = 2d \cdot N \cdot M$ (which is always the case in our experiments), the total run-time complexity is $\mathcal{O}(K d^2_L)$. 
This highlights an important feature of our approach: training time scales linearly with data size $M$, the number of particles $N$, and the spatial dimension $d$, given fixed settings for other variables.
The memory complexity during training is $\mathcal{O}(M N_e)$ and thus also scales linearly with the number of edges, and dataset size (see \citep{bolager-2023-swim} and \Cref{app_memory} for details).

\section{Computational experiments}\label{section:results}

\begin{wrapfigure}{l}{0.6\textwidth}

  \centering
  \begin{tabular}{cccc}
  \includegraphics[width=0.125\textwidth]{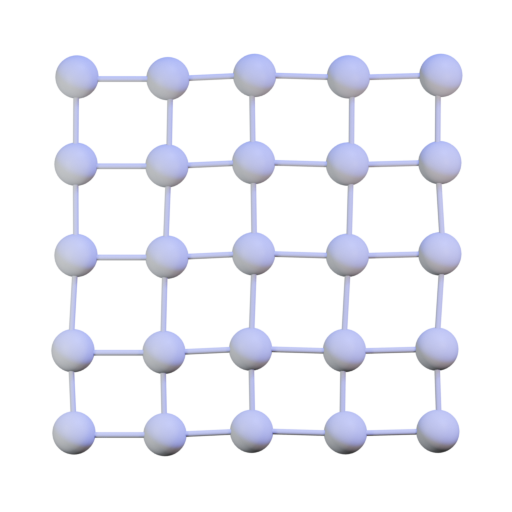}&
  \includegraphics[width=0.125\textwidth]{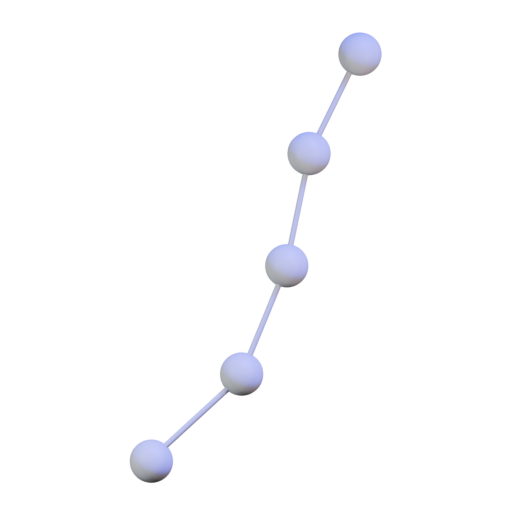}&
  \includegraphics[width=0.125\textwidth]{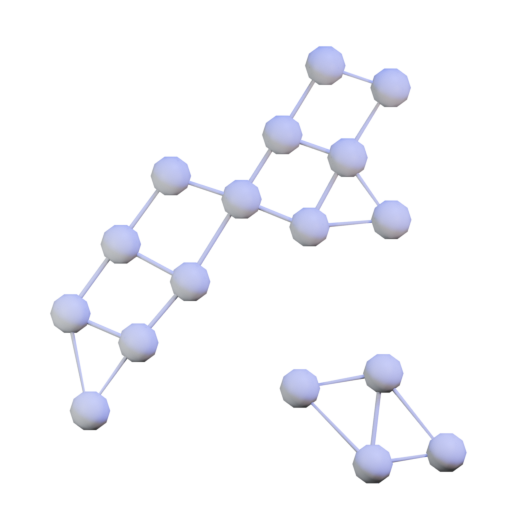}&
  \includegraphics[width=0.125\textwidth]{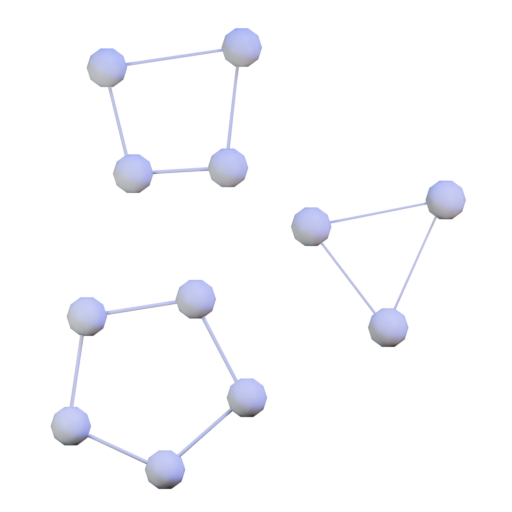}\\
  (a)&(b)&(c)&(d)
  \end{tabular}
  \vspace{4pt}
  \caption{\label{fig:experiments} Graphs considered in the experiments: \textbf{(a)} 3D lattice (nodes arranged on a 2D grid, moving in a 3D space - see \Cref{section:results:optim-study} and \Cref{section:results:zero-shot}), \textbf{(b)} an open chain (nodes moving in 2D space - see \Cref{section:results:zero-shot} {and \ref{sec:results:behcnmark-potentials})}, \textbf{(c)} molecules interacting through Lennard-Jones potential (nodes moving in 2D space with dynamic edges - see section \Cref{section:results:zero-shot}), and \textbf{(d)} 2D closed chain (nodes moving in 2D space - see \Cref{sec:benchmarking}).
    }
\end{wrapfigure}

We evaluate our method on mass-spring and molecular dynamics systems with two and three degrees of freedom (2D and 3D), as illustrated in \Cref{fig:experiments}.
We provide further details on the used datasets, setup, and hardware in \Cref{sec:datasets}, \Cref{sec:training_setup}, and \Cref{sec:hardware}, respectively.
We additionally discuss hyperparameter tuning and {ablation studies of increasing feature widths and number of message passes in \Cref{appendix:ablation-studies}}, test robustness against noise in \Cref{appendix:sec:noise-study}, demonstrate batch-wise training in \Cref{appendix:sec:batch-wise}, and provide further random feature benchmarks in \Cref{appendix:sec:different-rf}.
For $x_{\text{true}}, x_{\text{pred}} \in \R^{m}$ for $m \in \mathbb{N}$, we define the relative error as $ ||x_{\text{true}} - x_{\text{pred}}  ||_2 / ||x_{\text{true}}||_2$.

\captionsetup{skip=0pt} 

\subsection{Benchmarking against SOTA optimizers}
\label{section:results:optim-study}

\newcommand{\tabResultsOptimizerCaption}{Results of training the HGN architecture for the 3D lattice system (see \Cref{fig:experiments} (a)) with different optimizers.
Results show \textbf{mean (min, max)} over three runs on the same GPU hardware.
}

\begin{table}[h]
\caption{\label{tab:results_optimizers}\tabResultsOptimizerCaption}
\centering
\scriptsize
\begin{tabular}{p{3.7cm}lll}
\toprule
\textbf{{Optimizer}}&\textbf{{Test MSE}}&\textbf{Train time [s]}&\textbf{{Speed-up}}\\
\toprule
\textbf{RF-HGN (ours)}
&8.95e-5 (6.96e-5, 1.13e-4)&\textbf{0.16 (0.13, 0.22)}&-\\
LBFGS~\citep{liu-1989}&\textbf{3.56e-5 (1.21e-5, 7.94e-5)}&23.85 (23.71, 23.95)&148.96×\\
Rprop~\citep{riedmiller-1993}&9.59e-4 (7.49e-5, 2.63e-3)&30.84 (30.74, 30.94)&192.62×\\
RMSprop~\citep{tieleman-2012}&1.09e-3 (2.55e-5, 3.18e-3)&91.62 (91.13, 92.42)&572.24×\\
Adam~\citep{adam-optimizer}&2.90e-3 (4.11e-5, 8.53e-3)&91.64 (89.97,  92.66)&572.37×\\
SGD+momentum~\citep{sutskever-2013}&4.23e-3 (3.81e-3, 4.66e-3)&91.65 (91.14, 92.20)&572.43×\\
SGD~\citep{sgd-optimizer} &2.36e-2 (1.85e-2, 2.86e-2)&91.75 (91.51, 91.91)&573.07×\\
Adagrad~\citep{duchi-2011}&2.58e-2 (2.88e-3, 7.05e-2)&92.03 (91.58, 92.47) &574.84×\\
AdamW~\citep{loshchilov-2019}&2.91e-3 (4.30e-5, 8.53e-3)&92.15 (91.86, 92.31)&575.59×\\
Adamax~\citep{adam-optimizer}&1.85e-3 (1.55e-4, 4.32e-3)&92.33 (92.07, 92.69)&576.68×\\
Adadelta~\citep{zeiler-2012}&8.11e-3 (1.49e-3, 1.96e-2)&92.60 (92.25, 93.07) &578.39×\\
Radam~\citep{liu-2021}&1.69e-3 (5.36e-5, 4.75e-3)&93.00 (92.75, 93.42) &580.88×\\
Nadam~\citep{dozat-2016}&9.11e-4 (4.08e-5, 2.61e-3)&93.42 (92.88, 93.80) &583.54×\\
Averaged SGD~\citep{averaged-sgd-optimizer}&2.36e-2 (1.85e-2, 2.86e-2)&94.50 (94.01, 94.78)&590.26×\\
Adafactor~\citep{shazeer-2018}&2.41e-3 (1.06e-3, 4.71e-3) &96.36 (95.67, 96.88) & 601.92×\\
\bottomrule
\end{tabular}
\end{table}

The goal of this experiment is to demonstrate the efficiency of our training approach in comparison with the conventional training methods that rely on SOTA iterative optimization algorithms.
To this end, we consider all of the existing optimizers available in PyTorch~\citep{pytorch} as the current SOTA iterative training procedures.
%
In this experiment, the target function is the Hamiltonian of a generalized \(N_x \times N_y\)-body lattice mass-spring system with
a spatial dimension $d = 3$ given by \Cref{eq:lattice}.
\Cref{tab:results_optimizers} lists the results of training the HGN architecture for the lattice system with different optimizers, sorted by training time. The hyperparameters are tuned for each optimizer separately, and early stopping was used for all iterative approaches.
\textbf{Our proposed training method significantly outperforms all iterative approaches in terms of training time} by a factor of \textbf{148 up to 601}, and is only slightly less accurate compared to the LBFGS method (a second-order optimizer).
%
\Cref{sec:benchmarking} includes comparisons for other graph network architectures on benchmark datasets, where we observed similar results.

\subsection{Zero-shot generalization and comparison of random feature methods}
\label{section:results:zero-shot}

\begin{figure}[h]
    \centering
    \includegraphics[width=0.99\linewidth]{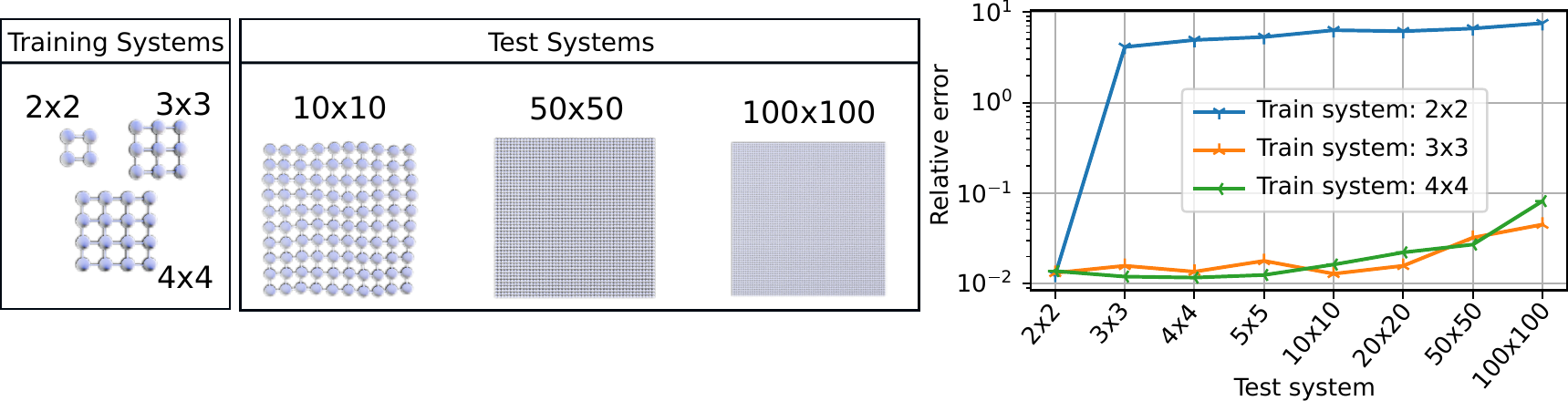}
    \vspace{4pt}
    \caption{
    Illustration of accurate zero-shot generalization for 3D lattice (see \Cref{fig:experiments} (a)): Training on smaller systems (left) enables accurate predictions (right) on extremely large test systems (middle).
    }
    \label{fig:zero_shot}
\end{figure}

We now study zero-shot generalization, where we train an RF-HGN on small systems of size 2x2, 3x3, and 4x4, and test on systems going from 2x2 up to 100x100. \Cref{fig:zero_shot} shows that we can accurately approximate a Hamiltonian for much smaller systems with 3x3 nodes and reliably predict it with extremely large systems of size 100x100 without retraining.
A 2x2 system is an edge case where all nodes have only two edges, lacking the nodes with four edges, as in the test data, explaining the poor zero-shot generalization.
%
%

\begin{figure}[h]
  \begin{center}
    \includegraphics[width=0.99\textwidth]{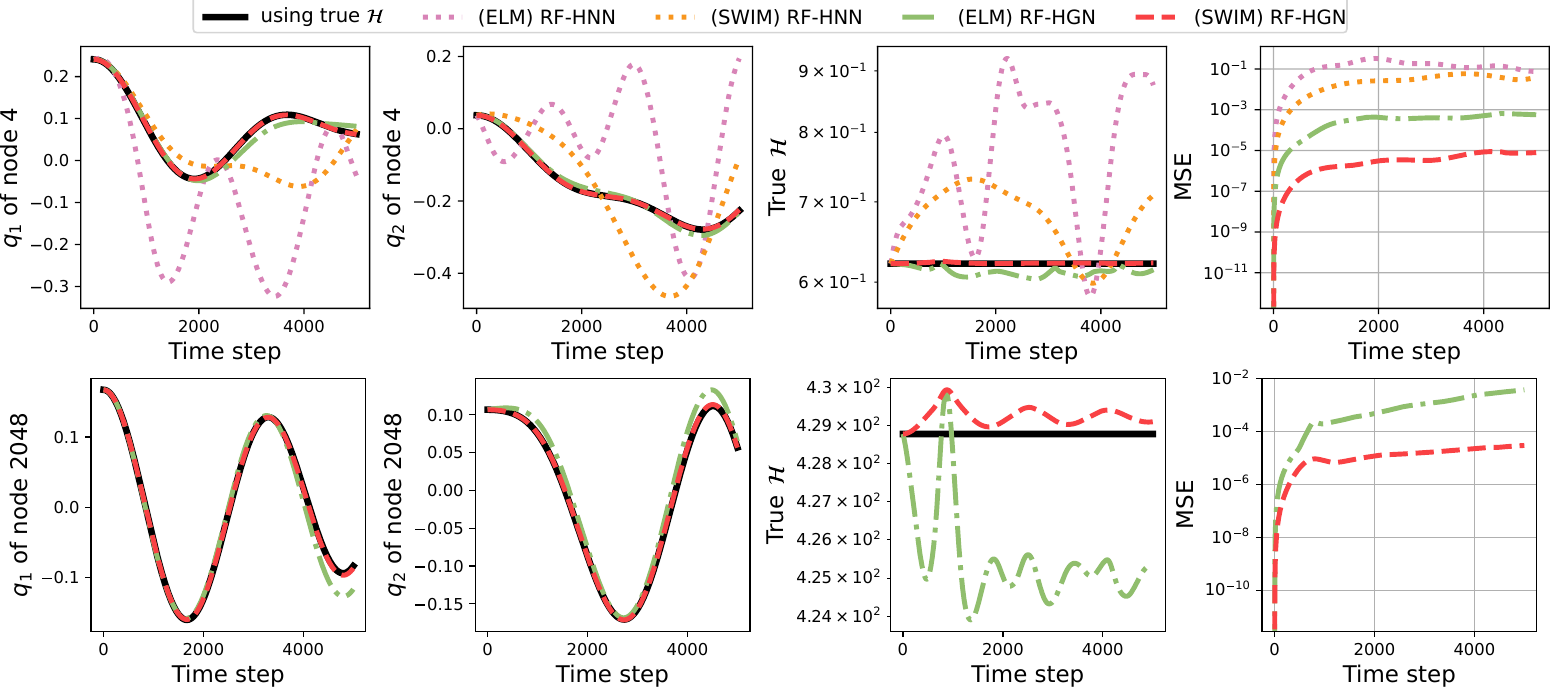}
  \end{center}
  \caption{
    Illustration of position trajectories (first two columns), their true Hamiltonian values (third column), and MSE (fourth column) over time from models trained on a system with eight nodes on the 2D open chain (see \Cref{fig:experiments} (b)).
    Top row: Results from RF-HNN and RF-HGN architectures are visualized along with the ground truth. A system of the same size for training and testing is used (\(2^3\) nodes). Bottom row: Results from RF-HGN architectures and ground truth for a zero-shot test case with a system size of \(2^{12}\) nodes (trained on \(2^3\) nodes).
  }
  \label{fig:results:node-scaling-integration-8-and-4096-all}
\end{figure}


We consider another example of an N-body chain mass-spring system in a 2D space (\cref{eq:chain-hamiltonian}). \Cref{fig:results:node-scaling} shows that by training on a much smaller system with $2^3$ nodes, RF-HGN trained with ELM and SWIM results in very low errors for systems as large as $2^{12}$ nodes.
This demonstrates robust and strong zero-shot generalization in graph-based architectures, with training using SWIM outperforming ELM by approximately an order of magnitude.
Since HNNs are not graph-based architectures, they have to be re-trained for each system, and even after re-training, they perform much poorly in comparison with graph-based architectures (by 1-2 orders of magnitude).
We also observe that the graph-based architectures (RF-HGN) are slower to train compared to their respective counterparts trained with fully connected networks (RF-HNN), but are more accurate by 1-2 orders of magnitude, especially for large systems, even without re-training.



\Cref{fig:results:node-scaling-integration-8-and-4096-all} (top row) shows that the trajectories evaluated with the RF-HGN accurately match the true ones closely while approximately conserving the Hamiltonian, unlike RF-HNN.
The RF-HGN trained with SWIM, in particular, outperforms the one trained with ELM by roughly two orders of magnitude.
In the bottom row of \Cref{fig:results:node-scaling-integration-8-and-4096-all}, we show zero-shot results, which are limited to only the graph networks.
Our (SWIM) RF-HGN is robust and again exhibits a low error, while the ELM-trained model exhibits deviations from the true Hamiltonian.

\label{section:results:learn-traj}

We now consider another Hamiltonian, using the Lennard-Jones potential (\cref{eq:lennard-jones-hamiltonian}) to investigate generalization properties to different geometries. First, a small system with 9 particles was trained with Adam and the random feature methods ELM and SWIM. The accuracy with ELM was poor (see \Cref{fig:results:9particles-lj-snaps}), and thus the results are omitted in \Cref{fig:results:9-lj-traj}. We speculate that the main reason for poor performance is the non-isotropic input variables, which makes normal distributions a bad choice for the weights. The results are given in \Cref{tab:9-lj-swim}, \Cref{fig:results:9-lj-traj}, \Cref{tab:9-lj-elm}, \Cref{tab:9-lj-adam}, and \Cref{fig:results:9particles-lj-snaps}.
To evaluate RF-HGN in a more complex scenario, we employed dynamic edge indices with a cutoff of \(2.0\), trained our model with 36 particles, and tested with 64 particles to test zero-shot generalization. We visualize the rollout trajectories in \Cref{fig:results:32train-64test-lj-snaps} and \Cref{fig:results:64-lj-traj}, and observe that SWIM sampling clearly outperformed ELM, while maintaining slightly worse approximation than the Adam-trained HGN. We note that none of the trainers could reach low approximation errors (\(\sim10\%\) relative error). To the best of our knowledge, our RF-HGN is the first random feature-based physics-informed graph network, and can be trained approximately 100 times faster than with the Adam optimizer at comparable accuracy.

\begin{table}[h]
    \centering
    \caption{Molecular dynamics evaluation with 9 particles. Mean squared error (MSE) and relative \(l^2\) error (rel. \(l^2\)) are reported together with the true Hamiltonian over the ground-truth trajectory and the (SWIM) RF-HGN predicted quantity over the rolled-out trajectory.}
    \label{tab:9-lj-swim}
    \scriptsize
    \begin{tabular}{llllll}
    \toprule
        & T=1 & T=25000 & T=50000 & T=74999 & T=99999 \\
    \midrule
    \(q\) MSE & 1.140e-13 & 3.239e-03 & 1.998e-02 & 4.932e-02 & 8.301e-02 \\
    \(q\) rel. \(l^2\) & 2.346e-07 &  3.978e-02 &  9.717e-02  & 1.545e-01  & 2.032e-01 \\
    True \(\mathcal{H}\) & -1.233e+01 & -1.233e+01 & -1.233e+01 & -1.233e+01 & -1.233e+01 \\
    Model \(\widehat{\mathcal{H}}\) & -1.223e+01 & -1.223e+01 & -1.248e+01 & -1.214e+01 & -1.217e+01 \\
    \bottomrule
    \end{tabular}
\end{table}


\subsection{Benchmarking with real-world potentials with increasing complexity}\label{sec:results:behcnmark-potentials}
%
%
%
To further evaluate the applicability of RF-HGN, we experiment with additional potentials from quantum mechanics (anharmonic oscillator \citet{anharmonic-potential} and molecular dynamics (the Morse potential \citet{morse-potential}), and list testing results with an unseen initial condition in \Cref{tab:chain-potential-gravity}. In addition to using more complicated potential, we also now apply an external (gravitational) force node-wise during integration and simulate for a long-time horizon (see \Cref{fig:spring-chain-traj}, \ref{fig:anharmonic-chain-traj}, \ref{fig:morse-chain-traj} for learned Hamiltonian plots over the trajectory; see \Cref{fig:spring-chain-snaps}, \ref{fig:anharmonic-chain-snaps}, \ref{fig:morse-chain-snaps} snapshot visualization of the predicted trajectories). The results show that more challenging potentials (non-linear forces) than the standard mass-spring (linear force) can also be approximated with the RF-HGN model with reasonable accuracy, compared to the Adam optimizer, still achieving 200-300× speed-ups, even without GPU acceleration. See \Cref{tab:chain-potential-gravity} for details, and note that the RF-HGN results are achieved without extensive hyperparameter tuning (as opposed to Adam).
\begin{table}[h]
    \centering
    \caption{Zero-shot (trained with 5 nodes, tested with 10 nodes) test evaluation by solving an unseen initial condition of a 2D chain (see \Cref{fig:experiments} (b)) using Hamiltonian graph models trained differently. The system is solved for 10,000 time steps with time step size \(\Delta t = 10^{-2}\) with gravitational force applied during integration, and the last position MSE is reported against the reference solution using the true Hamiltonian. The results are listed for the standard spring, anharmonic oscillator, and the Morse potential in that order.
    }
    \scriptsize
    \begin{tabular}{lccc}
        \toprule
        Potential & (Adam) HGN & (ELM) RF-HGN & (SWIM) RF-HGN \\
        \midrule
        \(V(r) = \frac{1}{2} \beta r^2\) & 3.875e-03 & 2.331e-03 & 3.408e-05 \\
        \(V(r) = \frac{1}{2} \beta r^2+\frac{1}{4} \eta r^4\) & 4.562e-02 & 4.324e-02 & 5.232e-04 \\
        \(V(r) = D(1-\exp( -ar))^2\) & 8.893e-02 & 7.398e-04 & 1.218e-03 \\
        \bottomrule
    \end{tabular}
    \label{tab:chain-potential-gravity}
\end{table}

\subsection{Benchmarking against SOTA architectures}
\label{sec:benchmarking}
The goal of this benchmark is to compare the results of our model with the existing state-of-the-art graph-based network architectures used to model physical systems.
The Adam optimizer~\citep{adam-optimizer}, widely regarded as the SOTA optimizer for physics-informed GNNs \citep{kumar-2023, thangamuthu-2022}, is used as the default in our comparisons.
We use the dataset and code from \citet{thangamuthu-2022}, introduced at the NeurIPS 2022 Datasets and Benchmarks Track.
In line with the other experiments, we observe excellent performance on the benchmark spring systems (\Cref{fig:experiments}, d) with orders of magnitude faster training times while maintaining a comparable accuracy.
Training times are reported in \Cref{tab:benchmarking}.
In our evaluations, certain specialized architectures such as Lagrangian Graph Networks (LGN) \citep{bhattoo-2022-lgn-ridig} occasionally exhibit instability and diverge on test trajectories, whereas our model maintains robust performance.
For a detailed problem setup, accuracy comparison, and the architectures, please see \Cref{sec:benchmarking_appendix}.


\begin{table}[h]
    \centering
    \caption{Comparison of training times (in seconds) for RF-HGN optimized using SWIM to existing physics-informed graph models optimized with Adam on a benchmark dataset from \cite{thangamuthu-2022} on a 2D closed chain (see \Cref{fig:experiments} (d)).
    }
    \scriptsize
    \begin{tabular}{p{1cm} p{1.4cm} p{0.8cm} p{1.15cm} p{1.1cm} p{1.2cm}p{1.2cm}p{1.2cm}p{1.2cm}}
        \toprule
        System size & (SWIM) RF-HGN & FGNN & FGNODE & GNODE & LGN & LGNN & HGN & HGNN \\
        \midrule
        \(N=3\) & \textbf{2.51} &  406.14 & 380.35  & 2367.37 &  12534.81 & 7225.88 & 1288.08  & 3568.12  \\
        \(N=4\) & \textbf{3.87} & 475.24 &  430.32 &  2499.04 &  20536.78  & 6259.58 & 1370.14 & 4021.59 \\
        \(N=5\)  & \textbf{5.42} & 536.27  &  520.54 & 2600.31 & 53148.24  & 8774.59  & 1676.78 & 4380.46 \\
         \bottomrule
    \end{tabular}
    \label{tab:benchmarking}
\end{table}

\section{Conclusion}
We propose a training algorithm for Hamiltonian graph networks via rapid random feature sampling and linear system solvers. Our approach completely avoids slow, iterative gradient-descent-based optimization, which is especially challenging in the graph network and the physics-informed settings. 
We demonstrate our approach on chain, lattice, and molecular systems in up to three spatial dimensions, encompassing N-body systems. 
By incorporating translation, rotation, and index-permutation invariances, we extend random feature methods to graph-based Hamiltonian network architectures.
Compared to 15 optimizer baselines, our method offers dramatic speedups (100× to 1000×) while achieving competitive accuracy in 3D physical systems. Remarkably, training on $3 \times 3$ systems suffices to accurately predict dynamics in systems of size $100 \times 100$, demonstrating strong zero-shot generalization capabilities. 
With this generalization from such small-scale training systems, one can deploy models without needing to re-train on full-scale data, enabling fast prototyping.

\paragraph{Limitations and future work:} 

%
For very small graphs, the HNN architecture is often faster to train than a graph-based approach, making it a better choice than HGN in these cases.
%
Generalization capabilities of the HGN models are typically limited to the same type of graphs, i.e., models trained on chains (edge degrees up to two) cannot be used to predict dynamics of lattices (edge degrees up to four) \citep{corso-2024}. We observed similar challenges when using dynamic edges in the molecular dynamics examples for all optimizers. Therefore, for those examples, one could either use more data or a less physically complex model (e.g., an SE(3) graph network that provides SE(3) equivariance by construction \citep{du-2022-se3}) to reduce data requirements.
Our approach does not easily generalize to other graph neural network architectures (e.g., convolution or self-attention on the individual node features). In future work, we intend to extend this work by employing multiple message passing layers and deeper architectures where random feature boosting might help \citep{zozoulenko-2025}.

\paragraph{Ethics statement:}
We demonstrate that data-driven construction of random features can significantly outperform many SOTA optimizers in terms of accuracy and training time. 
The tremendously increased training speeds we report may also speed up the development of nefarious and even dangerous applications. Similar to all HGN and HNN models, our specific training method is not designed for this purpose. However, specific bad intent as well as significant further development would be required for this to happen.
We hope that, instead, our work has a profound positive societal impact in the future, because training such accurate models from data is important in many sciences as well as in engineering -- but has been slow up to now, due to the difficulties in training.

\paragraph{Reproducibility statement:}
We provide details on the used datasets, model setup with hyperparameters, and hardware details in \Cref{sec:datasets}, \Cref{sec:training_setup}, and \Cref{sec:hardware}, respectively for all the numerical experiments discussed in the main text and in the appendix. In the supplementary materials we provide further instructions in a \texttt{README.md} on how to reproduce all results in the form of tables and plots. Our codebase is publicly available at 
\begin{center}
    \url{https://gitlab.com/fd-research/swimhgn}.    
\end{center}
\newpage

\paragraph{Acknowledgments:}
We are grateful for discussions with Samuel James Newcome, Manish Kumar Mishra, Markus Mühlhäußer, Jonas Schuhmacher, Iryna Burak, Nadiia Derevianko, Qing Sun, and Erik Lien Bolager. A.R. is supported by the BMBF (project AutoMD-AI), A.\v C. is supported by the TUM Georg Nemetschek Institute -- Artificial Intelligence for the Built World., and C.D. and F.D. are supported by the DFG (project no. 468830823), and acknowledge association with DFG-SPP-22. C.D. is partially funded by the Institute for Advanced Study (IAS) at the Technical University of Munich. The authors gratefully acknowledge the computational and data resources as well as the support provided by the Leibniz Supercomputing Centre (www.lrz.de).

\bibliography{iclr2026_conference}

@article{hamilton-1834,
  author = {William Rowan Hamilton},
  title = {On a General Method in Dynamics},
  journal = {Philosophical Transactions of the Royal Society},
  year = {1834},
  volume = {124},
  pages = {247-308},
}

@article{hamilton-1835,
  author = {William Rowan Hamilton},
  title = {Second Essay on a General Method in Dynamics},
  journal = {Philosophical Transactions of the Royal Society},
  year = {1835},
  volume = {125},
  pages = {95-144},
}

@article{morse-potential,
  title={Diatomic molecules according to the wave mechanics. II. Vibrational levels},
  author={Morse, Philip M},
  journal={Physical review},
  volume={34},
  number={1},
  pages={57},
  year={1929},
  publisher={APS}
}

@book{anharmonic-potential,
  title={Physical chemistry for the life sciences},
  author={Atkins, Peter William and Ratcliffe, R George and De Paula, Julio and Wormald, Mark},
  year={2023},
  publisher={Oxford University Press}
}

@inproceedings{adam-optimizer,
  title = {Adam: {{A Method}} for {{Stochastic Optimization}}},
  booktitle = {International {{Conference}} on {{Learning Representations ICLR}} 2015},
  author = {Kingma, D. P. and Ba, L. J.},
  year = {2015}
}

@incollection{pytorch,
  title = {PyTorch: An Imperative Style, High-Performance Deep Learning Library},
  author = {Paszke, Adam and Gross, Sam and Massa, Francisco and Lerer, Adam and Bradbury, James and Chanan, Gregory and Killeen, Trevor and Lin, Zeming and Gimelshein, Natalia and Antiga, Luca and Desmaison, Alban and Kopf, Andreas and Yang, Edward and DeVito, Zachary and Raison, Martin and Tejani, Alykhan and Chilamkurthy, Sasank and Steiner, Benoit and Fang, Lu and Bai, Junjie and Chintala, Soumith},
  booktitle = {Advances in Neural Information Processing Systems 32},
  pages = {8024--8035},
  year = {2019},
  publisher = {Curran Associates, Inc.}
}

@article{watters-2017-physics-priors,
  title={Visual interaction networks: Learning a physics simulator from video},
  author={Watters, Nicholas and Zoran, Daniel and Weber, Theophane and Battaglia, Peter and Pascanu, Razvan and Tacchetti, Andrea},
  journal={Advances in neural information processing systems},
  volume={30},
  year={2017}
}

@article{de-2018-physics-priors,
  title={End-to-end differentiable physics for learning and control},
  author={de Avila Belbute-Peres, Filipe and Smith, Kevin and Allen, Kelsey and Tenenbaum, Josh and Kolter, J Zico},
  journal={Advances in neural information processing systems},
  volume={31},
  year={2018}
}

@article{chang-2016-physics-priors,
  title={A compositional object-based approach to learning physical dynamics},
  author={Chang, Michael B and Ullman, Tomer and Torralba, Antonio and Tenenbaum, Joshua B},
  journal={arXiv},
  eprint={1612.00341},
  year={2016},
pubstate={prepublished}
}

@article{tenenbaum-2000-physics-priors,
  title={A global geometric framework for nonlinear dimensionality reduction},
  author={Tenenbaum, Joshua B and Silva, Vin de and Langford, John C},
  journal={science},
  volume={290},
  number={5500},
  pages={2319--2323},
  year={2000},
  publisher={American Association for the Advancement of Science}
}

@misc{nabian-2024,
      title={X-MeshGraphNet: Scalable Multi-Scale Graph Neural Networks for Physics Simulation}, 
      author={Mohammad Amin Nabian and Chang Liu and Rishikesh Ranade and Sanjay Choudhry},
      year={2024},
      eprint={2411.17164},
      archivePrefix={arXiv},
      primaryClass={cs.LG},
  pubstate = {prepublished}
}

@misc{yu-2025,
      title={PIORF: Physics-Informed Ollivier-Ricci Flow for Long-Range Interactions in Mesh Graph Neural Networks}, 
      author={Youn-Yeol Yu and Jeongwhan Choi and Jaehyeon Park and Kookjin Lee and Noseong Park},
      year={2025},
      eprint={2504.04052},
      archivePrefix={arXiv},
      primaryClass={cs.LG}
}

@inproceedings{schmidt-2021,
  title = {Descending through a Crowded Valley - Benchmarking Deep Learning Optimizers},
  booktitle = {Proceedings of the 38th International Conference on Machine Learning},
  author = {Schmidt, Robin M and Schneider, Frank and Hennig, Philipp},
  editor = {Meila, Marina and Zhang, Tong},
  year = {2021},
  series = {Proceedings of Machine Learning Research},
  volume = {139},
  pages = {9367--9376},
  publisher = {PMLR}
}

@inproceedings{gupta-2024,
    author = {Gupta, Vipul and Chen, Xin and Huang, Ruoyun and Meng, Fanlong and Chen, Jianjun and Yan, Yujun},
    title = {GraphScale: A Framework to Enable Machine Learning over Billion-node Graphs},
    year = {2024},
    isbn = {9798400704369},
    publisher = {Association for Computing Machinery},
    address = {New York, NY, USA},
    doi = {10.1145/3627673.3680021},
    booktitle = {Proceedings of the 33rd ACM International Conference on Information and Knowledge Management},
    pages = {4514–4521},
    numpages = {8},
    location = {Boise, ID, USA},
    series = {CIKM '24}
}

@inproceedings{zhu-2024,
    author = {Zhu, Zeyu and Wang, Peisong and Hu, Qinghao and Li, Gang and Liang, Xiaoyao and Cheng, Jian},
    title = {FastGL: A GPU-Efficient Framework for Accelerating Sampling-Based GNN Training at Large Scale},
    year = {2025},
    isbn = {9798400703911},
    publisher = {Association for Computing Machinery},
    address = {New York, NY, USA},
    doi = {10.1145/3622781.3674167},
    booktitle = {Proceedings of the 29th ACM International Conference on Architectural Support for Programming Languages and Operating Systems, Volume 4},
    pages = {94–110},
    numpages = {17},
    location = {Hilton La Jolla Torrey Pines, La Jolla, CA, USA},
    series = {ASPLOS '24}
}

@INPROCEEDINGS{zhang-2022,
  author={Zhang, Xin and Shen, Yanyan and Chen, Lei},
  booktitle={2022 IEEE International Conference on Data Mining (ICDM)}, 
  title={Feature-Oriented Sampling for Fast and Scalable GNN Training}, 
  year={2022},
  volume={},
  number={},
  pages={723-732},
  doi={10.1109/ICDM54844.2022.00083}}

@INPROCEEDINGS{zhang-2021,
  author={Zhang, Lizhi and Lai, Zhiquan and Li, Shengwei and Tang, Yu and Liu, Feng and Li, Dongsheng},
  booktitle={2021 IEEE International Conference on Cluster Computing (CLUSTER)}, 
  title={2PGraph: Accelerating GNN Training over Large Graphs on GPU Clusters}, 
  year={2021},
  volume={},
  number={},
  pages={103-113},
  doi={10.1109/Cluster48925.2021.00036}}

@inproceedings{yang-2022,
author = {Yang, Jianbang and Tang, Dahai and Song, Xiaoniu and Wang, Lei and Yin, Qiang and Chen, Rong and Yu, Wenyuan and Zhou, Jingren},
title = {GNNLab: a factored system for sample-based GNN training over GPUs},
year = {2022},
isbn = {9781450391627},
publisher = {Association for Computing Machinery},
address = {New York, NY, USA},
doi = {10.1145/3492321.3519557},
booktitle = {Proceedings of the Seventeenth European Conference on Computer Systems},
pages = {417–434},
numpages = {18},
location = {Rennes, France},
series = {EuroSys '22}
}

@article{zhou-2022,
author = {Zhou, Hongkuan and Zheng, Da and Nisa, Israt and Ioannidis, Vasileios and Song, Xiang and Karypis, George},
title = {TGL: a general framework for temporal GNN training on billion-scale graphs},
year = {2022},
issue_date = {April 2022},
publisher = {VLDB Endowment},
volume = {15},
number = {8},
issn = {2150-8097},
doi = {10.14778/3529337.3529342},
journal = {Proc. VLDB Endow.},
month = apr,
pages = {1572–1580},
numpages = {9}
}

@article{wan-2023,
author = {Wan, Xinchen and Xu, Kaiqiang and Liao, Xudong and Jin, Yilun and Chen, Kai and Jin, Xin},
title = {Scalable and Efficient Full-Graph GNN Training for Large Graphs},
year = {2023},
issue_date = {June 2023},
publisher = {Association for Computing Machinery},
address = {New York, NY, USA},
volume = {1},
number = {2},
doi = {10.1145/3589288},
journal = {Proc. ACM Manag. Data},
month = jun,
articleno = {143},
numpages = {23}
}

@article{pao-1992-random-vector-functional-link,
  title={Functional-link net computing: theory, system architecture, and functionalities},
  author={Pao, Y-H and Takefuji, Yoshiyasu},
  journal={Computer},
  volume={25},
  number={5},
  pages={76--79},
  year={1992},
  publisher={IEEE}
}

@inproceedings{schmidt-1992-schmidt-neural-network,
  title={Feed forward neural networks with random weights},
  author={Schmidt, Wouter F and Kraaijveld, Martin A and Duin, Robert PW and others},
  booktitle={International conference on pattern recognition},
  pages={1--1},
  year={1992},
  organization={IEEE Computer Society Press}
}

@inproceedings{cai-2021,
author = {Cai, Zhenkun and Yan, Xiao and Wu, Yidi and Ma, Kaihao and Cheng, James and Yu, Fan},
title = {DGCL: an efficient communication library for distributed GNN training},
year = {2021},
isbn = {9781450383349},
publisher = {Association for Computing Machinery},
address = {New York, NY, USA},
doi = {10.1145/3447786.3456233},
booktitle = {Proceedings of the Sixteenth European Conference on Computer Systems},
pages = {130–144},
numpages = {15},
location = {Online Event, United Kingdom},
series = {EuroSys '21}
}

@article{shao-2024,
author = {Shao, Yingxia and Li, Hongzheng and Gu, Xizhi and Yin, Hongbo and Li, Yawen and Miao, Xupeng and Zhang, Wentao and Cui, Bin and Chen, Lei},
title = {Distributed Graph Neural Network Training: A Survey},
year = {2024},
issue_date = {August 2024},
publisher = {Association for Computing Machinery},
address = {New York, NY, USA},
volume = {56},
number = {8},
issn = {0360-0300},
doi = {10.1145/3648358},
journal = {ACM Comput. Surv.},
month = apr,
articleno = {191},
numpages = {39}
}

@inproceedings{lin-2020,
author = {Lin, Zhiqi and Li, Cheng and Miao, Youshan and Liu, Yunxin and Xu, Yinlong},
title = {PaGraph: Scaling GNN training on large graphs via computation-aware caching},
year = {2020},
isbn = {9781450381376},
publisher = {Association for Computing Machinery},
address = {New York, NY, USA},
doi = {10.1145/3419111.3421281},
booktitle = {Proceedings of the 11th ACM Symposium on Cloud Computing},
pages = {401–415},
numpages = {15},
location = {Virtual Event, USA},
series = {SoCC '20}
}

@article{kose-2023,
title={Fast\&Fair: Training Acceleration and Bias Mitigation for {GNN}s},
author={Oyku Deniz Kose and Yanning Shen},
journal={Transactions on Machine Learning Research},
issn={2835-8856},
year={2023},
url={https://openreview.net/forum?id=nOk4XEB7Ke},
note={}
}

@inproceedings{kaler-2022,
 author = {Kaler, Tim and Stathas, Nickolas and Ouyang, Anne and Iliopoulos, Alexandros-Stavros and Schardl, Tao and Leiserson, Charles E. and Chen, Jie},
 booktitle = {Proceedings of Machine Learning and Systems},
 editor = {D. Marculescu and Y. Chi and C. Wu},
 pages = {172--189},
 title = {Accelerating Training and Inference of Graph Neural Networks with Fast Sampling and Pipelining},
 volume = {4},
 year = {2022}
}

@inproceedings{wang-2021,
author = {Wang, Lei and Yin, Qiang and Tian, Chao and Yang, Jianbang and Chen, Rong and Yu, Wenyuan and Yao, Zihang and Zhou, Jingren},
title = {FlexGraph: a flexible and efficient distributed framework for GNN training},
year = {2021},
isbn = {9781450383349},
publisher = {Association for Computing Machinery},
address = {New York, NY, USA},
doi = {10.1145/3447786.3456229},
booktitle = {Proceedings of the Sixteenth European Conference on Computer Systems},
pages = {67–82},
numpages = {16},
location = {Online Event, United Kingdom},
series = {EuroSys '21}
}

@article{bravo-hermsdorff-2019,
  title={A unifying framework for spectrum-preserving graph sparsification and coarsening},
  author={Bravo Hermsdorff, Gecia and Gunderson, Lee},
  journal={Advances in Neural Information Processing Systems},
  volume={32},
  year={2019}
}

@misc{hashemi-2024,
      title={A Comprehensive Survey on Graph Reduction: Sparsification, Coarsening, and Condensation}, 
      author={Mohammad Hashemi and Shengbo Gong and Juntong Ni and Wenqi Fan and B. Aditya Prakash and Wei Jin},
      year={2024},
      eprint={2402.03358},
      archivePrefix={arXiv},
      primaryClass={cs.SI},
}

@inproceedings{jin-2022,
author = {Jin, Wei and Tang, Xianfeng and Jiang, Haoming and Li, Zheng and Zhang, Danqing and Tang, Jiliang and Yin, Bing},
title = {Condensing Graphs via One-Step Gradient Matching},
year = {2022},
isbn = {9781450393850},
publisher = {Association for Computing Machinery},
address = {New York, NY, USA},
doi = {10.1145/3534678.3539429},
booktitle = {Proceedings of the 28th ACM SIGKDD Conference on Knowledge Discovery and Data Mining},
pages = {720–730},
numpages = {11},
location = {Washington DC, USA},
series = {KDD '22}
}

@inproceedings{jin-2022a,
title={Graph Condensation for Graph Neural Networks},
author={Wei Jin and Lingxiao Zhao and Shichang Zhang and Yozen Liu and Jiliang Tang and Neil Shah},
booktitle={International Conference on Learning Representations},
year={2022},
url={https://openreview.net/forum?id=WLEx3Jo4QaB}
}

@InProceedings{kumar-2023,
  title = 	 {Featured Graph Coarsening with Similarity Guarantees},
  author =       {Kumar, Manoj and Sharma, Anurag and Saxena, Shashwat and Kumar, Sandeep},
  booktitle = 	 {Proceedings of the 40th International Conference on Machine Learning},
  pages = 	 {17953--17975},
  year = 	 {2023},
  editor = 	 {Krause, Andreas and Brunskill, Emma and Cho, Kyunghyun and Engelhardt, Barbara and Sabato, Sivan and Scarlett, Jonathan},
  volume = 	 {202},
  series = 	 {Proceedings of Machine Learning Research},
  month = 	 jul,
  publisher =    {PMLR}
}

@inproceedings{vignac-2020,
 author = {Vignac, Cl\'{e}ment and Loukas, Andreas and Frossard, Pascal},
 booktitle = {Advances in Neural Information Processing Systems},
 editor = {H. Larochelle and M. Ranzato and R. Hadsell and M.F. Balcan and H. Lin},
 pages = {14143--14155},
 publisher = {Curran Associates, Inc.},
 title = {Building powerful and equivariant graph neural networks with structural message-passing},
 volume = {33},
 year = {2020}
}

@article{marino-2025,
author = {Marino, Antonio and Pacchierotti, Claudio and Robuffo Giordano, Paolo},
title = {A Gated Graph Neural Network Approach to Fast-Convergent Dynamic Average Estimation},
year = {2025},
publisher = {Association for Computing Machinery},
address = {New York, NY, USA},
issn = {2157-6904},
doi = {10.1145/3725857},
note = {Just Accepted},
journal = {ACM Trans. Intell. Syst. Technol.},
month = mar
}

@ARTICLE{zhao-2025,
  author={Zhao, Xinge and Cheah, Chien Chern},
  journal={IEEE Transactions on Artificial Intelligence}, 
  title={Ensuring Reliable Learning in Graph Convolutional Networks: Convergence Analysis and Training Methodology}, 
  year={2025},
  volume={},
  number={},
  pages={1-15},
  doi={10.1109/TAI.2025.3550458}}

@article{dezoort-2023,
  title={Graph neural networks at the Large Hadron Collider},
  author={DeZoort, Gage and Battaglia, Peter W and Biscarat, Catherine and Vlimant, Jean-Roch},
  journal={Nature Reviews Physics},
  volume={5},
  number={5},
  pages={281--303},
  year={2023},
  publisher={Nature Publishing Group UK London}
}

@article{zhao-2024,
  title={A review of graph neural network applications in mechanics-related domains},
  author={Zhao, Yingxue and Li, Haoran and Zhou, Haosu and Attar, Hamid Reza and Pfaff, Tobias and Li, Nan},
  journal={Artificial Intelligence Review},
  volume={57},
  number={11},
  pages={315},
  year={2024},
  publisher={Springer}
}

@article{xue-2022,
  title={Physics-embedded graph network for accelerating phase-field simulation of microstructure evolution in additive manufacturing},
  author={Xue, Tianju and Gan, Zhengtao and Liao, Shuheng and Cao, Jian},
  journal={npj Computational Materials},
  volume={8},
  number={1},
  pages={201},
  year={2022},
  publisher={Nature Publishing Group UK London}
}

@article{peng-2023,
  title={Physics-informed graph convolutional neural network for modeling fluid flow and heat convection},
  author={Peng, Jiang-Zhou and Hua, Yue and Li, Yu-Bai and Chen, Zhi-Hua and Wu, Wei-Tao and Aubry, Nadine},
  journal={Physics of Fluids},
  volume={35},
  number={8},
  year={2023},
  publisher={AIP Publishing}
}

@article{li-2024,
  title={Physics-constrained and flow-field-message-informed graph neural network for solving unsteady compressible flows},
  author={Li, Siye and Sun, Zhensheng and Zhu, Yujie and Zhang, Chi},
  journal={Physics of Fluids},
  volume={36},
  number={4},
  year={2024},
  publisher={AIP Publishing}
}

@InProceedings{choromanski-2023,
  title = 	 {Taming graph kernels with random features},
  author =       {Choromanski, Krzysztof Marcin},
  booktitle = 	 {Proceedings of the 40th International Conference on Machine Learning},
  pages = 	 {5964--5977},
  year = 	 {2023},
  editor = 	 {Krause, Andreas and Brunskill, Emma and Cho, Kyunghyun and Engelhardt, Barbara and Sabato, Sivan and Scarlett, Jonathan},
  volume = 	 {202},
  series = 	 {Proceedings of Machine Learning Research},
  month = 	 jul,
  publisher =    {PMLR},
}

@inproceedings{reid-2023,
 author = {Reid, Isaac and Choromanski, Krzysztof M and Weller, Adrian},
 booktitle = {Advances in Neural Information Processing Systems},
 editor = {A. Oh and T. Naumann and A. Globerson and K. Saenko and M. Hardt and S. Levine},
 pages = {14770--14796},
 publisher = {Curran Associates, Inc.},
 title = {Quasi-Monte Carlo Graph Random Features},
 volume = {36},
 year = {2023}
}

@misc{reid-2023a,
    title={General Graph Random Features},
    author={Isaac Reid and Krzysztof Choromanski and Eli Berger and Adrian Weller},
    year={2023},
    eprint={2310.04859},
    archivePrefix={arXiv},
    primaryClass={stat.ML},
  pubstate = {prepublished}
}

@inproceedings{gallicchio-2020,
  title={Fast and deep graph neural networks},
  author={Gallicchio, Claudio and Micheli, Alessio},
  booktitle={Proceedings of the AAAI conference on artificial intelligence},
  volume={34},
  pages={3898--3905},
  year={2020}
}

@article{thangamuthu-2022,
  title={Unravelling the performance of physics-informed graph neural networks for dynamical systems},
  author={Thangamuthu, Abishek and Kumar, Gunjan and Bishnoi, Suresh and Bhattoo, Ravinder and Krishnan, NM and Ranu, Sayan},
  journal={Advances in Neural Information Processing Systems},
  volume={35},
  pages={3691--3702},
  year={2022}
}

@inproceedings{sanchez-2020,
  title={Learning to simulate complex physics with graph networks},
  author={Sanchez-Gonzalez, Alvaro and Godwin, Jonathan and Pfaff, Tobias and Ying, Rex and Leskovec, Jure and Battaglia, Peter},
  booktitle={International conference on machine learning},
  pages={8459--8468},
  year={2020},
  organization={PMLR}
}

@inproceedings{gallicchio-2010,
  title={Graph echo state networks},
  author={Gallicchio, Claudio and Micheli, Alessio},
  booktitle={The 2010 international joint conference on neural networks (IJCNN)},
  pages={1--8},
  year={2010},
  organization={IEEE}
}

@article{wang-2023,
  title={Echo state graph neural networks with analogue random resistive memory arrays},
  author={Wang, Shaocong and Li, Yi and Wang, Dingchen and Zhang, Woyu and Chen, Xi and Dong, Danian and Wang, Songqi and Zhang, Xumeng and Lin, Peng and Gallicchio, Claudio and others},
  journal={Nature Machine Intelligence},
  volume={5},
  number={2},
  pages={104--113},
  year={2023},
  publisher={Nature Publishing Group UK London}
}

@article{liu-1989,
  title={On the limited memory BFGS method for large scale optimization},
  author={Liu, Dong C and Nocedal, Jorge},
  journal={Mathematical programming},
  volume={45},
  number={1},
  pages={503--528},
  year={1989},
  publisher={Springer}
}

@inproceedings{riedmiller-1993,
  title={A direct adaptive method for faster backpropagation learning: The RPROP algorithm},
  author={Riedmiller, Martin and Braun, Heinrich},
  booktitle={IEEE international conference on neural networks},
  pages={586--591},
  year={1993},
  organization={IEEE}
}

@misc{tieleman-2012,
  title={Rmsprop: Divide the gradient by a running average of its recent magnitude. Lecture 6.5},
  author={Tieleman, Tijmen and Hinton, Geoffrey},
  year={2012}
}

@inproceedings{sutskever-2013,
  title={On the importance of initialization and momentum in deep learning},
  author={Sutskever, Ilya and Martens, James and Dahl, George and Hinton, Geoffrey},
  booktitle={International conference on machine learning},
  pages={1139--1147},
  year={2013},
  organization={PMLR}
}

@article{duchi-2011,
  author  = {John Duchi and Elad Hazan and Yoram Singer},
  title   = {Adaptive Subgradient Methods for Online Learning and Stochastic Optimization},
  journal = {Journal of Machine Learning Research},
  year    = {2011},
  volume  = {12},
  number  = {61},
  pages   = {2121--2159},
}

@misc{loshchilov-2019,
      title={Decoupled Weight Decay Regularization}, 
      author={Ilya Loshchilov and Frank Hutter},
      year={2019},
      eprint={1711.05101},
      archivePrefix={arXiv},
      primaryClass={cs.LG},
}

@article{zeiler-2012,
  title={Adadelta: an adaptive learning rate method},
  author={Zeiler, Matthew D},
  journal={arXiv preprint arXiv:1212.5701},
  year={2012}
}

@misc{liu-2021,
      title={On the Variance of the Adaptive Learning Rate and Beyond}, 
      author={Liyuan Liu and Haoming Jiang and Pengcheng He and Weizhu Chen and Xiaodong Liu and Jianfeng Gao and Jiawei Han},
      year={2021},
      eprint={1908.03265},
      archivePrefix={arXiv},
      primaryClass={cs.LG},
  pubstate = {prepublished}
}

@article{dozat-2016,
  title={Incorporating {Nesterov} momentum into {Adam}},
  author={Dozat, Timothy},
 journal={Proceedings of the 4th International Conference on Learning Representations, Workshop Track},
month=may,
  year={2016}
}

@inproceedings{shazeer-2018,
  title={Adafactor: Adaptive learning rates with sublinear memory cost},
  author={Shazeer, Noam and Stern, Mitchell},
  booktitle={International Conference on Machine Learning},
  pages={4596--4604},
  year={2018},
  organization={PMLR}
}

@article{hairer-2003,
  title={Geometric numerical integration illustrated by the St{\"o}rmer--Verlet method},
  author={Hairer, Ernst and Lubich, Christian and Wanner, Gerhard},
  journal={Acta numerica},
  volume={12},
  pages={399--450},
  year={2003},
  publisher={Cambridge University Press}
}

@article{sgd-optimizer,
  title={A Stochastic Approximation Method},
  author={Herbert E. Robbins},
  journal={Annals of Mathematical Statistics},
  year={1951},
  volume={22},
  pages={400-407},
}

@misc{averaged-sgd-optimizer,
      title={SGD: General Analysis and Improved Rates}, 
      author={Robert Mansel Gower and Nicolas Loizou and Xun Qian and Alibek Sailanbayev and Egor Shulgin and Peter Richtarik},
      year={2019},
      eprint={1901.09401},
      archivePrefix={arXiv},
      primaryClass={cs.LG},
pubstate={prepublished}
}

@article{meng-2014-leastsquares,
  title = {{{LSRN}}: {{A}} Parallel Iterative Solver for Strongly Over- or Underdetermined Systems},
  shorttitle = {{{LSRN}}},
  author = {Meng, Xiangrui and Saunders, Michael A. and Mahoney, Michael W.},
  year = {2014},
  month = jan,
  journal = {SIAM Journal on Scientific Computing},
  volume = {36},
  number = {2},
  pages = {C95-C118},
  issn = {1064-8275, 1095-7197},
  doi = {10.1137/120866580},
  langid = {english}
}

@inproceedings{zozoulenko-2025,
title={Random Feature Representation Boosting},
author={Nikita Zozoulenko and Thomas Cass and Lukas Gonon},
booktitle={Forty-second International Conference on Machine Learning},
year={2025},
}

@article{mahoney-2014-lsrn,
  title={LSRN: A parallel iterative solver for strongly over-or underdetermined systems},
  author={Meng, Xiangrui and Saunders, Michael A and Mahoney, Michael W},
  journal={SIAM Journal on Scientific Computing},
  volume={36},
  number={2},
  pages={C95--C118},
  year={2014},
  publisher={SIAM}
}

@article{paige-1982-lsqr,
  title={LSQR: An algorithm for sparse linear equations and sparse least squares},
  author={Paige, Christopher C and Saunders, Michael A},
  journal={ACM Transactions on Mathematical Software (TOMS)},
  volume={8},
  number={1},
  pages={43--71},
  year={1982},
  publisher={ACM New York, NY, USA}
}

@article{fong-2011-lsmr,
  title={LSMR: An iterative algorithm for sparse least-squares problems},
  author={Fong, David Chin-Lung and Saunders, Michael},
  journal={SIAM Journal on Scientific Computing},
  volume={33},
  number={5},
  pages={2950--2971},
  year={2011},
  publisher={SIAM}
}

@inproceedings{huang-2004-elm,
  title={Extreme learning machine: a new learning scheme of feedforward neural networks},
  author={Huang, Guang-Bin and Zhu, Qin-Yu and Siew, Chee-Kheong},
  booktitle={2004 IEEE international joint conference on neural networks (IEEE Cat. No. 04CH37541)},
  volume={2},
  pages={985--990},
  year={2004}
}

@ARTICLE{zhang-2012-augmented-elm,
  author={Zhang, Rui and Lan, Yuan and Huang, Guang-Bin and Xu, Zong-Ben},
  journal={IEEE Transactions on Neural Networks and Learning Systems},
  title={Universal Approximation of Extreme Learning Machine With Adaptive Growth of Hidden Nodes},
  year={2012},
  volume={23},
  number={2},
  pages={365-371},
}

@ARTICLE{leung-2019-elm,
  author={Leung, Ho Chun and Leung, Chi Sing and Wong, Eric Wing Ming},
  journal={IEEE Access},
  title={Fault and Noise Tolerance in the Incremental Extreme Learning Machine},
  year={2019},
  volume={7},
  pages={155171-155183},
}

@article{huang-2006-universal-approx-elm,
  author = {Huang, Guang-Bin and Chen, Lei and Siew, Chee},
  year = {2006},
  pages = {879-92},
  title = {Universal Approximation Using Incremental Constructive Feedforward Networks With Random Hidden Nodes},
  volume = {17},
  journal = {IEEE transactions on neural networks / a publication of the IEEE Neural Networks Council},
}

@inproceedings{rahimi-2008-random-feature,
  title={Uniform approximation of functions with random bases},
  author={Rahimi, Ali and Recht, Benjamin},
  booktitle={2008 46th annual allerton conference on communication, control, and computing},
  pages={555--561},
  year={2008},
  organization={IEEE}
}

@article{rahimi-2007random-fourier-features,
  title={Random features for large-scale kernel machines},
  author={Rahimi, Ali and Recht, Benjamin},
  journal={Advances in neural information processing systems},
  volume={20},
  year={2007}
}

@article{datar-2024-swim-pde,
  title={Solving partial differential equations with sampled neural networks},
  author={Datar, Chinmay and Kapoor, Taniya and Chandra, Abhishek and Sun, Qing and Burak, Iryna and Bolager, Erik Lien and Veselovska, Anna and Fornasier, Massimo and Dietrich, Felix},
  journal={arXiv preprint arXiv:2405.20836},
  year={2024}
}

@article{bolager-2024-swim-rnn,
  title={Gradient-free training of recurrent neural networks},
  author={Bolager, Erik Lien and Cukarska, Ana and Burak, Iryna and Monfared, Zahra and Dietrich, Felix},
  journal={arXiv preprint arXiv:2410.23467},
  year={2024}
}

@inproceedings{bolager-2023-swim,
  title = {Sampling Weights of Deep Neural Networks},
  booktitle = {Advances in Neural Information Processing Systems},
  author = {Bolager, Erik L and Burak, Iryna and Datar, Chinmay and Sun, Qing and Dietrich, Felix},
  year = {2023},
  volume = {36},
  pages = {63075--63116},
  publisher = {Curran Associates, Inc.}
}

@inproceedings{rahma-2024-swim-hnn,
  title = {Training {{Hamiltonian}} Neural Networks without Backpropagation},
  booktitle = {Workshop on {{Machine Learning}} and the {{Physical Sciences}}},
  author = {Rahma, Atamert and Datar, Chinmay and Dietrich, Felix},
  year = {2024},
  month = nov,
  publisher = {NeurIPS 2024},
}

@article{fabiani-2021,
  title = {Numerical Solution and Bifurcation Analysis of Nonlinear Partial Differential Equations with Extreme Learning Machines},
  author = {Fabiani, Gianluca and Calabr{\`o}, Francesco and Russo, Lucia and Siettos, Constantinos},
  year = {2021},
  month = nov,
  journal = {Journal of Scientific Computing},
  volume = {89},
  number = {2},
  pages = {44},
  issn = {0885-7474, 1573-7691},
  doi = {10.1007/s10915-021-01650-5},
  langid = {english}
}

@misc{fabiani-2024,
  title = {Random {{Projection Neural Networks}} of {{Best Approximation}}: {{Convergence}} Theory and Practical Applications},
  shorttitle = {Random {{Projection Neural Networks}} of {{Best Approximation}}},
  author = {Fabiani, Gianluca},
  year = {2024},
  month = feb,
  number = {arXiv:2402.11397},
  eprint = {2402.11397},
  primaryclass = {cs},
  publisher = {arXiv},
  archiveprefix = {arXiv},
  pubstate = {prepublished}
}

@article{fabiani-2025,
  title = {{{RandONets}}: {{Shallow}} Networks with Random Projections for Learning Linear and Nonlinear Operators},
  shorttitle = {{{RandONets}}},
  author = {Fabiani, Gianluca and Kevrekidis, Ioannis G. and Siettos, Constantinos and Yannacopoulos, Athanasios N.},
  year = {2025},
  month = jan,
  journal = {Journal of Computational Physics},
  volume = {520},
  pages = {113433},
  issn = {00219991},
  doi = {10.1016/j.jcp.2024.113433},
  langid = {english}
}

@article{galaris-2022,
  title = {Numerical {{Bifurcation Analysis}} of {{PDEs From Lattice Boltzmann Model Simulations}}: A {{Parsimonious Machine Learning Approach}}},
  shorttitle = {Numerical {{Bifurcation Analysis}} of {{PDEs From Lattice Boltzmann Model Simulations}}},
  author = {Galaris, Evangelos and Fabiani, Gianluca and Gallos, Ioannis and Kevrekidis, Ioannis and Siettos, Constantinos},
  year = {2022},
  month = aug,
  journal = {Journal of Scientific Computing},
  volume = {92},
  number = {2},
  pages = {34},
  issn = {0885-7474, 1573-7691},
  doi = {10.1007/s10915-022-01883-y},
  langid = {english}
}

@inproceedings{schlichtkrull-2018-cond-edge-feature-gnn,
  author = {Schlichtkrull, Michael and Kipf, Thomas N. and Bloem, Peter and van den Berg, Rianne and Titov, Ivan and Welling, Max},
  title = {Modeling Relational Data with Graph Convolutional Networks},
  year = {2018},
  isbn = {978-3-319-93416-7},
  publisher = {Springer-Verlag},
  address = {Berlin, Heidelberg},
  booktitle = {The Semantic Web: 15th International Conference, ESWC 2018, Heraklion, Crete, Greece, June 3–7, 2018, Proceedings},
  pages = {593–607},
  numpages = {15},
  location = {Heraklion, Greece}
}

@InProceedings{brockschmidt-2020-cond-edge-feature-gnn,
  title = 	 {{GNN}-{F}i{LM}: Graph Neural Networks with Feature-wise Linear Modulation},
  author =       {Brockschmidt, Marc},
  booktitle = 	 {Proceedings of the 37th International Conference on Machine Learning},
  pages = 	 {1144--1152},
  year = 	 {2020},
  editor = 	 {III, Hal Daumé and Singh, Aarti},
  volume = 	 {119},
  series = 	 {Proceedings of Machine Learning Research},
  month = 	 jul,
  publisher =    {PMLR},
}

@InProceedings{gilmer-2017,
  title = 	 {Neural Message Passing for Quantum Chemistry},
  author =       {Justin Gilmer and Samuel S. Schoenholz and Patrick F. Riley and Oriol Vinyals and George E. Dahl},
  booktitle = 	 {Proceedings of the 34th International Conference on Machine Learning},
  pages = 	 {1263--1272},
  year = 	 {2017},
  editor = 	 {Precup, Doina and Teh, Yee Whye},
  volume = 	 {70},
  series = 	 {Proceedings of Machine Learning Research},
  month = 	 aug,
  publisher =    {PMLR},
}

@article{corso-2024,
  title = {Graph Neural Networks},
  author = {Corso, Gabriele and Stark, Hannes and Jegelka, Stefanie and Jaakkola, Tommi and Barzilay, Regina},
  year = {2024},
  month = mar,
  journal = {Nature Reviews Methods Primers},
  volume = {4},
  number = {1},
  pages = {1--13},
  publisher = {Nature Publishing Group},
  issn = {2662-8449},
  doi = {10.1038/s43586-024-00294-7},
  copyright = {2024 Springer Nature Limited},
  langid = {english},
}

@article{nagarajan-2023,
    title={{FASTRAIN}-{GNN}: Fast and Accurate Self-Training for Graph Neural Networks},
    author={Amrit Nagarajan and Anand Raghunathan},
    journal={Transactions on Machine Learning Research},
    issn={2835-8856},
    year={2023},
}

@inproceedings{du-2022-se3,
  title={SE(3) equivariant graph neural networks with complete local frames},
  author={Du, Weitao and Zhang, He and Du, Yuanqi and Meng, Qi and Chen, Wei and Zheng, Nanning and Shao, Bin and Liu, Tie-Yan},
  booktitle={International Conference on Machine Learning},
  pages={5583--5608},
  year={2022},
  organization={PMLR}
}

@inproceedings{pfaff-2021-learning-mesh,
  title={Learning Mesh-Based Simulation with Graph Networks},
  author={Tobias Pfaff and Meire Fortunato and Alvaro Sanchez-Gonzalez and Peter Battaglia},
  booktitle={International Conference on Learning Representations},
  year={2021},
}

@misc{sanchez-gonzalez-2019-hgnn,
  title = {Hamiltonian {{Graph Networks}} with {{ODE Integrators}}},
  author = {{Sanchez-Gonzalez}, Alvaro and Bapst, Victor and Cranmer, Kyle and Battaglia, Peter},
  year = {2019},
  month = sep,
  number = {arXiv:1909.12790},
  eprint = {1909.12790},
  primaryclass = {cs},
  publisher = {arXiv},
  doi = {10.48550/arXiv.1909.12790},
  urldate = {2025-04-28},
  archiveprefix = {arXiv},
}

@article{tierz-2025-generic-gnn,
  title = {Graph Neural Networks Informed Locally by Thermodynamics},
  author = {Tierz, Alicia and Alfaro, Ic{\'i}ar and Gonz{\'a}lez, David and Chinesta, Francisco and Cueto, El{\'i}as},
  year = {2025},
  month = mar,
  journal = {Engineering Applications of Artificial Intelligence},
  volume = {144},
  pages = {110108},
  issn = {09521976},
  doi = {10.1016/j.engappai.2025.110108},
  urldate = {2025-04-28},
}

@article{varghese-2025-sympgnn,
  title = {{{SympGNNs}}: {{Symplectic Graph Neural Networks}} for Identifying High-Dimensional {{Hamiltonian}} Systems and Node Classification},
  shorttitle = {{{SympGNNs}}},
  author = {Varghese, Alan John and Zhang, Zhen and Karniadakis, George Em},
  year = {2025},
  month = jul,
  journal = {Neural Networks},
  volume = {187},
  pages = {107397},
  issn = {08936080},
  doi = {10.1016/j.neunet.2025.107397},
  urldate = {2025-04-28},
  langid = {english}
}

@inproceedings{bhattoo-2022-lgn-ridig,
 author = {Bhattoo, Ravinder and Ranu, Sayan and Krishnan, N M Anoop},
 booktitle = {Advances in Neural Information Processing Systems},
 editor = {S. Koyejo and S. Mohamed and A. Agarwal and D. Belgrave and K. Cho and A. Oh},
 pages = {29789--29800},
 publisher = {Curran Associates, Inc.},
 title = {Learning Articulated Rigid Body Dynamics with Lagrangian Graph Neural Network},
 volume = {35},
 year = {2022}
}

@misc{sharma-2025,
      title={Dynami-CAL GraphNet: A Physics-Informed Graph Neural Network Conserving Linear and Angular Momentum for Dynamical Systems}, 
      author={Vinay Sharma and Olga Fink},
      year={2025},
      eprint={2501.07373},
      archivePrefix={arXiv},
      primaryClass={cs.LG},
  pubstate = {prepublished},
}

@article{shukla-2022, 
        title={Scalable algorithms for physics-informed neural and graph networks}, 
        volume={3}, 
        DOI={10.1017/dce.2022.24}, 
        journal={Data-Centric Engineering}, 
        author={Shukla, Khemraj and Xu, Mengjia and Trask, Nathaniel and Karniadakis, George E.}, 
        year={2022}, 
        pages={e24}
}

@inproceedings{greydanus-2019-hnn,
 author = {Greydanus, Samuel and Dzamba, Misko and Yosinski, Jason},
 booktitle = {Advances in Neural Information Processing Systems},
 editor = {H. Wallach and H. Larochelle and A. Beygelzimer and F. d\textquotesingle Alch\'{e}-Buc and E. Fox and R. Garnett},
 pages = {},
 publisher = {Curran Associates, Inc.},
 title = {Hamiltonian Neural Networks},
 volume = {32},
 year = {2019}
}

@article{bertalan-2019-hnn,
  year = 2019,
  publisher = {{AIP} Publishing},
  volume = {29},
  number = {12},
  author = {Tom Bertalan and Felix Dietrich and Igor Mezic and Ioannis G. Kevrekidis},
  title = {On learning Hamiltonian systems from data},
  journal = {Chaos: An Interdisciplinary Journal of Nonlinear Science}
}

@article{offen-2022-GP-inverse-modified-hamiltonian,
  title = {Symplectic Integration of Learned {{Hamiltonian}} Systems},
  author = {Offen, C. and {Ober-Bloebaum}, S.},
  year = {2022},
  journal = {Chaos: An Interdisciplinary Journal of Nonlinear Science},
  volume = {32},
  number = {1},
  pages = {013122},
}

@article{dierkes-2023-hnn-with-symplectic-prior,
  title={Hamiltonian neural networks with automatic symmetry detection},
  author={Dierkes, Eva and Offen, Christian and Ober-Bl{\"o}baum, Sina and Fla{\ss}kamp, Kathrin},
  journal={Chaos: An Interdisciplinary Journal of Nonlinear Science},
  volume={33},
  number={6},
  year={2023},
  publisher={AIP Publishing}
}

@article{ober-blobaum-2023,
  title = {Variational Learning of {{Euler}}--{{Lagrange}} Dynamics from Data},
  author = {{Ober-Bloebaum}, Sina and Offen, Christian},
  year = {2023},
  journal = {Journal of Computational and Applied Mathematics},
  volume = {421},
  pages = {114780},
}

@article{shaan-2021-port-hnn,
  title={Port-Hamiltonian neural networks for learning explicit time-dependent dynamical systems},
  author={Desai, Shaan A and Mattheakis, Marios and Sondak, David and Protopapas, Pavlos and Roberts, Stephen J},
  journal={Physical Review E},
  volume={104},
  number={3},
  pages={034312},
  year={2021},
  publisher={APS}
}

@misc{roth-2025-stable-port-hnn,
  title = {Stable {{Port-Hamiltonian Neural Networks}}},
  author = {Roth, Fabian J. and Klein, Dominik K. and Kannapinn, Maximilian and Peters, Jan and Weeger, Oliver},
  year = {2025},
  month = feb,
  number = {arXiv:2502.02480},
  eprint = {2502.02480},
  primaryclass = {cs},
  publisher = {arXiv},
  doi = {10.48550/arXiv.2502.02480},
  urldate = {2025-03-04},
  archiveprefix = {arXiv},
}

@inproceedings{cranmer-2019-lagrangian-nn,
  title={Lagrangian Neural Networks},
  author={Miles Cranmer and Sam Greydanus and Stephan Hoyer and Peter Battaglia and David Spergel and Shirley Ho},
  booktitle={ICLR 2020 Workshop on Integration of Deep Neural Models and Differential Equations},
  year={2019},
}

@inproceedings{lutter-2018-lagrangian-deep-nn,
  title={Deep Lagrangian Networks: Using Physics as Model Prior for Deep Learning},
  author={Michael Lutter and Christian Ritter and Jan Peters},
  booktitle={International Conference on Learning Representations},
  year={2019},
}

@article{hernandez-2021-generic-spnn,
  title={Structure-preserving neural networks},
  author={Hern{\'a}ndez, Quercus and Bad{\'\i}as, Alberto and Gonz{\'a}lez, David and Chinesta, Francisco and Cueto, El{\'\i}as},
  journal={Journal of Computational Physics},
  volume={426},
  pages={109950},
  year={2021},
  publisher={Elsevier}
}

@article{lee-2021-generic-gnode,
  title={Machine learning structure preserving brackets for forecasting irreversible processes},
  author={Lee, Kookjin and Trask, Nathaniel and Stinis, Panos},
  journal={Advances in Neural Information Processing Systems},
  volume={34},
  pages={5696--5707},
  year={2021}
}

@article{zhang-2022-generic-gfinn,
  title = {{{GFINNs}}: {{GENERIC}} Formalism Informed Neural Networks for Deterministic and Stochastic Dynamical Systems},
  shorttitle = {{{GFINNs}}},
  author = {Zhang, Zhen and Shin, Yeonjong and Em Karniadakis, George},
  year = {2022},
  month = aug,
  journal = {Philosophical Transactions of the Royal Society A: Mathematical, Physical and Engineering Sciences},
  volume = {380},
  number = {2229},
  pages = {20210207},
  issn = {1364-503X, 1471-2962},
  doi = {10.1098/rsta.2021.0207},
  urldate = {2025-04-28},
}

@inproceedings{gruber-2025-generic-nms,
title={Efficiently Parameterized Neural Metriplectic Systems},
author={Anthony Gruber and Kookjin Lee and Haksoo Lim and Noseong Park and Nathaniel Trask},
booktitle={The Thirteenth International Conference on Learning Representations},
year={2025},
}

@inproceedings{
    xiong-2021-nonseparable-hard-training-with-integrator,
    title={Nonseparable Symplectic Neural Networks},
    author={Shiying Xiong and Yunjin Tong and Xingzhe He and Shuqi Yang and Cheng Yang and Bo Zhu},
    booktitle={International Conference on Learning Representations},
    year={2021},
}

@misc{bishnoi-2023,
  title = {Discovering {{Symbolic Laws Directly}} from {{Trajectories}} with {{Hamiltonian Graph Neural Networks}}},
  author = {Bishnoi, Suresh and Bhattoo, Ravinder and Jayadeva and Ranu, Sayan and Krishnan, N. M. Anoop},
  year = 2023,
  month = jul,
  number = {arXiv:2307.05299},
  eprint = {2307.05299},
  primaryclass = {cs},
  publisher = {arXiv},
  doi = {10.48550/arXiv.2307.05299},
  urldate = {2025-11-20},
  archiveprefix = {arXiv},
}

@book{rosenblatt-1962,
  title={Perceptions and the theory of brain mechanisms},
  author={Rosenblatt, Frank},
  year={1962},
  publisher={Spartan books}
}

@article{barron-1993,
  title = {Universal Approximation Bounds for Superpositions of a Sigmoidal Function},
  author = {Barron, A.R.},
  year = 1993,
  month = may,
  journal = {IEEE Transactions on Information Theory},
  volume = {39},
  number = {3},
  pages = {930--945},
  issn = {0018-9448, 1557-9654},
  doi = {10.1109/18.256500},
  urldate = {2022-10-28},
}

@incollection{johnson-1984,
  title = {Extensions of {{Lipschitz}} Mappings into a {{Hilbert}} Space},
  booktitle = {Contemporary {{Mathematics}}},
  author = {Johnson, William B. and Lindenstrauss, Joram},
  editor = {Beals, Richard and Beck, Anatole and Bellow, Alexandra and Hajian, Arshag},
  year = 1984,
  volume = {26},
  pages = {189--206},
  publisher = {American Mathematical Society},
  address = {Providence, Rhode Island},
  doi = {10.1090/conm/026/737400},
  isbn = {978-0-8218-5030-5 978-0-8218-7611-4},
  langid = {english}
}
\bibliographystyle{iclr2026_conference}

\newpage
\appendix

\newcommand{\xmark}{%
\tikz[scale=0.23] {
    \draw[line width=0.7,line cap=round] (0,0) to [bend left=6] (1,1);
    \draw[line width=0.7,line cap=round] (0.2,0.95) to [bend right=3] (0.8,0.05);
}}

\appendix

\renewcommand{\thefigure}{\thesection.\arabic{figure}}
\renewcommand{\theequation}{\thesection.\arabic{equation}}
\renewcommand{\thealgocf}{\thesection.\arabic{algocf}}
\renewcommand{\thetable}{\thesection.\arabic{table}}

\section*{Appendix}
\section{Notation}\label{app_notation}
We define the notation we use in the main text and in the Appendix in
\Cref{tab:notation}.

\begin{table}\caption{Notation used in \Cref{sections:method} and in the appendix.}
    \begin{center}
    \begin{tabular}{c p{10cm} }
    \toprule
    \textbf{Problem setup} & \\
    \(N\) & Number of nodes/particles in the system \\
    \(d\) & Spatial dimension \\
    \(M\) & Number of training data points \\
    \(N_e\) & Number of edges in the graph\\
    \(q, p \in \R^{d \cdot N} \) & Positions and momenta of all the nodes \\
    \(q_i, p_i \in \R^{d} \) & Positions and momenta of the $i^{\text{th}}$ node \\
    \(\mathcal{H}: \R^{2d \cdot N} \rightarrow \R\)  &  True Hamiltonian function\\
    $\widehat{\mathcal{H}}: \R^{2d \cdot N} \rightarrow \R$ &  Predicted Hamiltonian function\\
    \(A\) & Symmetric adjacency matrix\\
     \midrule
    \textbf{RF-HGN setup} & \\
     \(V\)  & Set of node feature encodings\\
     \(E\) & Set of edge feature encodings\\
     \(d_V \in \N\) & Number of node features\\
     \(d_E \in \N\) & Number of edge features\\
     \(d_h \in \N \) & Latent (hidden) dimension for encoding node and edge features\\
     \(d_M \in \N\) & Latent (hidden) dimension for encoding messages\\
     \(d_L \in \N\) & Size of the input to the linear layer (here also the network width)\\
     \(\bar{q}_i,\,\bar{p}_i  \in \R^{d}\) & Invariant representation \\
    \( v_i \in \R^{d_V}\) & Node features for the $i^{\text{th}}$ node \\
     $ e_{ij} \in \R^{d_E}$ &  Edge features for the edge \((i,j)\) with \(i > j\) \\
     \(\mathcal{N}_j\) & Set of source nodes connected to destination node \(j\) \\
     \(\phi_V: \R^{d_V} \rightarrow \R^{d_h}\)& dense layer for encoding the node features\\
     \(\phi_E: \R^{d_E} \rightarrow \R^{d_h}\) & dense layer for encoding the edge features \\
     \(\phi_M: \R^{2d_h} \rightarrow \R^{d_M}\) & Message encoder \\
     \(W_V \in \R^{d_h \times d_V},  b_V \in \R^{d_h}\) & Weights and biases of the node encoder \\
     \(W_E \in \R^{d_h \times d_E},  b_E \in \R^{d_h}\) & Weights and biases of the edge encoder \\
    \(W_M \in \R^{d_M \times \R^{2d_h}}\),  \(b_M \in \R^{d_M}\) & Weights and biases of the message encoder \\
    \(W_L \in \R^{d_L}\), \(b_L \in \R\) & Weights and biases of the linear layer \\
    \( m_{j} \in  \R^{d_M}\) & aggregated incoming message of node $j$\\
    \( h_i^{V} \in \R^{d_h}\) & Node feature encoding for the $i^{\text{th}}$ node\\
    \( h_{ij}^{E} \in \R^{d_h}\) & Edge feature encoding for the edge defined by nodes $i, j$\\
    \(h^{M}_{ij} \in \R^{d_M}\) & Constructed message encoding \\
    \(h_G \in \R^{d_L}\) & Global encoding of the graph \\
    \bottomrule
    \end{tabular}
    \end{center}
    \label{tab:notation}
\end{table}

\section{Additional details on Random-Feature Hamiltonian Graph Networks}
We discuss the problem setup used in
\Cref{fig:illustration_invariances} in \Cref{app_fig2_details}, how to construct rotation-invariant representation for spatial dimension $d = 3$ in \Cref{app_rot_inv}, the algorithm for the forward pass for RF-HGN in \Cref{app_alg} and run-time and memory complexity in \Cref{app_memory}.

\subsection{Additional details for Figure 2} \label{app_fig2_details}
The initial conditions for training the networks in \Cref{fig:illustration_invariances} are generated by displacing the positions by \(dq \sim U(-0.5,+0.5)\) and with momenta \(p \sim U(-2,+2)\) from some fixed reference frame illustrated on the left of \Cref{fig:illustration_invariances}.
The system is integrated using symplectic Störmer-Verlet~\cite{hairer-2003} with \(\Delta t = 10^{-4}\) for 100 steps using the true dynamics and predictions of invariant and non-invariant variations of (SWIM) RF-HGN.

\subsection{Rotation-invariant representation for spatial dimension \(\mathbf{d=3}\)}\label{app_rot_inv}

The details on obtaining rotation-invariant representations for the spatial dimension $d=2$ are discussed in \Cref{sections:method}.
Here, we extend this approach to the spatial dimension $d = 3$, leveraging the classical Gram-Schmidt orthogonalization method.

We pick \(q_1 \in \R^{3}\) and \(q_2 \in \R^{3}\) as the reference vectors and let \(e_1 = \frac{q_1}{\lVert q_1 \rVert}\) and \(e_{2}' = \frac{q_2}{\lVert q_2 \rVert}\). \(e_1\) is the first basis vector of the new frame. If \(|e_1\tran e_2'| > \epsilon\)  (near colinear), set \(e_2' = e_1 \times e_2'\) (cross product). We then project \(u_2 = e_2' - \mathrm{proj}_{e_1}e_2'\) and scale \(e_2 = \frac{u_2}{\lVert u_2 \rVert}\) (Gram-Schmidt) to compute the second basis vector \(e_2\), where \(\mathrm{proj}_{e_1}e_2' = ({e_2'}\tran e_1) e_1 \) is the projection of \(e_2'\) onto \(e_1\).
We then construct \(e_3 = e_1 \times e_2\), where \(\times\) is the cross-product.
Finally, we define the orthonormal basis \(\mathcal{B} = \begin{bmatrix}
    e_1 & e_2 & e_3
\end{bmatrix}\).
We set \(\epsilon = 0.98\) if not specified otherwise.

One can uniquely identify the first two points, independent of node ordering or orientation, as the ones closest to the mean $\bar{q}$.
In case of ties, we select the point with the smallest angle relative to the first coordinate axis centered at $\bar{q}$.
If the ties persist, we can then select the point with the smallest angle relative to the second coordinate axis.

\subsection{Algorithm}\label{app_alg}
Here, we outline the algorithm for the forward pass of the Random-Feature Hamiltonian Graph Network (RF-HGN) using the notation introduced in \Cref{sections:method}.

\begin{algorithm}
\caption{Forward pass for RF-HGN: The parameters of all dense layers \( \phi_V, \phi_E, \phi_M\) are computed leveraging random sampling techniques and last layer parameters \(W_L\) and \(b_L\) are computed using least squares (see \Cref{sec_training}).
We denote the set of neighbors that transmit information to node \(j\)\ by \(\mathcal{N}_j\). In the following, we use a single subscript, for instance, for $v_i$, to denote that we compute $v_i$ for all values of $i \in \{1, 2, \hdots, N\}$ for brevity.
Also, we use a double subscript, for instance, for $e_{ij}$, to denote that we compute $e_{ij}$ for \(i,j \in \{1,\dots, N\}\) and $i>j$, and set $e_{ji} = e_{ij}$.
}
\label{alg:forward_pass_rfhgn}
\begin{algorithmic}
\STATE \textbf{Input: } Positions and momenta of the $N$ bodies in spatial dimension $d$ \( (p, q \in \R^{2d \cdot N}) \), adjacency matrix $A \in \R^{N \times N}$
\STATE \textbf{Output:} Approximation of Hamiltonian \( \widehat{\mathcal{H}} \in \R\)
\STATE \textbf{Parameters: } Node/edge encoder dimension $d_h\in \mathbb{N}$ and message encoder dimension $d_M \in \mathbb{N}$
\STATE \(\bar{q}_i, \bar{p}_i \in \R^{2d \cdot N}\) $\gets$ \texttt{encode\_invariances}(\(p\,,\, q\)) for each \(i \in \{1,\dots,N\}\) \hfill\algorithmiccomment{Encode translation- and rotation-invariance}
\STATE $v_i \gets \begin{bmatrix}
      \bar{q}_i & \bar{p}_i
  \end{bmatrix}\tran \in \R^{2 \cdot d}$  \hfill\algorithmiccomment{Node features}
  \STATE \( e_{ij} \gets
    \begin{bmatrix}
        (\bar{q}_i - \bar{q}_j)\tran ; \lVert \bar{q}_i - \bar{q}_j \rVert
    \end{bmatrix}\tran \in \R^{d + 1}\,\)  \hfill\algorithmiccomment{Edge features}
\STATE $h_i^{V} \gets \phi_{V}(v_i) \in \R^{d_h} \,$  \hfill\algorithmiccomment{Node encoding}
\STATE $ h_{ij}^{E} \gets \phi_{E}(e_{ij}) \in \R^{d_h} \,$ \hfill\algorithmiccomment{Edge encoding}
\STATE $h^{M}_{ij} \gets \phi_M\left(
  \begin{bmatrix}
      (h_i^V)\tran & (h^E_{ij})\tran
  \end{bmatrix}\tran\right) \in \R^{d_M}$ \hfill\algorithmiccomment{Message encoding}
\STATE \(
    m_j \gets \sum_{i \in \mathcal{N}_j}{h_{ij}^M} \in \R^{d_M} \,
\)
\hfill\algorithmiccomment{Message passing (local pooling)}
\STATE $
    h_G \gets \sum_{j = 1}^{N}{\begin{bmatrix}
        (h^V_j)\tran & (m_j)\tran
    \end{bmatrix}\tran} \in \R^{d_L},$
where $d_L := d_h + d_M$ \hfill\algorithmiccomment{Message passing (global pooling)}
\STATE $
    \widehat{\mathcal{H}} \gets W_L \cdot h_G + b_L,$
where \(W_L \in \R^{d_L}\) and \(b_L \in \R\) \hfill\algorithmiccomment{Linear layer}
\STATE \textbf{return $\widehat{\mathcal{H}}$}
\end{algorithmic}
\end{algorithm}

The forward pass discussed here is independent of how the network parameters are computed.
The training leverages random sampling, automatic differentiation to compute gradients of the Hamiltonian with respect to inputs to compute \(\nabla \mathcal{H}\) using PyTorch~\citep{pytorch}, and least squares solvers as described in \Cref{sec_training}. The network can be formulated more compactly as
\begin{equation}\label{eq:compact-rfhgn}
    \widehat{\mathcal{H}}(G) = W_L \bigg( \sum_{v_j \in V}{\phi_V(v_j) \Vert \sum_{i \in \mathcal{N_j}}{\phi_M\Big(\phi_V(v_i), \phi_E(e_{ij}) \Big)}} \bigg) + b_L,
\end{equation}
where \(\Vert\) is concatenation, given a graph \(G=(V,E)\) with a node feature set \(V\) and edge feature set \(E\) as explained in \Cref{sections:method} problem setup.

\subsection{Run-time and memory complexity of training} \label{app_memory}
We use the notation defined in \Cref{tab:notation}.

\textbf{Run-time complexity: } The bottleneck of the run-time complexity is described in \Cref{sec_training}. Encoding the translational symmetry requires a mean-shift of the particles which can be done in \(\mathcal{O}(MNd)\) because for each system first the mean value of the positions has to be computed and then the values are updated which can all be done linearly in \(M\), \(N\), and \(d\) given \(d\) positions we have to shift. For encoding rotational symmetry we have implemented Gram-Schmidt orthogonalization, which is in \(\mathcal{O}(MNd^2)\).

Also note that there are \(\frac{M(M-1)}{2}\) pairs of data points to choose from when sampling random features with SWIM. In practice, we do not consider all possible pairs, but rather subsample this set uniformly by choosing the candidate number of pairs to be \(\Big\lceil \frac{|W|}{M} \Big\rceil M\), where \(|W|\) is the number of neurons. This is much less than the theoretically possible number of pairs, and still results in a robust sampling method.

\textbf{Memory complexity: }
Memory requirements for a training set of size $M$ graphs include $\mathcal{O} (d \cdot N \cdot M)$ node features, $\mathcal{O}(d \cdot N_e \cdot M)$ for edge features, and $\mathcal{O}(N_e)$ for the sparse adjacency matrix, assuming the graph stays the same for each example in the training set. For sparsity, we assume $\mathcal{O}(1)$ number of neighbors for each node.
The three dense layers (node, edge, and message encoders incur costs of $\mathcal{O}(d_h \cdot d_V)$, $\mathcal{O}(d_h \cdot d_E)$, $\mathcal{O}(d_M \cdot d_h)$.
The linear readout layer adds a further $\mathcal{O}(d_L) = \mathcal{O}(d_h + d_M)= \mathcal{O}(d_M)$.

Unlike gradient-descent-based iterative optimization schemes, we only need to compute the gradients of the Hamiltonian $\widehat{\mathcal{H}}$ with respect to inputs, and not with respect to parameters.
For this, we additionally need to store the partial derivatives of the output with respect to the input of each dense layer for back-propagation.
This amounts to an additional cost of \(\mathcal{O}(d_L \cdot d \cdot N \cdot M)\) for the partial derivatives of the global graph value with respect to inputs.

For a fixed spatial dimension $d<4$ and network width \(d_L\), since the dominant terms depend on the dataset size $M$, the number of nodes $N$, and the number of edges in a graph $N_e$, the total memory footprint during training is $\mathcal{O}(M(N + N_e))$. If we further assume zero-shot generalization with a fixed training system, then the total memory requirement is in \(\mathcal{O}(M)\) and the geometry of the system (the number of nodes \(N\) and edges \(N_e\)) can grow independently of this training.

\section{Datasets}
\label{sec:datasets}
\Cref{tab:datasets} lists summary information of the datasets used in our experiments, which are explained in more detail in the following subsections. All the constants (masses, spring, and Lennard-Jones constants) are set to one in all the experiments, and for the chain and lattice examples we have used relative distances (all positions are given as displacements relative to the equilibrated state). More information can be found in the code repository: 
\url{https://gitlab.com/fd-research/swimhgn}.
\begin{table}
    \caption{Summary of the datasets used in our main experiments. Low and high specify the uniform distribution used to sample the dataset. In \Cref{app:datasets:optim-study}, \Cref{app:datasets:zero-shot} and \Cref{app:datasets:benchmarks} we give more details on how we generate the datasets.
    }

    \centering
    \scriptsize
    \begin{tabular}{p{3.5cm} p{2.7cm} p{2.7cm} p{1.4cm} p{1.4cm}}
         \toprule
         \textbf{Experiment} & \textbf{Train points} & \textbf{Test points} & \textbf{Low} & \textbf{High} \\
         \midrule
         \Cref{tab:results_optimizers} & \(1000 \cdot N \cdot 6\) & \(1000 \cdot N \cdot 6\) & \(-0.5\) & \(+0.5\) \\
         \Cref{tab:chain-potential-gravity} & \(3000 \cdot N \cdot 4\) & \(3000 \cdot N \cdot 4\) & \(-1.0\) & \(+1.0\)
         \\
         \midrule
         \Cref{fig:zero_shot} (\(N_x\)x\(N_y\) train) & \(1000 \cdot N_x \cdot N_y \cdot 6\) & \(1000 \cdot N_x \cdot N_y \cdot 6 \) & \(-0.5\) & \(+0.5\) \\
         \Cref{fig:zero_shot} (\(N_x\)x\(N_y\) test) & --- & \(1000 \cdot N_x \cdot N_y \cdot 6 \) & \(-0.5\) & \(+0.5\) \\
         \midrule
         \Cref{fig:results:node-scaling} and \ref{fig:results:node-scaling-integration-8-and-4096-all} (train \(N\)) & \(2000 \cdot N \cdot 4\) & \(2000 \cdot N \cdot 4\) & \(-1.0\) & \(+1.0\) \\
         \Cref{fig:results:node-scaling} (test \(N\)) & ---  & \(2000 \cdot N \cdot 4\) & \(-1.0\) & \(+1.0\) \\
         \midrule
         \Cref{tab:9-lj-swim}, \ref{tab:9-lj-elm}, \ref{tab:9-lj-adam}; \Cref{fig:results:9particles-lj-snaps} and \ref{fig:results:9-lj-traj} & \(810 \cdot N \cdot 4\) & \(90 \cdot N \cdot 4\) & \(-0.1\) & \(+0.1\) \\
         \Cref{fig:results:32train-64test-lj-snaps} and \ref{fig:results:64-lj-traj} & \(540 \cdot N \cdot 4\) & \(60 \cdot N \cdot 4\) & \(-0.1\) & \(+0.1\) \\
         \midrule
         \Cref{sec:benchmarking} & \( 10000  \cdot N \cdot 4 \) & \( 100 \cdot N \cdot 4 \) & --- & --- \\
         \bottomrule
    \end{tabular}
    \label{tab:datasets}
\end{table}

\subsection{Benchmarking against SOTA optimizers}\label{app:datasets:optim-study}
The target Hamiltonian is
\begin{equation}\label{eq:lattice}
    \begin{split}
    \mathcal{H}_1(q,p) &= \frac{1}{2}\bigg( \sum_{i=1}^{Nx}\sum_{j=1}^{Ny}{\frac{\lVert p_{ij} \rVert^2}{\alpha_{ij}}} \\
    &+ \sum_{i=1}^{Nx}\sum_{j=1}^{Ny-1}{\beta_{ij}^x \lVert q_{i, j+1} - q_{ij} \rVert^2} + \sum_{j=1}^{Ny}\sum_{i=1}^{Nx-1}{\beta_{ij}^y \lVert q_{i+1,j} - q_{ij} \rVert^2} \bigg),
    \end{split}
\end{equation}
where \(q_{ij}, p_{ij} \in \R^3\), and \(\alpha_{ij}, \beta_{ij} \in \R \) denote masses and spring constants, respectively. All constants are equal to one if not specified otherwise. \(N_x\) and \(N_y\) are set to three to build a 3x3 lattice structure (with number of total nodes \(N=9\)), which moves in 3D (\(d=3\)). We generate a synthetic dataset of \(2000\) structures (graphs) with their true time derivatives \(\{ q_i, p_i, \dot{q}_i, \dot{p}_i \}_{i=1}^{2000}\) where \(q_i,p_i,\dot{q}_i,\dot{p}_i \in \R^{d \cdot N} \,\, \forall i\). We first set all \(q_i, p_i\) to be in the equilibrium state. Then we sample the displacements \(dq_i\) and \(dp_i\) from the uniform distribution \(U(-0.5, +0.5)\), and compute \(q_i \gets q_i + dq_i\) and \(p_i \gets p_i + dp_i\). We then compute the ground truths \(\dot{q}_i, \dot{p}_i\) using \Cref{eq:hamiltons-pde} and the ground truth gradient \(\nabla \mathcal{H}_1\). We shuffle and split the dataset into train (1000) and test (1000) sets. All the errors reported in \Cref{tab:results_optimizers} are the average test errors of three independent runs using different seeds. The total number of training and test points then becomes \(1000 \cdot N \cdot d \cdot 2 = 54000\) each.

Additional to the standard spring potential \(V(r)=\frac{1}{2}\beta r^2\) given distance \(r\) with spring constant \(\beta=1\), we use an anrharmonic spring potential \(V(r)=\frac{1}{2}\beta r^2 + \frac{1}{4}\eta r^4\) with nonlinearity coefficient \(\eta=1\) and the Morse potential \(V(r)=D(1 -\exp(-ar))^2\) \citep{morse-potential} with well-depth \(D=1\) and potential-width \(a=1.0\) for the 2D chain potential experiments in \Cref{tab:chain-potential-gravity}, (also in \Cref{fig:spring-chain-traj}, \ref{fig:spring-chain-snaps}, \ref{fig:anharmonic-chain-traj}, \ref{fig:anharmonic-chain-snaps}, \ref{fig:morse-chain-traj} and \ref{fig:morse-chain-snaps}). The data generation follows the same procedure explained above, with \(N=5\), resulting in a total of \(3000 \cdot N \cdot d \cdot 2 = 60000\) train and test points each.

\subsection{Zero-shot generalization and comparison of random feature methods}\label{app:datasets:zero-shot}
The experiment in \Cref{fig:zero_shot} uses the same procedure explained in \Cref{app:datasets:optim-study}, with \(N_x\) and \(N_y\) set to two, three, and four to build 2x2, 3x3, and 4x4 lattice structures.

For the experiment in \Cref{fig:results:node-scaling}, the procedure is again similar, but the structure of the experiment and data is different (an open chain). The target function for the open chain system is given in \Cref{section:results:zero-shot} as
\begin{equation}
\label{eq:chain-hamiltonian}
    \mathcal{H}_2(q,p) = \frac{1}{2} \big( \sum_{i=1}^{N}{\frac{\rVert p_i \lVert^2}{\alpha_i}} + \sum_{i=1}^{N-1}{\beta_{i} \lVert q_{i+1} - q_i \rVert^2} \big),
\end{equation}
where \(q_i,p_i \in \R^{2}\), \(\alpha_i, \beta_i \in \R\) are positions, momenta, masses, and spring constants in the system, respectively, for \(i\in\left\lbrace 1,\dots,N\right\rbrace\). All constants are equal to one if not specified otherwise. \(\mathcal{H}_2\). \(N\) is scaled from exponentially from \(2^1\) to \(2^{12}\) in the experiment, which always moves in 2D (\(d=2\)). For each \(N\), we again generate a synthetic dataset of 4000 structures (graphs) with their true time derivatives \(\{ q_i, p_i, \dot{q}_i, \dot{p}_i \}_{i=1}^{4000}\) where \(q_i, p_i, \dot{q}_i, \dot{p}_i \,\, \forall i\). We first set all \(q_i, p_i\) to be in the equilibrium state. Then we sample the displacements \(dq_i\) and \(dp_i\) from the uniform distribution \(U(-1.0, +1.0)\), and compute \(q_i \gets q_i + dq_i\) and \(p_i \gets p_i + dp_i\). We then compute the ground truths \(\dot{q}_i, \dot{p}_i\) using \Cref{eq:hamiltons-pde} and the ground truth gradient \(\nabla \mathcal{H}_2\). We shuffle and split the dataset into train (2000) and test (2000) sets.

For the molecular dynamics scenarios with the Lennard-Jones (LJ) potential the Hamiltonian is defined as
\begin{equation}
    \label{eq:lennard-jones-hamiltonian}
    \mathcal{H}_3(q,p) = \frac{1}{2}\sum_{i=1}^{N}{\frac{\lVert p_i \rVert}{\alpha_i}} + \sum_{i=1}^{N}\sum_{j=i+1}^{N}{V^{\texttt{LJ}}(\lVert q_j - q_i \rVert)},
\end{equation}
where
\[
    V^{\texttt{LJ}}(r_{ij}) = 4 \epsilon \bigg[ \Big( \frac{\sigma}{r_{ij}} \Big)^{12} - \Big( \frac{\sigma}{r_{ij}} \Big)^{6} \bigg],
\]
and where \(r_{ij} = \lVert q_j - q_i \rVert\). We set the parameters \(\alpha_i\), \(\epsilon\), \(\sigma\) to \(1.0\) and the cutoff to \(2.0\) when computing the dynamic edge indices (\Cref{fig:results:32train-64test-lj-snaps}, \ref{fig:results:64-lj-traj}), and static edge indices when training with 9 particles and testing with 9 particles (\Cref{fig:results:9particles-lj-snaps}, \ref{fig:results:9-lj-traj}, \Cref{tab:9-lj-swim}, \ref{tab:9-lj-elm}, \ref{tab:9-lj-adam}). For the static edge experiment, we generated \(300\) trajectories with \(9\) particles, and for the dynamic edge experiment, we generated \(200\) trajectories with \(36\) particles with the \(q\) displacement specified in \Cref{tab:datasets} from the equilibrium state with momenta set to zero. Each trajectory is simulated for \(50\) time steps with \update{\(\Delta t =5 \cdot 10^{-3}\)} and snapshots are taken every \(20th\) step with a train-test ratio of \(0.9\).

\subsection{Benchmarking against SOTA architectures}
\label{app:datasets:benchmarks}

To benchmark our model, we considered the N-body spring system from \citet{thangamuthu-2022}, for which details are available in the original work. Nonetheless, we mention the key properties of the dataset for completeness.

A system of N bodies with equal masses, connected by elastic springs such that each body has two connections and the system forms a closed loop. The system's physical behavior additionally depends on the spring's stiffness and its undeformed length, both are set to one. Initial positions $q_0$ are sampled as $q_0 \sim U(0,2)$ and initial momenta ${p_0}$ are sampled as ${p_0} \sim U(0,0.1)$ and subsequently mean-centered. The symplectic Störmer-Verlet~\citep{hairer-2003} integrator with a timestep of $10^{-3}$ is used to generate 100000 datapoints, which are subsampled to 100 datapoints. The approach is repeated for 100 trajectories to obtain a dataset that is split in a 75:25 ratio for a training and validation set. Unlike the original work, the test data we use consists of only one trajectory because with 100 trajectories, we were often experiencing failed simulations with the existing Adam-trained benchmarks (in particular with the LGN architecture), which significantly hinders comparison with our method.

\section{Training and zero-shot integration setup}
\label{sec:training_setup}

\subsection{Benchmarking against SOTA optimizers}
\begin{table}
    \caption{Model parameters (see \Cref{fig:swim-hgnn-architecture}) used in \Cref{section:results:optim-study}.}
    \centering
    \scriptsize
    \begin{tabular}{lcccc}
         \toprule
         \textbf{Model} & \textbf{Encoder width} (\(d_h\)) & \textbf{Network width} (\(d_L\)) & \textbf{Activation} & \textbf{Precision} \\
         \midrule
         HGN (\cref{fig:swim-hgnn-architecture}) & \(48\) & \(384\) & \texttt{softplus} & \texttt{single} \\
         \bottomrule
    \end{tabular}
    \label{tab:appendix:optim-study-model-params}
\end{table}
\begin{table}
    \caption{Hyperparameters used in \Cref{section:results:optim-study} and \Cref{section:results:zero-shot} are listed for SWIM. \texttt{Driver} and \texttt{rcond} (\(l^2\) reg.) are the parameters of \texttt{torch.linalg.lstsq}~\citep{pytorch}. Resample duplicates specifies to resample till we get a unique pair of points in the SWIM algorithm \citep{bolager-2023-swim}.}
    \centering
    \scriptsize
    \begin{tabular}{lcccc}
         \toprule
         \textbf{Optimizer} & \textbf{Driver} & \textbf{Parameter sampler} & \textbf{Resample duplicates} & \textbf{\(l^2\) reg.} \\
         \midrule
         SWIM~\cite{bolager-2023-swim} & \texttt{GELS} & \texttt{relu} & \texttt{True} &  1e-6 \\
         \bottomrule
    \end{tabular}
    \label{tab:appendix:optim-study-swim-hyperparams}
\end{table}
\begin{table}
    \caption{Hyperparameters used in \Cref{section:results:optim-study} and \Cref{section:results:zero-shot} are listed for SOTA optimizers. SGD(+m) represents both SGD~\cite{sgd-optimizer} and SGD+momentum~\citep{sutskever-2013}. The \texttt{momentum} parameter is set to \(0.9\). Avg. SGD specified the Averaged SGD~\citep{averaged-sgd-optimizer}. Default values are given in the first row (Defaults) for all the optimizers not present in this table, but are listed in \Cref{tab:results_optimizers}. \#steps is the number of total iterations (one iteration per batch). If LR schedule is specified, exponential decay is used as the learning rate scheduler. All optimizers use the \texttt{kaiming\_normal}~\citep{pytorch} weight initialization. \(l^2\) regularization (\(l^2\) reg.) is specified using the \texttt{weight\_decay} parameter~\citep{pytorch}. Full batch size is \(1000\).}
    \centering
    \scriptsize
    \begin{tabular}{lcccccc}
         \toprule
         \textbf{Optimizer} & \textbf{\#steps} & \textbf{Batch size} & \textbf{LR schedule} & \textbf{LR (start, end)} & \textbf{\(l^2\) reg.} \\
         \midrule
         Defaults & \(10000\) & \(256\) & Yes & 1e-2, 5e-5 & 1e-6 \\
         SGD(+m.)
         & \(10000\) & \(256\) & Yes & 5e-4, 5e-5 &  1e-6 \\
         Avg. SGD & \(10000\) & \(256\) & Yes & 5e-4, 5e-5 & 0 \\
         Adadelta
         & \(10000\) & \(256\) & No & 1e-1 (fixed) & 1e-6 \\
         Rprop
         & \(2560\) & Full & No & 1e-2 (fixed) & 0 \\
         RMSprop
         & \(10000\) & \(256\) & Yes & 1e-2, 5e-5 & 0 \\
         LBFGS
         & \(100\) & Full & No & 1e-1 (fixed) & 0 \\
         \bottomrule
    \end{tabular}
    \label{tab:appendix:optim-study-sota-hyperparams}
\end{table}
\Cref{tab:appendix:optim-study-model-params} and \Cref{tab:appendix:optim-study-swim-hyperparams} list the model and SWIM hyperparameters, respectively. \Cref{tab:appendix:optim-study-sota-hyperparams} lists the hyperparameters used for the SOTA optimizers listed in \Cref{tab:results_optimizers}. All the optimizers are run once with the default settings that are optimized initially for the Adam optimizer~\citep{adam-optimizer}, and tuned further with multiple iterations. Note that we only want to give a \textbf{``time to solution''}, with similar accuracies in order to compare the SOTA optimizers against our method, since the iterative routines can be done arbitrarily long and may be tuned further to reach lower approximation errors than our method with excessive hyperparameter tuning and larger number of iterations for each optimizer--at the cost of even longer training times. \textbf{The SGD family} (SGD~\citep{sgd-optimizer}, SGD+momentum~\citep{sutskever-2013}, and Averaged SGD~\citep{averaged-sgd-optimizer}) required lower learning rate starts than the adaptive-gradient based optimizers, otherwise they led to \texttt{NaN} (Not a Number) results. Even with a very low learning rate, starting at 5e-4, they all produced one \texttt{NaN} value out of three experiments, which shows their instability and difficulty in setup. In our results, we therefore only average the two valid results of the SGD-family. \textbf{In Averaged SGD}~\citep{averaged-sgd-optimizer}, the averaging may have acted as an implicit regularizer, and required no weight decay to perform similarly. Also, regularization was not necessary for \textbf{Rprop}~\citep{riedmiller-1993}, \textbf{LBFGS}~\citep{liu-1989}, and \textbf{RMSprop}~\citep{tieleman-2012}. \textbf{Adadelta}~\citep{zeiler-2012} is an adaptive method that dynamically scales updates; therefore, it does not require any scheduler. Also, \textbf{Rprop}~\citep{riedmiller-1993} uses the sign of the gradients and adapts the step size dynamically, which makes it suitable to be used with large batch updates and no scheduler. Since it uses full batch updates (with batch size of \(1000\)), its number of gradient steps is reduced to provide around the same epoch as the other optimizers. \textbf{LBFGS} is a second-order method and outperformed the other optimizers with only \(100\) steps using full batch updates and no learning rate scheduler.

\Cref{tab:appendix:chain-potential-model-params}, \ref{tab:appendix:chain-potential-adam-hyperparams}, \ref{tab:appendix:chain-potential-swim-hyperparams}, and \ref{tab:appendix:chain-potential-adam-hyperparams} list model hyperparameters for the chain potential experiment in \Cref{tab:chain-potential-gravity} for training. \Cref{tab:appendix:zero-shot-integration-params-for-chain-potential} lists the integration hyperparameters used in the same experiment when zero-shot testing. Note that during testing, we apply a constant gravitational force \([0, -0.075]\tran\) to every node in the negative y-axis direction.

\begin{table}
    \caption{Parameters used in \Cref{tab:chain-potential-gravity} (also in \Cref{fig:spring-chain-traj}, \ref{fig:spring-chain-snaps}, \ref{fig:anharmonic-chain-traj}, \ref{fig:anharmonic-chain-snaps}, \ref{fig:morse-chain-traj}, \ref{fig:morse-chain-snaps}). \#steps is the total number of time steps, \(\Delta t\) is the time step size.}
    \centering
    \scriptsize
    \begin{tabular}{cc}
         \toprule
         \textbf{Parameter} & \textbf{Value} \\
         \midrule
         Symplectic solver & Störmer-Verlet~\citep{hairer-2003} \\ \#steps & \(1e4\) \\
         \(\Delta t\) & 1e-2 \\
         \bottomrule
    \end{tabular}
    \label{tab:appendix:zero-shot-integration-params-for-chain-potential}
\end{table}
\begin{table}
    \caption{
    Model parameters (see \Cref{fig:swim-hgnn-architecture}) used in \Cref{tab:chain-potential-gravity}.
    }
    \centering
    \scriptsize
    \begin{tabular}{lcccc}
         \toprule
         \textbf{Model} & \textbf{Encoder width} (\(d_h\)) & \textbf{Network width} (\(d_L\)) & \textbf{Activation} & \textbf{Precision} \\
         \midrule
         HGN (\cref{fig:swim-hgnn-architecture}) & \(64\) & \(1024\) & \texttt{softplus} & \texttt{double} \\
         \bottomrule
    \end{tabular}
    \label{tab:appendix:chain-potential-model-params}
\end{table}
\begin{table}
    \caption{
    Hyperparameters used in \Cref{tab:chain-potential-gravity} are listed for SWIM. \texttt{Driver} and \texttt{rcond} (\(l^2\) reg.) are the parameters of \texttt{torch.linalg.lstsq}~\citep{pytorch}. Resample duplicates specifies to resample till we get a unique pair of points in the SWIM algorithm \citep{bolager-2023-swim}.
}
    \centering
    \scriptsize
    \begin{tabular}{lcccc}
         \toprule
         \textbf{Optimizer} & \textbf{Driver} & \textbf{Parameter sampler} & \textbf{Resample duplicates} & \textbf{\(l^2\) reg.} \\
         \midrule
         SWIM~\cite{bolager-2023-swim} & \texttt{GELS} & \texttt{relu} & \texttt{True} & 1e-10 \\
         \bottomrule
    \end{tabular}
    \label{tab:appendix:chain-potential-swim-hyperparams}
\end{table}
\begin{table}
    \caption{
Hyperparameters used in  \Cref{tab:chain-potential-gravity} are listed for ELM~\citep{huang-2004-elm}. \texttt{Driver} and \texttt{rcond} (\(l^2\) reg.) are the parameters of \texttt{torch.linalg.lstsq}~\citep{pytorch}. Bias low and high specify the uniform distribution of low and high values, from which the biases of the random feature layers are sampled. The weights are sampled using the standard normal distribution as explained in \Cref{sec_training}.
}
    \centering
    \scriptsize
    \begin{tabular}{lcccc}
         \toprule
         \textbf{Optimizer} & \textbf{Driver} & \textbf{Bias low} & \textbf{Bias high} & \textbf{\(l^2\) reg.} \\
         \midrule
         ELM~\citep{huang-2004-elm} & \texttt{GELS} & -1.0 & +1.0 & 1e-10 \\
         \bottomrule
    \end{tabular}
    \label{tab:appendix:chain-potential-elm-hyperparams}
\end{table}

\begin{table}
    \caption{
    Hyperparameters used in \Cref{tab:chain-potential-gravity} are listed for Adam.
}
    \centering
    \scriptsize
    \begin{tabular}{lcccccccc}
         \toprule
         \textbf{Optimizer} & \textbf{\#steps} & \textbf{Batch size} & \textbf{LR schedule} & \textbf{LR (start, end)} & \textbf{\(l^2\) reg.} & \textbf{Initialization} & \textbf{Patience} \\
         \midrule
         Adam & \(10000\) & \(256\) & exponential decay & 1e-2, 5e-5 & 1e-6 & Kaiming normal & 1000 \\
         \bottomrule
    \end{tabular}
    \label{tab:appendix:chain-potential-adam-hyperparams}
\end{table}

\subsection{Zero-shot generalization and comparison of random feature methods}
\Cref{tab:appendix:zero-shot-model-params}, \Cref{tab:appendix:optim-study-swim-hyperparams}, and \Cref{tab:appendix:zero-shot-elm-hyperparams} list the model, SWIM, and ELM hyperparameters used for the experiments in \Cref{section:results:zero-shot}, respectively. For the zero-shot evaluation presented in \Cref{fig:zero_shot}, we have trained (SWIM) RF-HGN ten times with different random seeds (also see \Cref{fig:results:adam-comparison}, \Cref{fig:errors_times}, and \Cref{tab:times_table_adam_comparison}), and used the pretrained (SWIM) RF-HGN model with the median test error to evaluate on the zero-shot test cases in order to avoid any statistical bias, as this is a random feature method. \Cref{tab:appendix:zero-shot-integration-params} lists the parameters used to integrate the system in \Cref{fig:results:node-scaling-integration-8-and-4096-all} and \Cref{fig:results:traj_2D_chain_all_bothcorners}.
\begin{table}
    \caption{Model parameters used in \Cref{section:results:zero-shot}. The network width specifies the size of the input to the last linear layer in both RF-HNN~\citep{bertalan-2019-hnn, greydanus-2019-hnn} and RF-HGN.}
    \centering
    \scriptsize
    \begin{tabular}{lcccc}
         \toprule
         \textbf{Model} & \textbf{Encoder width} (\(d_h\)) & \textbf{Network width} & \textbf{Activation} & \textbf{Precision} \\
         \midrule
         RF-HGN (\cref{fig:swim-hgnn-architecture}) & \(64\) & \(512\) & \texttt{softplus} & \texttt{single} \\
         RF-HNN & --- & \(512\) & \texttt{softplus} & \texttt{single} \\
         \bottomrule
    \end{tabular}
    \label{tab:appendix:zero-shot-model-params}
\end{table}
\begin{table}
    \caption{Hyperparameters used in  \Cref{section:results:zero-shot} are listed for ELM~\citep{huang-2004-elm}. \texttt{Driver} and \texttt{rcond} (\(l^2\) reg.) are the parameters of \texttt{torch.linalg.lstsq}~\citep{pytorch}. Bias low and high specify the uniform distribution of low and high values, from which the biases of the random feature layers are sampled. The weights are sampled using the standard normal distribution as explained in \Cref{sec_training}.}
    \centering
    \scriptsize
    \begin{tabular}{lcccc}
         \toprule
         \textbf{Optimizer} & \textbf{Driver} & \textbf{Bias low} & \textbf{Bias high} & \textbf{\(l^2\) reg.} \\
         \midrule
         ELM~\citep{huang-2004-elm} & \texttt{GELS} & -1.0 & +1.0 &  1e-6 \\
         \bottomrule
    \end{tabular}
    \label{tab:appendix:zero-shot-elm-hyperparams}
\end{table}
\begin{table}
    \caption{Parameters used in \Cref{fig:results:node-scaling-integration-8-and-4096-all} (and consequently \Cref{fig:results:traj_2D_chain_all_bothcorners}). \#steps is the total number of time steps, \(\Delta t\) is the time step size.}
    \centering
    \scriptsize
    \begin{tabular}{cc}
         \toprule
         \textbf{Parameter} & \textbf{Value} \\
         \midrule
         Symplectic solver & Störmer-Verlet~\citep{hairer-2003} \\ \#steps & \(5000\) \\
         \(\Delta t\) & 1e-3 \\
         \bottomrule
    \end{tabular}
    \label{tab:appendix:zero-shot-integration-params}
\end{table}

\begin{table}
    \caption{Parameters used in \Cref{fig:results:9particles-lj-snaps} and \Cref{fig:results:32train-64test-lj-snaps} (and consequently \Cref{tab:9-lj-swim}, \ref{tab:9-lj-elm}, \ref{tab:9-lj-adam}, \Cref{fig:results:9-lj-traj} \ref{fig:results:64-lj-traj}). \#steps is the total number of time steps, \(\Delta t\) is the time step size.}
    \centering
    \scriptsize
    \begin{tabular}{cc}
         \toprule
         \textbf{Parameter} & \textbf{Value} \\
         \midrule
         Symplectic solver & Störmer-Verlet~\citep{hairer-2003} \\ \#steps & \(1e5\) \\
         \(\Delta t\) & 1e-5 \\
         \bottomrule
    \end{tabular}
    \label{tab:appendix:zero-shot-integration-params-for-md}
\end{table}

For the molecular systems, \Cref{tab:appendix:md-model-params} lists the model parameters, \Cref{tab:appendix:md-swim-hyperparams} lists SWIM hyperparameters, \Cref{tab:appendix:md-elm-hyperparams} lists ELM hyperparameters, and \Cref{tab:appendix:md-adam-hyperparams} lists Adam hyperparameters. No early stopping was triggered in these experiments. \Cref{tab:appendix:zero-shot-integration-params-for-md} lists the parameters used to integrate all the molecular systems presented in this paper.
\begin{table}
    \caption{Model parameters used in \Cref{tab:9-lj-swim}, \ref{tab:9-lj-elm}, \ref{tab:9-lj-adam}, \Cref{fig:results:9particles-lj-snaps}, \ref{fig:results:9-lj-traj} (top row) and in \Cref{fig:results:32train-64test-lj-snaps}, \ref{fig:results:64-lj-traj} (bottom row) are listed for the RF-HGNs and Adam-HGN.}
    \centering
    \scriptsize
    \begin{tabular}{lcccc}
         \toprule
         \textbf{Model} & \textbf{Encoder width} (\(d_h\)) & \textbf{Network width} (\(d_L\)) & \textbf{Activation} & \textbf{Precision} \\
         \midrule
         HGN & \(40\) & \(800\) & \texttt{softplus} & \texttt{single} \\
         HGN & \(32\) & \(256\) & \texttt{gelu} & \texttt{single} \\
         \bottomrule
    \end{tabular}
    \label{tab:appendix:md-model-params}
\end{table}
\begin{table}
    \caption{Hyperparameters used in \Cref{tab:9-lj-swim}, \ref{tab:9-lj-elm}, \ref{tab:9-lj-adam}, \Cref{fig:results:9particles-lj-snaps}, \ref{fig:results:9-lj-traj} and in \Cref{fig:results:32train-64test-lj-snaps}, \ref{fig:results:64-lj-traj} are listed for SWIM. \texttt{Driver} and \texttt{rcond} (\(l^2\) reg.) are the parameters of \texttt{torch.linalg.lstsq}~\citep{pytorch}. Resample duplicates specifies to resample till we get a unique pair of points in the SWIM algorithm \citep{bolager-2023-swim}.}
    \centering
    \scriptsize
    \begin{tabular}{lcccc}
         \toprule
         \textbf{Optimizer} & \textbf{Driver} & \textbf{Parameter sampler} & \textbf{Resample duplicates} & \textbf{\(l^2\) reg.} \\
         \midrule
         SWIM~\cite{bolager-2023-swim} & \texttt{GELSD} & \texttt{relu} & \texttt{True} &  1e-10 \\
         \bottomrule
    \end{tabular}
    \label{tab:appendix:md-swim-hyperparams}
\end{table}
\begin{table}
    \caption{Hyperparameters used in \Cref{tab:9-lj-swim}, \ref{tab:9-lj-elm}, \ref{tab:9-lj-adam}, \Cref{fig:results:9particles-lj-snaps}, \ref{fig:results:9-lj-traj} and in \Cref{fig:results:32train-64test-lj-snaps}, \ref{fig:results:64-lj-traj} are listed for ELM~\citep{huang-2004-elm}. \texttt{Driver} and \texttt{rcond} (\(l^2\) reg.) are the parameters of \texttt{torch.linalg.lstsq}~\citep{pytorch}. Bias low and high specify the uniform distribution of low and high values, from which the biases of the random feature layers are sampled. The weights are sampled using the standard normal distribution as explained in \Cref{sec_training}.}
    \centering
    \scriptsize
    \begin{tabular}{lcccc}
         \toprule
         \textbf{Optimizer} & \textbf{Driver} & \textbf{Bias low} & \textbf{Bias high} & \textbf{\(l^2\) reg.} \\
         \midrule
         ELM~\citep{huang-2004-elm} & \texttt{GELSD} & -1.0 & +1.0 &  1e-10 \\
         \bottomrule
    \end{tabular}
    \label{tab:appendix:md-elm-hyperparams}
\end{table}
\begin{table}
    \caption{Hyperparameters used in \Cref{tab:9-lj-swim}, \ref{tab:9-lj-elm}, \ref{tab:9-lj-adam}, \Cref{fig:results:9particles-lj-snaps}, \ref{fig:results:9-lj-traj} and in \Cref{fig:results:32train-64test-lj-snaps}, \ref{fig:results:64-lj-traj} are listed for Adam.}
    \centering
    \scriptsize
    \begin{tabular}{lcccccccc}
         \toprule
         \textbf{Optimizer} & \textbf{\#steps} & \textbf{Batch size} & \textbf{LR schedule} & \textbf{LR (start, end)} & \textbf{\(l^2\) reg.} & \textbf{Initialization} & \textbf{Patience} \\
         \midrule
         Adam & \(10000\) & \(8\) & exponential decay & 1e-3, 5e-5 & 1e-10 & Kaiming normal & 500 \\
         \bottomrule
    \end{tabular}
    \label{tab:appendix:md-adam-hyperparams}
\end{table}

\subsection{Benchmarking against SOTA architectures}

\Cref{tab:appendix:benchmark-model-params} lists the model parameters, \Cref{tab:appendix:benchmark-swim-hyperparams} lists SWIM hyperparameters, \Cref{tab:appendix:benchmark-architecture-hyperparams} lists SOTA architecture hyperparameters used in \Cref{sec:benchmarking}.

\begin{table}
    \caption{Random feature Hamiltonian graph network (RF-HGN) parameters used in \Cref{sec:benchmarking}. The network width specifies the size of the input to the last linear layer in the RF-HGN.}
    \centering
    \scriptsize
    \begin{tabular}{lcccc}
         \toprule
         \textbf{Model} & \textbf{Encoder width} (\(d_h\)) & \textbf{Network width} & \textbf{Activation} & \textbf{Precision} \\
         \midrule
         RF-HGN (\cref{fig:swim-hgnn-architecture}) & \(32\) & \(512\) & \texttt{softplus} & \texttt{double} \\
         \bottomrule
    \end{tabular}
    \label{tab:appendix:benchmark-model-params}
\end{table}
\begin{table}
    \caption{Hyperparameters used in \Cref{sec:benchmarking} are listed for SWIM~\citep{bolager-2023-swim}. \texttt{Driver} and \texttt{rcond} (\(l^2\) reg.) are the parameters of \texttt{torch.linalg.lstsq}~\citep{pytorch}.}
    \centering
    \scriptsize
    \begin{tabular}{lcccc}
         \toprule
         \textbf{Optimizer} & \textbf{Driver} & \textbf{Parameter sampler} & \textbf{Resample duplicates} & \textbf{\(l^2\) reg.} \\
         \midrule
         SWIM~\citep{bolager-2023-swim} & \texttt{GELS} & \texttt{relu} & \texttt{True} & 1e-15 \\
         \bottomrule
    \end{tabular}
    \label{tab:appendix:benchmark-swim-hyperparams}
\end{table}

\begin{table}
    \caption{
    Hyperparameters used in \Cref{sec:benchmarking} are listed.
    }
    \centering
    \scriptsize
    \begin{tabular}{lcc}
         \toprule
         \textbf{Model} &\textbf{FGNN, FGNODE, LGN, HGN} & \textbf{GNODE, LGNN, HGNN}  \\
         \midrule
         Node embedding dim. & 8 & 5 \\
         Edge embedding dim. & 8 & 5 \\
         \# hidden layers & 2 & 2 \\
         \# hidden neurons (per layer) &  16 & 5 \\
         \# message passing layers & 1 & 1 \\
         Activation & \texttt{squareplus} & \texttt{squareplus} \\
         Optimizer & Adam & Adam \\
         Learning rate & $10^{-3}$ & $10^{-3}$ \\
         Batch size & 100 & 100 \\
         Epochs & 10000 & 10000 \\
         Precision & \texttt{double} & \texttt{double} \\
         \(l^2\) regularization & --- & --- \\
         \bottomrule
    \end{tabular}
    \label{tab:appendix:benchmark-architecture-hyperparams}
\end{table}

\section{Hardware}\label{sec:hardware}
\Cref{tab:hardware} lists hardware used for all the experiments presented in \Cref{section:results}. The experiments presented in \cref{section:results:optim-study}, \Cref{fig:zero_shot} (training of the 2x2 and 3x3 lattice systems), and \Cref{sec:benchmarking} are conducted on a CUDA GPU. \Cref{fig:zero_shot} (training of the 4x4 lattice), \Cref{fig:zero_shot} (testing), and \Cref{fig:results:node-scaling} are conducted on CPUs because of their larger memory requirements.
\begin{table}
    \caption{Hardware used for the experiments is listed with details on CPUs (Intel i7 and AMD EPYC), memory, GPU (NVIDIA), CUDA version (driver version, CUDA version), and operating system (OS) versions of Ubuntu LTS, together with memory requirements (in GB).}
    \centering
    \scriptsize
    \begin{tabular}{lccccc}
         \toprule
         \textbf{Experiment} & \textbf{CPU} (cores) & \textbf{Memory} & \textbf{GPU} (vram) & \textbf{CUDA} & \textbf{OS} \\
         \midrule
         \cref{section:results:optim-study} & i7-14700K (20) & 66 & RTX 4070 (12) & 550.120, 12.4 & 24.04.2 \\
         \midrule
         \cref{fig:zero_shot} train & i7-14700K (20) & 66 & RTX 4070 (12) & 550.120, 12.4 & 24.04.2 \\
         \cref{fig:zero_shot} 4x4 train & i7-14700K (20) & 66 & --- & --- & 24.04.2 \\
         \cref{fig:zero_shot} test & EPYC 7402 (24) & 256 & --- & --- & 20.04.2 \\
         \cref{fig:results:node-scaling} & EPYC 7402 (24) & 256 & --- & --- & 20.04.2 \\
         \cref{section:results:zero-shot} molecular systems & i7-14700K (20) & 66 & --- & --- & 24.04.2 \\
         \midrule
         \cref{sec:benchmarking} & EPYC 7402 (24) & 256 & --- & --- & 20.04.2 \\
         \bottomrule
    \end{tabular}
    \label{tab:hardware}
\end{table}

\section{Ablation studies}\label{appendix:ablation-studies}
\begin{table}
    \centering
    \caption{Ablation study showing the influence of widths of the dense layers encoding the node and edge features and the linear layer on the mean squared error between predicted and true system dynamics \(\dot{y}\).
    }
    \scriptsize
    \begin{tabular}{llll}
        \toprule
        & \(d_L = 128\) & \(d_L = 256\) & \(d_L = 512\) \\
        \midrule
        \(d_h = 8\)  & 2.879e-02 & 1.336e-02 & 3.666e-03 \\
        \(d_h = 16\) & 1.657e-03 & 1.634e-04 & 6.571e-05 \\
        \(d_h = 32\) & 8.290e-04 & 2.011e-04 & 1.278e-05\\
        \(d_h = 64\) & 7.826e-04 & 3.037e-05 & 7.713e-06\\
        \bottomrule
    \end{tabular}
    \label{tab:ablation}
\end{table}

\begin{figure}
    \centering
    \includegraphics[width=0.97\linewidth]{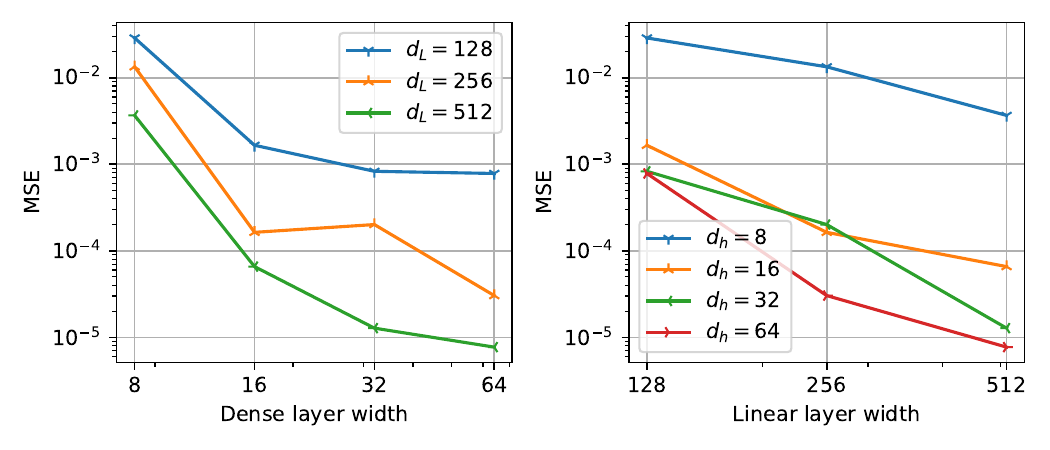}
    \caption{Ablation study for widths of the dense and linear layers.}
    \label{fig:abl_widths}
\end{figure}
We vary the widths of the encoders and linear layers to understand how they affect the mean squared error defined on the true and predicted solutions. We experimented with the Hamiltonian \(\mathcal{H}_2\) explained in \Cref{section:results:zero-shot}, chain of 8 nodes in 2D. We evaluated the model on the training and testing sets with 2000 samples each in the phase space and report the test errors. The message encoder's width is chosen by subtracting the width of the hidden layer from the width of the linear layer in the ablation study, i.e., $d_M = d_L-d_h$.
\Cref{fig:abl_widths} and \Cref{tab:ablation} reveal that increasing either the linear layer width $d_L$ or the hidden dimension $d_h$ while keeping the other parameter fixed consistently reduces the mean squared error. In the same experiment, we also computed the condition number \(\kappa(Z) = \frac{\sigma_1}{\sigma_n}\) in terms of the singular values \(\sigma_1 > \dots > \sigma_n\) of the matrix \(Z\) associated with the linear system in \Cref{eq:fully-linear-system} to assess the sensitivity of the solution. We avoided the bias term when computing the condition, as it was only used to fit the integration constant in practice. \Cref{tab:ablation-cond} reveals larger values as the system size (network width \(d_L\)) increases, but increasing the feature width of the encoders \(d_h\) slightly stabilizes the system for large network widths, as expected.

\begin{table}
    \centering
    \caption{Ablation study showing the condition number of the matrix associated with \Cref{eq:fully-linear-system} with increasing feature widths.}
    \scriptsize
    \begin{tabular}{lccc}
        \toprule
        & \(d_L = 128\) & \(d_L = 256\) & \(d_L = 512\) \\
        \midrule
        \(d_h = 8\)  & 4.266e+05 & 8.985e+06 & 2.612e+08 \\
        \(d_h = 16\) & 1.525e+05 & 1.687e+06 & 3.545e+07 \\
        \(d_h = 32\) & 8.105e+04 & 1.352e+06 & 2.974e+07 \\
        \(d_h = 64\) & 1.072e+05 & 9.625e+05 & 1.766e+07 \\
        \bottomrule
    \end{tabular}
    \label{tab:ablation-cond}
\end{table}

We additionally vary the number of message passes (\#msg) by recursively applying the local-pooling \(h^V_j \gets \sum_{i \in \mathcal{N}_j}{h^V_i}\) aggregating the node encodings \(h^V_i\) of the neighboring nodes \(i \in \mathcal{N}_j\) (\#msg$-1$ times), and then applying the final message scheme explained in \Cref{section:method:model:message-passing} with two different schemes: summing (\(h^V_j \gets \sum_{i \in \mathcal{N}_j}{h^V_i}\)) and averaging \((h^V_j \gets \frac{1}{|\mathcal{N}_j|} \sum_{i \in \mathcal{N}_j}{h^V_i}\)).
\Cref{tab:ablation-message-passing} reveals that having multiple message-passes can improve the accuracy for the 8-particle mass-spring system when averaging is used.
We believe that summing works better than averaging because it implicitly encodes the node degree information by aggregating the neighboring messages.
Each neighboring message is the output of a \texttt{softplus} activation function and has non-negative values.
In all the other experiments presented in this paper, we use only a single message pass and do not optimize the number of message passes, as all the ground truth systems we consider only require a single step of neighborhood information.

\begin{table}
    \centering
    \caption{Ablation study showing the influence of applying recursive message-passing on the mean squared error between predicted and true system dynamics \(\dot{y}\). \texttt{\#msg} is the number of message passes.
    }
    \scriptsize
    \begin{tabular}{lllllll}
        \toprule
        & \texttt{\#msg} \(=1\) & \texttt{\#msg}\(=2\) & \texttt{\#msg}\(=3\) & \texttt{\#msg}\(=4\) & \texttt{\#msg}\(=5\) & \texttt{\#msg}\(=6\) \\
        \midrule
        Summing & 7.713e-06 & 1.866e-05 & 1.271e-04 & 1.150e-03 & 7.620e-04 & 2.222e-03 \\
        Averaging & 2.020e-02 & 1.265e-02 & 1.280e-02 & 1.057e-02 & 1.175e-02 & 1.089e-02 \\
        \bottomrule
    \end{tabular}
    \label{tab:ablation-message-passing}
\end{table}

\section{Comparison with a benchmark dataset}
\label{sec:benchmarking_appendix}
To further support our claims, here we perform benchmarking of our model against existing suitable graph network approaches. We made use of the existing publication from the NeurIPS 2022 Datasets and Benchmarks Track by \citet{thangamuthu-2022} and their corresponding repository. The considered models for comparison include: \begin{itemize}
    \item \textbf{Full Graph Neural Network (FGNN)} : Based on the work of \citet{sanchez-2020}, these models utilize message-passing as a key feature to enable a simulation framework. Note that in the original work the architecture is called Graph Network-based Simulators (GNS) but for benchmarking it is called FGNN and we use this name as well.
    \item \textbf{Full Graph Neural ODE (FGNODE)} : An ODE version of FGNN is what we refer to as FGNODE \citep{sanchez-gonzalez-2019-hgnn}.
    \item \textbf{Graph Neural ODE (GNODE)} : This architecture uses a graph topology to parameterize the force of a system using a neural ODE approach, it was introduced by \citet{thangamuthu-2022}.
    \item \textbf{Lagrangian Graph Network (LGN)} : This architecture uses an FGNN to predict the Lagrangian of the system \citep{bhattoo-2022-lgn-ridig}.
    \item \textbf{Lagrangian Graph Neural Network (LGNN)} : Similar to LGN, this architecture decoples the kinetic and potential energies \citep{thangamuthu-2022}.
    \item \textbf{Hamiltonian Graph Network (HGN)} : In this architecture an FGNN predicts the Hamiltonian of the system \citep{sanchez-gonzalez-2019-hgnn, thangamuthu-2022}.
    \item \textbf{Hamiltonian Graph Neural Netwrok (HGNN)} : Analogously, this architecture is similar to HGN but it decouples the potential and kinetic energies of the Hamiltonian \citep{thangamuthu-2022}.
\end{itemize}

First, we highlight the similarities and differences in model properties in table \ref{tab:properties}, noting that our model satisfies requirements necessary for modeling physical systems while maintaining energy conservation.

\begin{table}
    \centering
    \caption{Comparison of the SOTA physics-informed graph network architectures (also see \Cref{tab:benchmarking}) and our (SWIM) RF-HGN.}
    \scriptsize
    \begin{tabular}{p{3.3cm} p{1.4cm} p{0.7cm} p{1.155cm} p{1.1cm} p{0.55cm}p{0.7cm}p{0.55cm}p{0.7cm}}
        \toprule
        Model & (SWIM) RF-HGN & FGNN & FGNODE & GNODE & LGN & LGNN & HGN & HGNN \\
        \midrule
        Translation invariance & \checkmark &  \checkmark & \xmark & \xmark & \xmark & \xmark  & \xmark  & \xmark  \\
        Rotation invariance & \checkmark & \xmark  & \xmark & \xmark & \xmark & \xmark & \xmark & \xmark \\
        Energy conservation & \checkmark & \xmark & \xmark & \xmark & \checkmark & \checkmark & \checkmark & \checkmark \\
        Gradient-descent-free \\ training & \checkmark & \xmark & \xmark & \xmark & \xmark  & \xmark  & \xmark & \xmark \\
        \bottomrule
    \end{tabular}
    \label{tab:properties}
\end{table}


A key metric of interest for our work is the training time, thus we have re-trained models from \citep{thangamuthu-2022} on their datasets of the spring system with 3, 4 and 5 nodes and record the training time. All runs were performed on the same machine as the experiments in the main paper. The resulting training times in table \Cref{tab:benchmarking} show clearly that our proposed approach is much faster to train, especially compared to the specialized Lagrangian and Hamiltonian graph networks.


Of course, a model is only useful if it can accurately make predictions, thus we plot errors on a test trajectory for all mentioned models in \Cref{fig:benchmark_errors}. We observe that our (SWIM) RF-HGN has a similar predictive ability as the SOTA architectures. It should be noted that for the test trajectory shown for $N=4$ the LGN model diverged after around 25 steps. Similar results of diverging models were also observed from LGNN and the NODE architectures when we attempted to test on 100 trajectories, where multiple predicted trajectories would diverge from the true trajectory.

\section{Additional results}
\label{sec:additional_results}

Our submitted folder contains an animation of the test system shown in Figure 1 of the main text, as well as the molecular dynamics systems (see \Cref{fig:results:9particles-lj-snaps} and \Cref{fig:results:32train-64test-lj-snaps}).

\subsection{Benchmarking against SOTA optimizers}

\begin{figure}
  \begin{center}
  \includegraphics[width=1.0\textwidth]{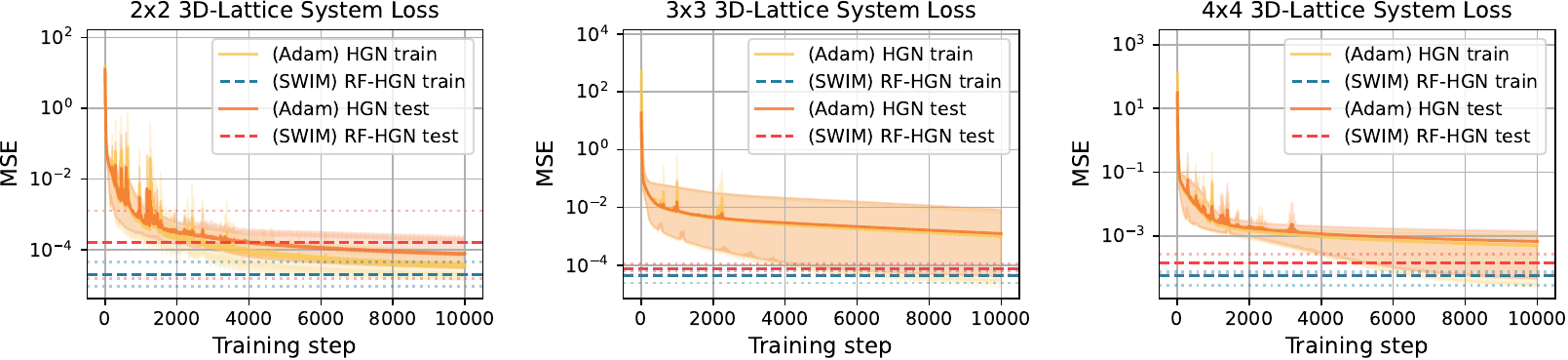}
  \end{center}
  \caption{MSE losses on the training and test dataset for a 2x2 (left), 3x3 (middle) and 4x4 (right) lattice during iterative training are given with solid lines for the average over ten runs; the shaded region extends from the minimum to the maximum value. The dashed lines denote the (constant) MSE losses for our non-iterative optimization, and shaded dashes show the minimum and maximum.
  }
  \label{fig:results:adam-comparison}
\end{figure}

\begin{figure}
    \centering
    \includegraphics[width=0.99\linewidth]{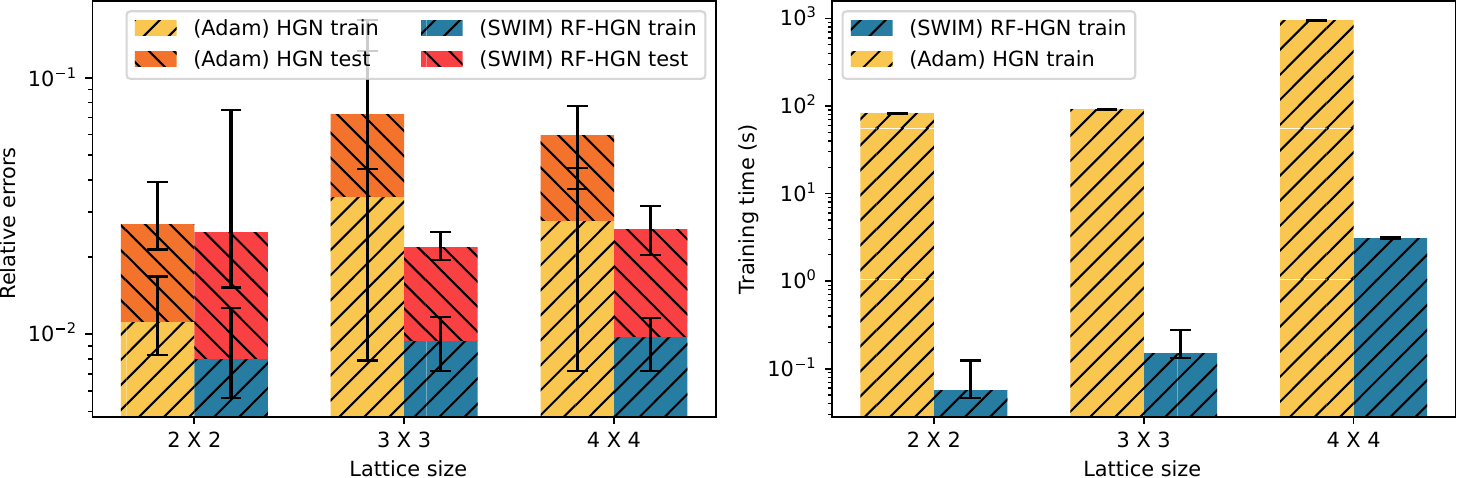}
    \vspace{12pt}
    \caption{
    Relative error and training time are shown for different lattice sizes. Boxplots show the mean and error bars based on ten runs with different random seeds.}
    \label{fig:errors_times}
\end{figure}

\begin{figure}
    \centering
    \includegraphics[width=0.99\linewidth]{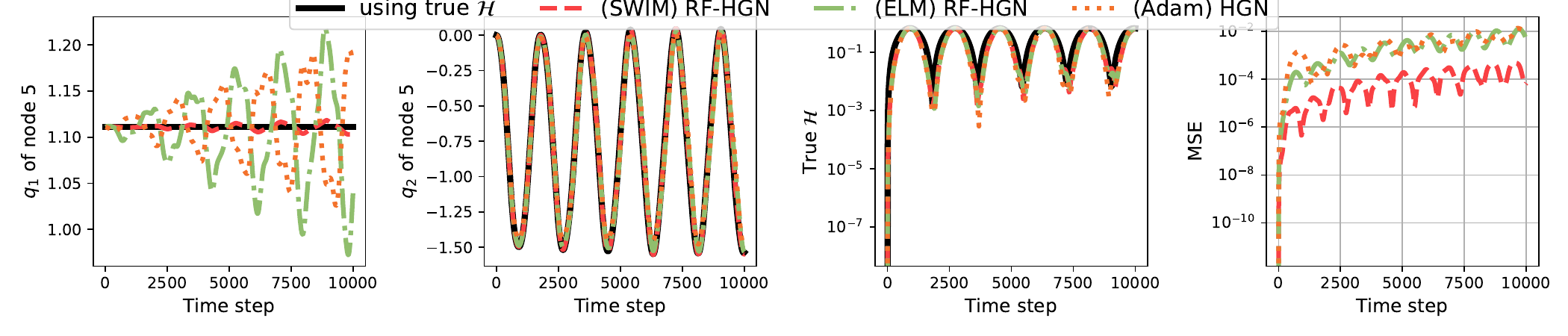}
    \caption{
    Illustration of position trajectories of the middle node over time (also see \Cref{fig:spring-chain-snaps}). Models are trained with 5 nodes and tested (here) with 10 nodes using the standard spring potential (see \Cref{app:datasets:optim-study}).
    }
    \label{fig:spring-chain-traj}
\end{figure}

\begin{figure}
    \centering
    \includegraphics[width=0.99\linewidth]{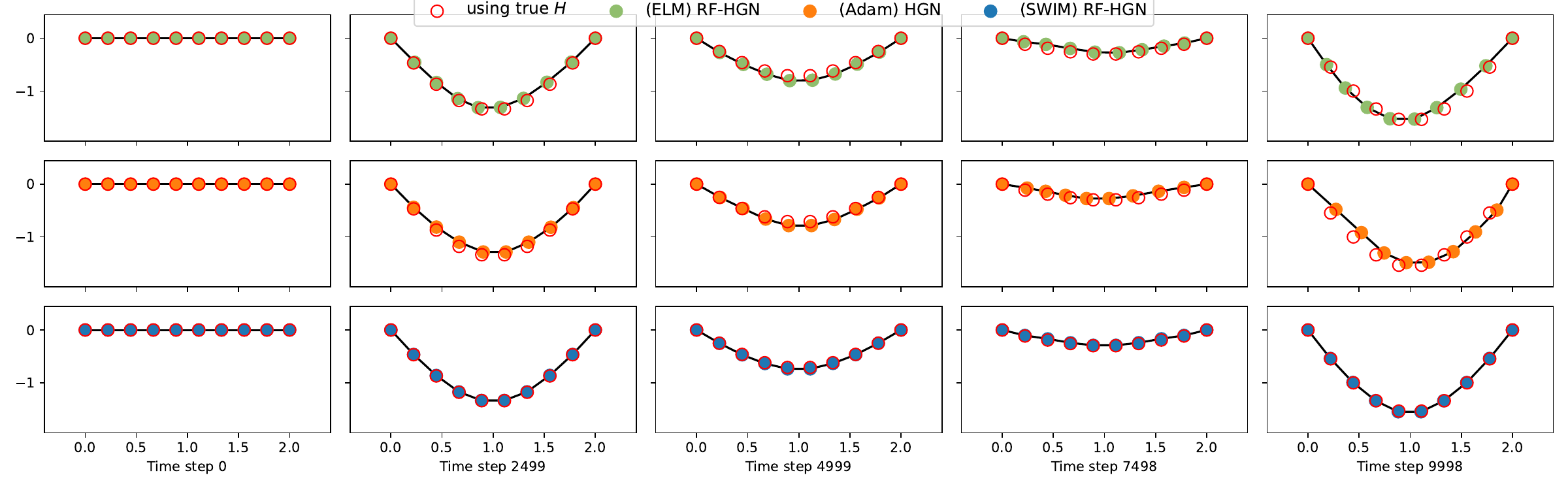}
    \caption{
    Illustration of position trajectories over time from models trained on a system with 5 nodes using the spring chain potential (see \Cref{app:datasets:optim-study}) and zero-shot tested with 10 nodes with an external force (gravitational).
    }
    \label{fig:spring-chain-snaps}
\end{figure}

\begin{figure}
    \centering
    \includegraphics[width=0.99\linewidth]{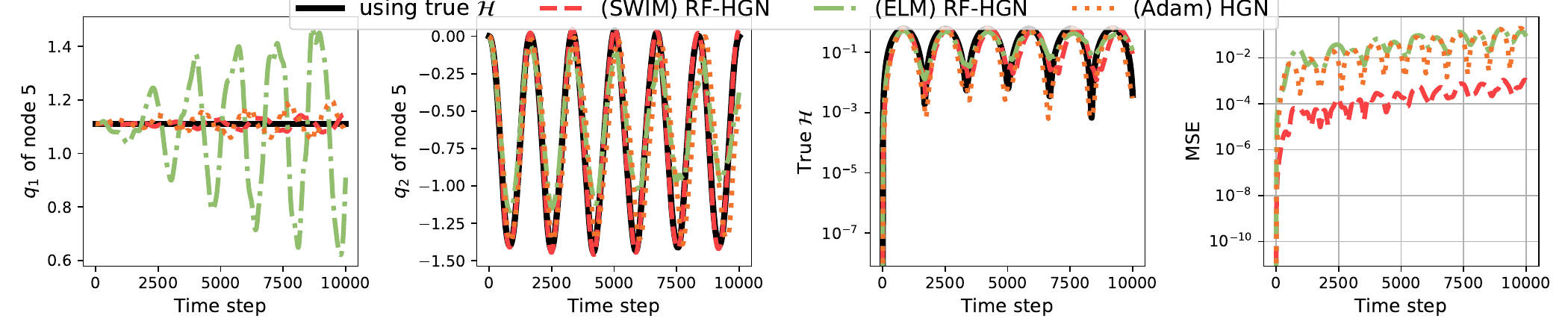}
    \caption{
    Illustration of position trajectories of the middle node over time (also see \Cref{fig:spring-chain-snaps}). Models are trained with 5 nodes and tested (here) with 10 nodes using anharmonic spring potential (see \Cref{app:datasets:optim-study}).
    }
    \label{fig:anharmonic-chain-traj}
\end{figure}

\begin{figure}
    \centering
    \includegraphics[width=0.99\linewidth]{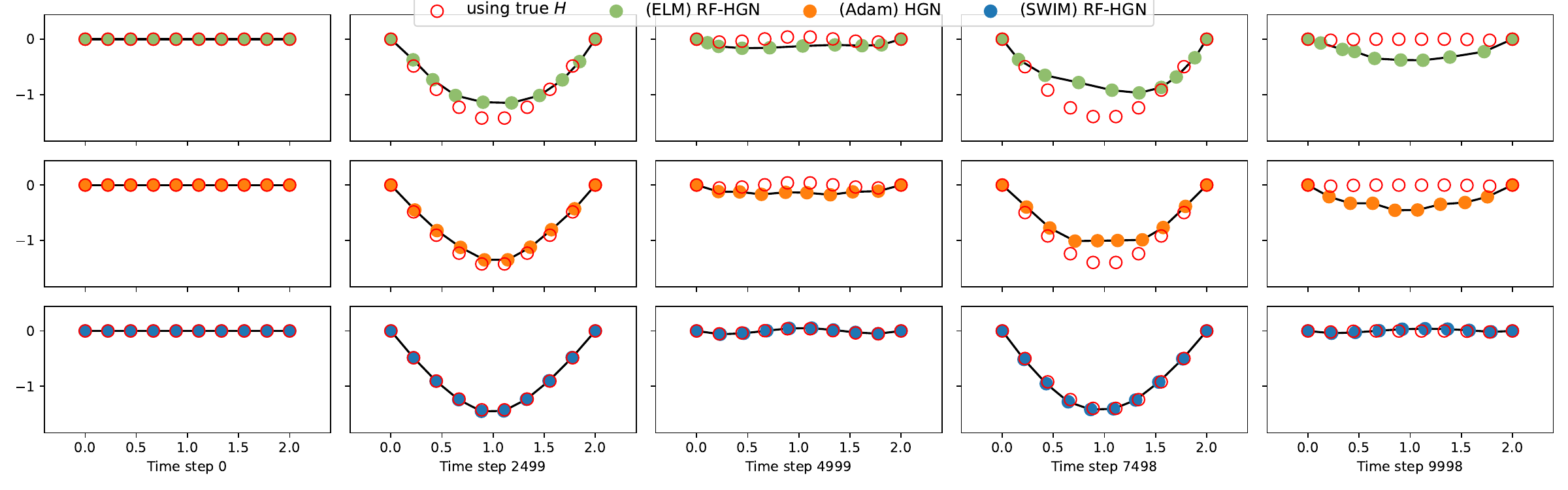}
    \caption{
    Illustration of position trajectories over time from models trained on a system with 5 nodes using anharmonic spring potential (see \Cref{app:datasets:optim-study}) and zero-shot tested with 10 nodes with an external force (gravitational).
    }
    \label{fig:anharmonic-chain-snaps}
\end{figure}

\begin{figure}
    \centering
    \includegraphics[width=0.99\linewidth]{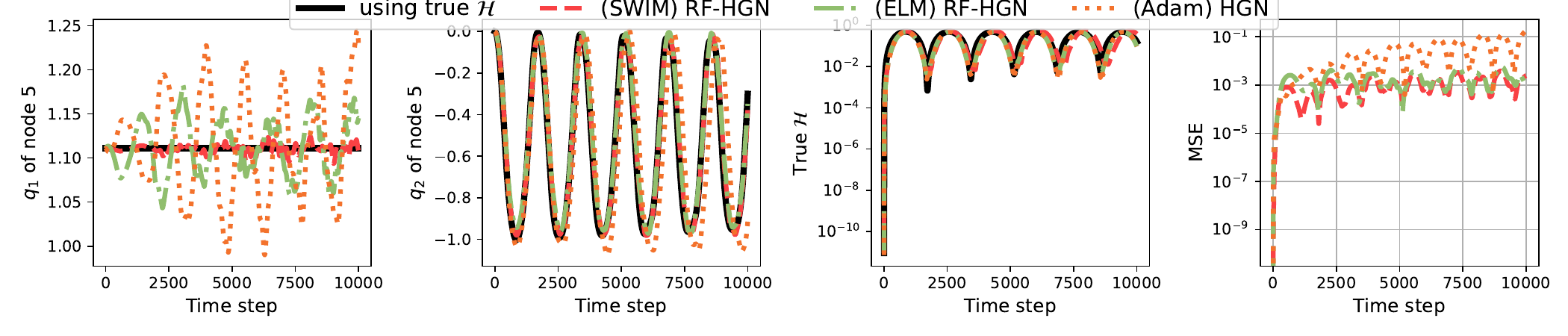}
    \caption{
    Illustration of position trajectories of the middle node over time (also see \Cref{fig:spring-chain-snaps}). Models are trained with 5 nodes and tested (here) with 10 nodes using the Morse potential (see \Cref{app:datasets:optim-study}).
    }
    \label{fig:morse-chain-traj}
\end{figure}

\begin{figure}
    \centering
    \includegraphics[width=0.99\linewidth]{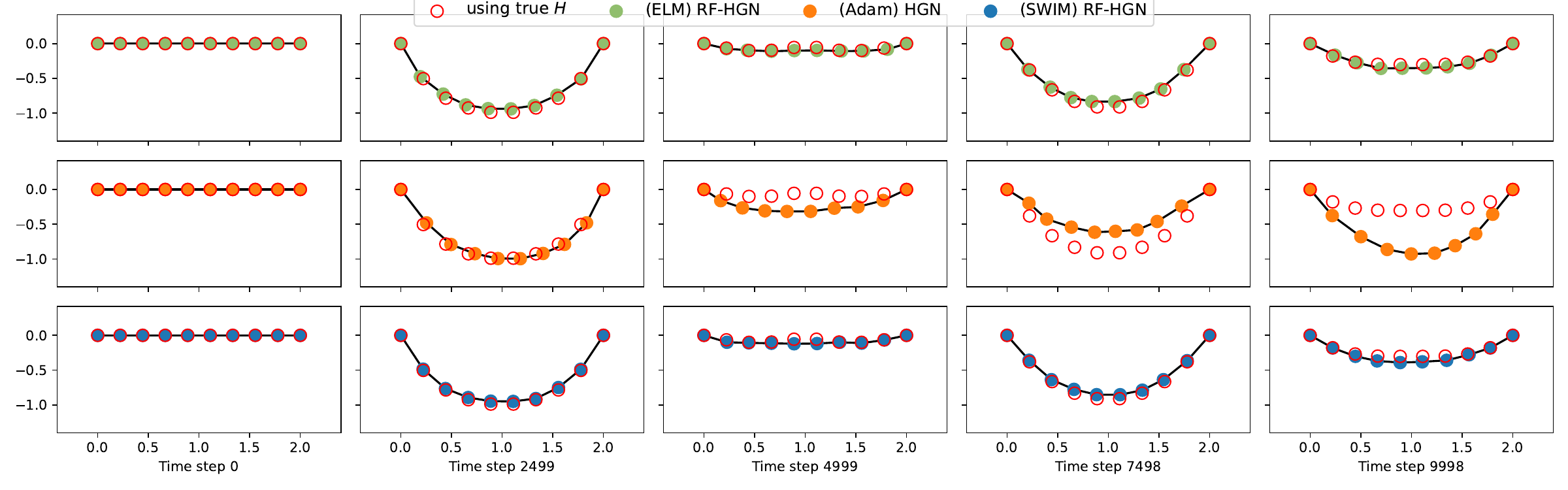}
    \caption{
    Illustration of position trajectories over time from models trained on a system with 5 nodes using the Morse spring potential (see \Cref{app:datasets:optim-study}) and zero-shot tested with 10 nodes with an external force (gravitational).
    }
    \label{fig:morse-chain-snaps}
\end{figure}

\begin{figure}
    \centering
    \includegraphics[width=0.7\linewidth]{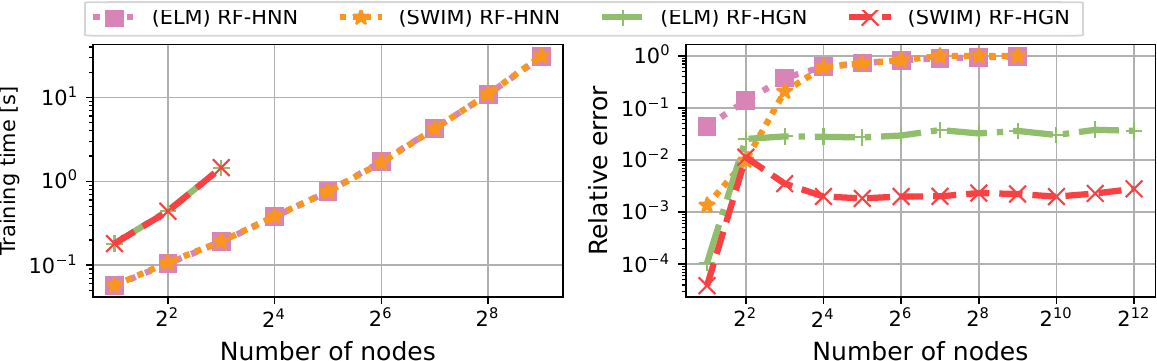}
    \vspace{8pt}
    \caption{Zero-shot generalization in 2D open chain (see \Cref{fig:experiments} (b)): RF-HGN trained up to \(N=8\) accurately generalizes up to  \(N=4096\), outperforming retrained RF-HNN (right).
    RF-HGN with zero-shot generalization is also faster than RF-HNN for node counts larger than $2^6$ (left).
    }
    \label{fig:results:node-scaling}
\end{figure}

In \Cref{fig:results:adam-comparison}, we show loss curves for the Adam optimizer, highlighting how its train and test losses evolve over time relative to the loss of our non-iterative approach in 2x2, 3x3, and 4x4 systems. The model and optimizer hyperparameters are set accordingly as explained in \Cref{app:datasets:optim-study}, \Cref{tab:appendix:optim-study-model-params}, \Cref{tab:appendix:optim-study-sota-hyperparams}, and \Cref{tab:appendix:optim-study-swim-hyperparams}.
We observe comparable accuracies of Adam and SWIM~\citep{bolager-2023-swim}, even after $10000$ gradient descent iterations using the Adam~\citep{adam-optimizer} optimizer.
Moreover, \Cref{fig:errors_times} reveals that our method scales better than iterative optimization, maintaining low error as system size increases. And \Cref{tab:times_table_adam_comparison} reveals two to three orders of magnitude quicker training of (SWIM) RF-HGN than (Adam) HGN in different 3D lattice systems.

\begin{table}
    \caption{Summary of the training times of the experiment presented in \Cref{fig:errors_times} for (SWIM) RF-HGN and (Adam) HGN in seconds. For the systems of sizes \(2\times2\) (GPU trained), \(3\times3\) (GPU trained), and \(4\times4\) (CPU trained), we observe approximately three, two, and three orders of magnitude faster training, respectively.}
    \centering
    \scriptsize
    \begin{tabular}{cll}
        \toprule
        System size & (SWIM) RF-HGN & (Adam) HGN \\
        \midrule
        \(2 \times 2\) & \(\approx 0.06\) seconds & \(\approx 82.93\) seconds \\
        \(3 \times 3\) & \(\approx 0.15\) seconds & \(\approx 92.1\) seconds \\
        \(4 \times 4\) & \(\approx 3.06\) seconds & \(\approx 936.12\) seconds \\
        \bottomrule
    \end{tabular}
    \label{tab:times_table_adam_comparison}
\end{table}

\subsection{Zero-shot generalization and comparison of random feature methods}\label{app:sec:additional-results:zero-shot}
\begin{figure}
  \begin{center}
    \includegraphics[width=0.99\textwidth]{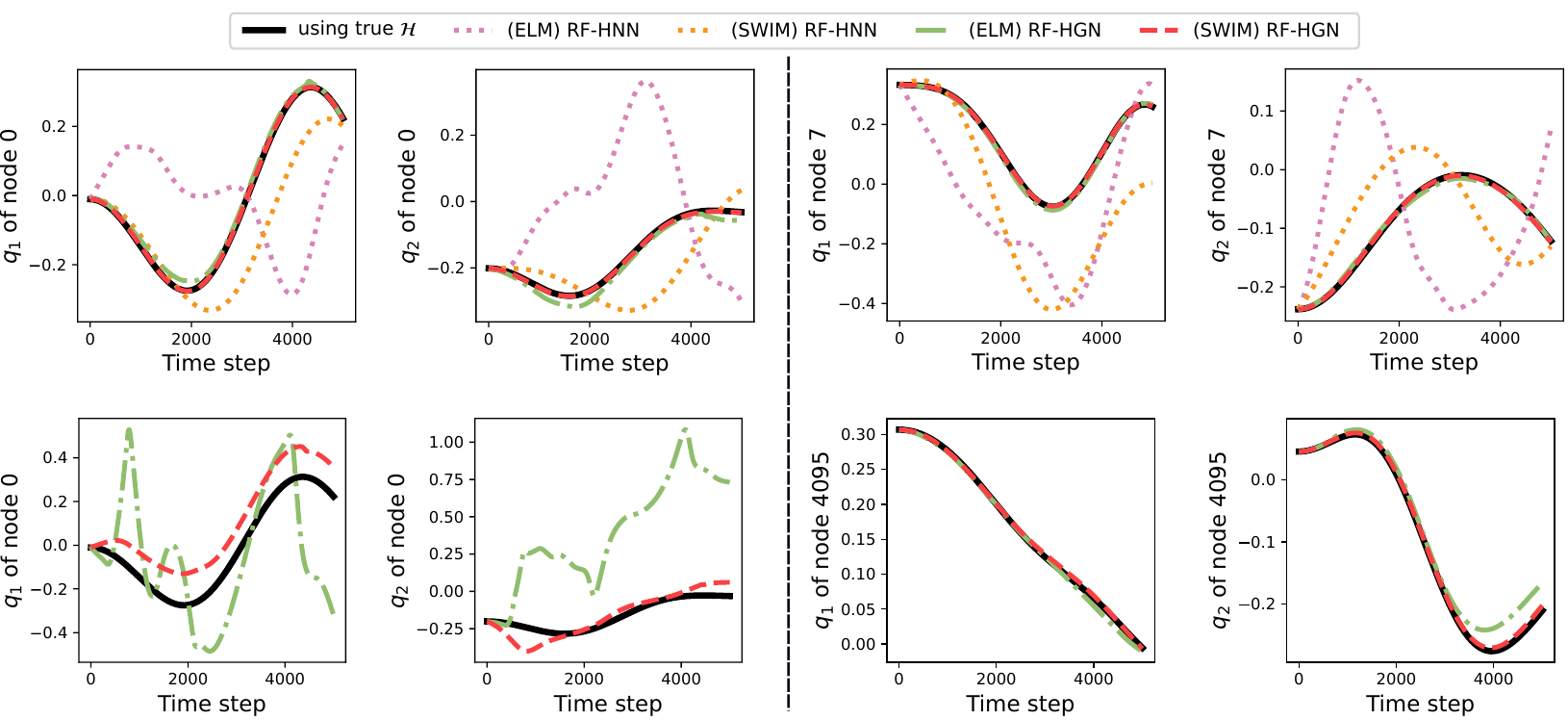}
  \end{center}
  \caption{Illustration of position trajectories of the corner nodes over time (also see \Cref{fig:results:node-scaling-integration-8-and-4096-all}). Top: Trained with \(2^3\) nodes and tested with \(2^3\) nodes. Bottom: Trained with \(2^3\) nodes and tested with \(2^{12}\) nodes.}
  \label{fig:results:traj_2D_chain_all_bothcorners}
\end{figure}

\begin{figure}
  \begin{center}
    \includegraphics[width=0.99\textwidth]{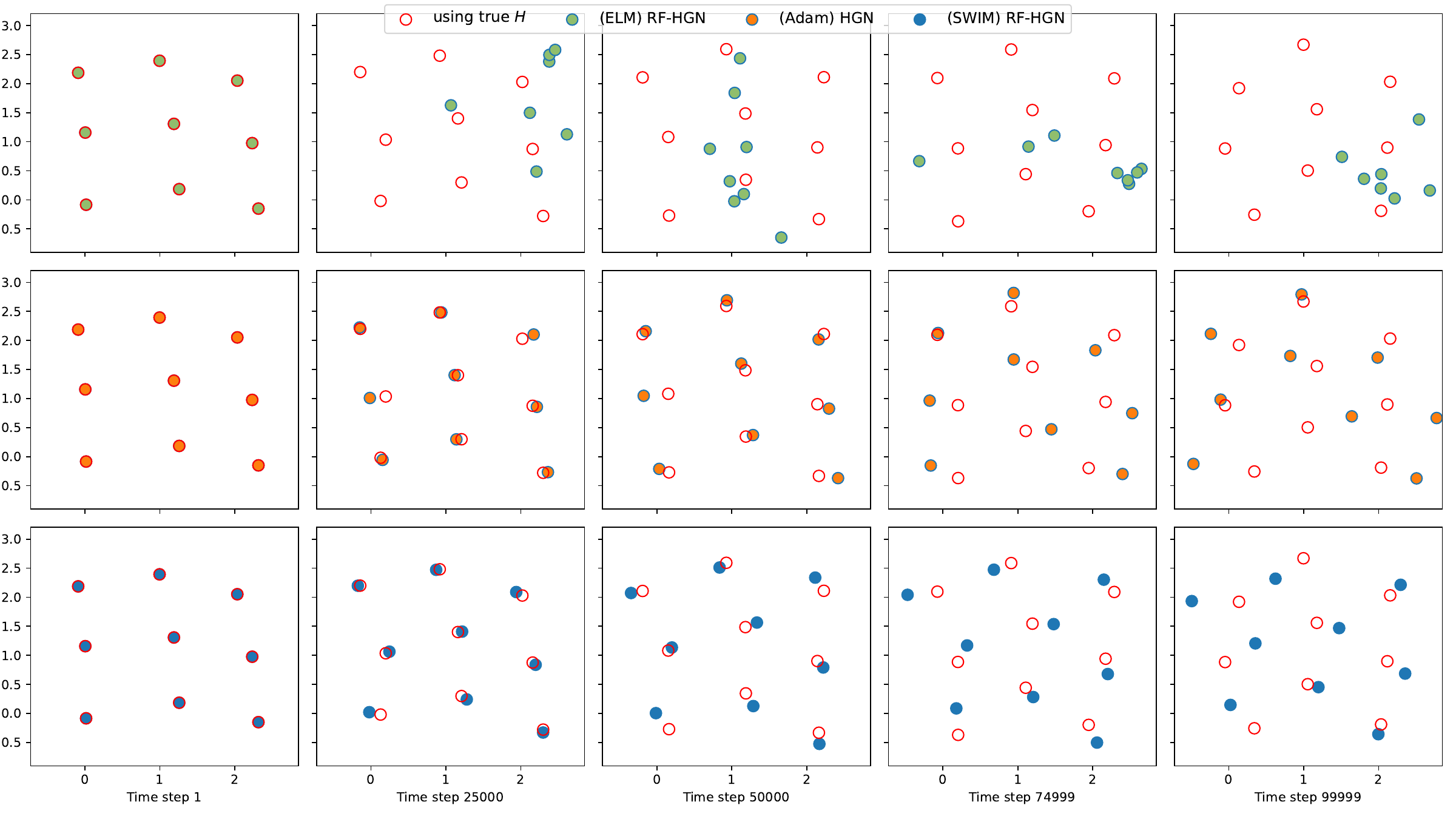}
  \end{center}
  \caption{
    Illustration of position trajectories over time from models trained on a system with 9 nodes on the 2D molecular dynamics system (see \Cref{fig:experiments} (c)).
  }
  \label{fig:results:9particles-lj-snaps}
\end{figure}

\begin{figure}
  \begin{center}
    \includegraphics[width=0.99\textwidth]{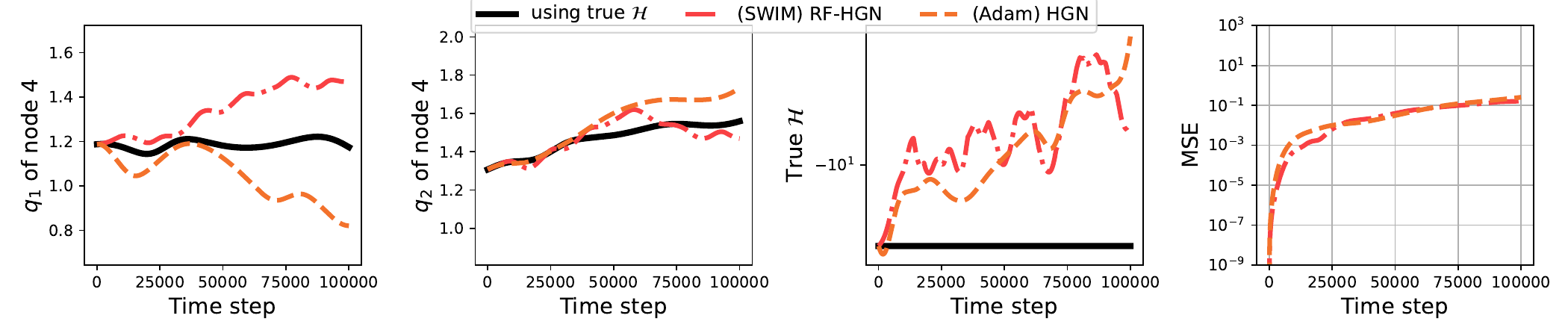}
  \end{center}
  \caption{
    Illustration of position trajectories over time from models trained on a system with 9 nodes on the 2D molecular dynamics system (see \Cref{fig:experiments} (c)).
    Results from ELM RF-HGN training are omitted due to very large errors which distort the representations in the plots.
  }
  \label{fig:results:9-lj-traj}
\end{figure}

\begin{figure}
  \begin{center}
    \includegraphics[width=0.99\textwidth]{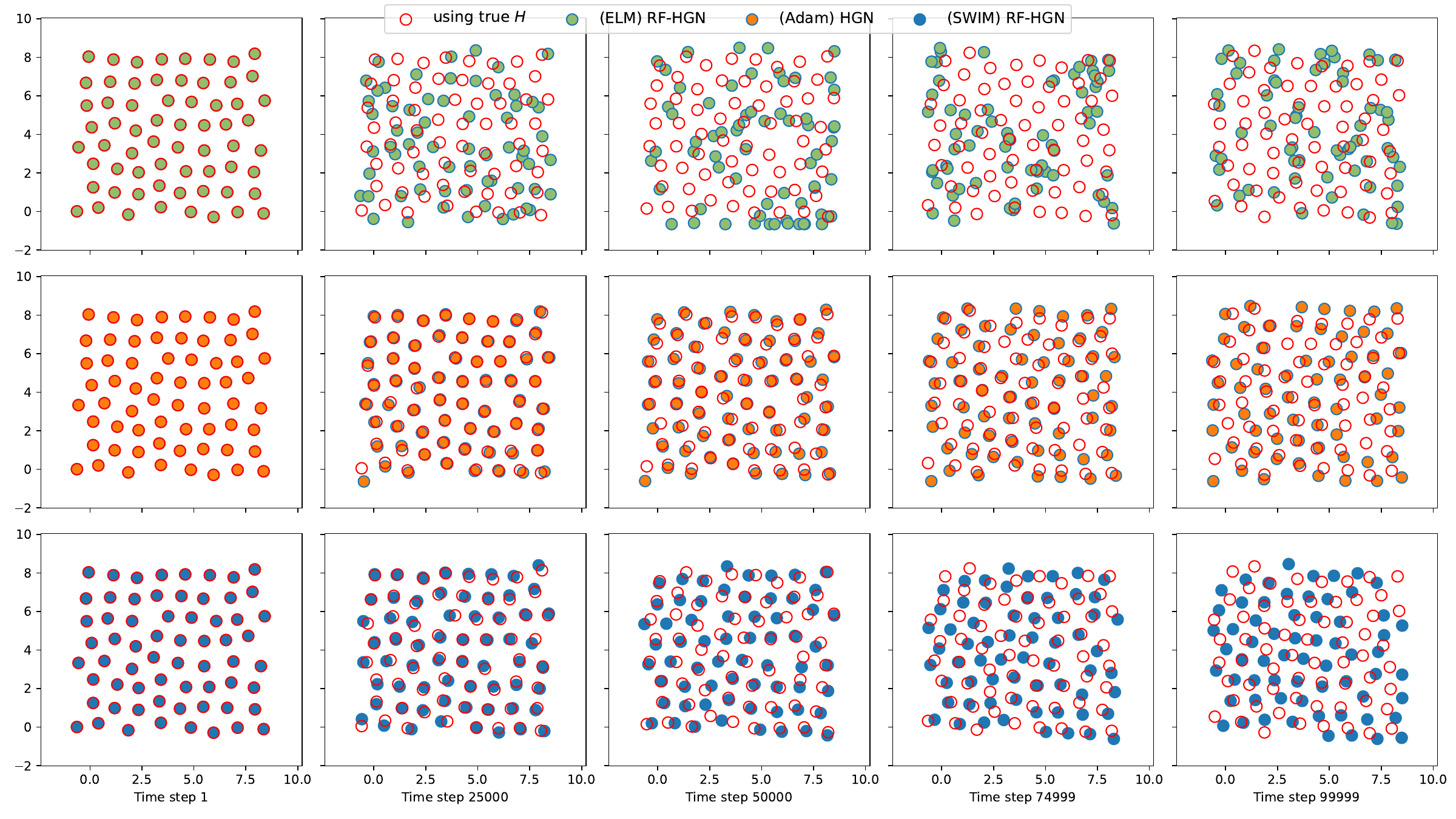}
  \end{center}
  \caption{
    Illustration of position trajectories over time from models trained on a system with 36 nodes and zero-shot tested with 64 nodes on the 2D molecular dynamics system (see \Cref{fig:experiments} (d)).
  }
  \label{fig:results:32train-64test-lj-snaps}
\end{figure}

\begin{figure}
  \begin{center}
    \includegraphics[width=0.99\textwidth]{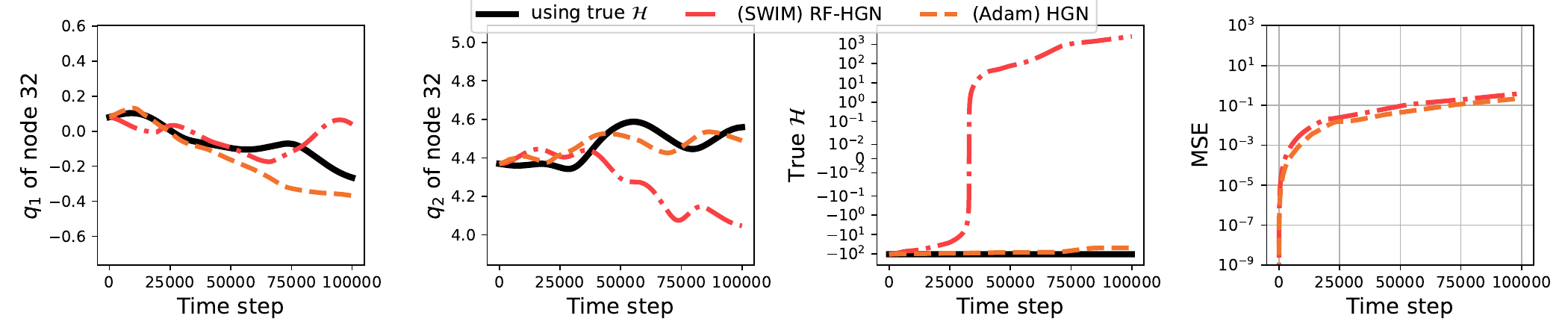}
  \end{center}
  \caption{
    Illustration of position trajectories over time from models trained on a system with 36 nodes and tested with 64 nodes on the 2D molecular dynamics system (see \Cref{fig:experiments} (c)). Results from ELM RF-HGN training are omitted due to very large errors which distort the representation in the plots.
  }
  \label{fig:results:64-lj-traj}
\end{figure}

\begin{figure}
    \centering
    \includegraphics[width=0.98\linewidth]{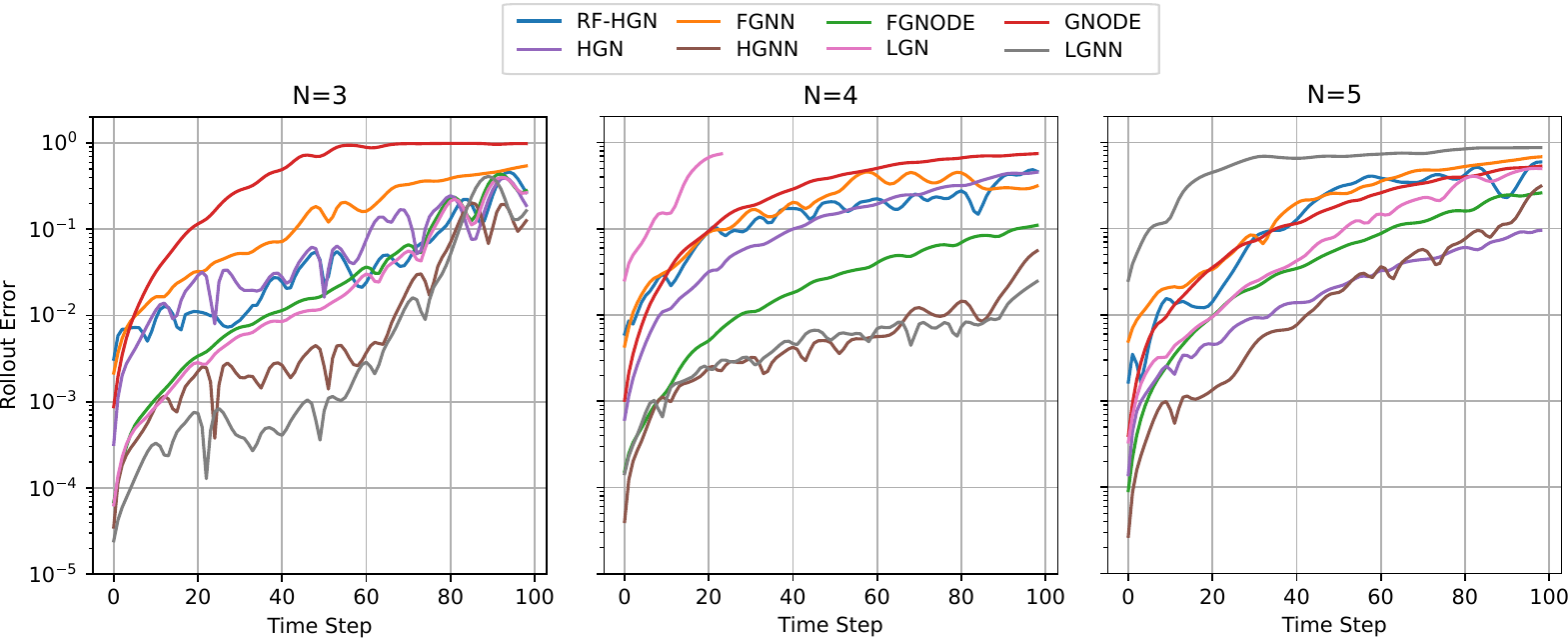}
    \vspace{12pt}
    \caption{Rollout errors on test trajectory benchmark for $N=3,4,5$.}
    \label{fig:benchmark_errors}
\end{figure}

\Cref{fig:results:traj_2D_chain_all_bothcorners} illustrates the trajectories of the corner nodes of the experiment in \Cref{fig:results:node-scaling-integration-8-and-4096-all}. For this particular example the left corner trajectory (node with id 0) seems to be harder to capture than the other nodes in the system for the extreme zero-shot case (trained with 8 nodes, tested with 4096 nodes) case, hence slightly higher error on the trajectories compared to the non-zero-shot case (trained with 8 nodes and tested with 8 nodes) as one can see in \Cref{fig:results:node-scaling-integration-8-and-4096-all}.

\begin{table}
    \centering
    \caption{Molecular dynamics evaluation with 9 particles. Mean squared error (MSE) and relative \(l^2\) error (rel. \(l^2\)) are reported together with the true Hamiltonian over the ground-truth trajectory and the (ELM) RF-HGN predicted quantity over the rolled-out trajectory.}
    \label{tab:9-lj-elm}
    \scriptsize
    \begin{tabular}{llllll}
    \toprule
        & T=1 & T=25000 & T=50000 & T=74999 & T=99999 \\
    \midrule
    \(q\) MSE & 8.651e-08 &  9.369e+07 &  2.926e+09 &  9.456e+08 &  6.337e+09 \\
    \(q\) rel. \(l^2\) & 2.044e-04 &  6.766e+03 &  3.718e+04 &  2.139e+04 &  5.615e+04 \\
    True \(\mathcal{H}\) &
    -1.233e+01 & -1.233e+01 & -1.233e+01 & -1.233e+01 & -1.233e+01 \\
    Model \(\widehat{\mathcal{H}}\) &
    -1.450e+01 &  2.979e+10 &  -1.122e+10 & -3.085e+10 &   8.765e+10 \\
    \bottomrule
    \end{tabular}
\end{table}

\begin{table}
    \centering
    \caption{Molecular dynamics evaluation with 9 particles. Mean squared error (MSE) and relative \(l^2\) error (rel. \(l^2\)) are reported together with the true Hamiltonian over the ground-truth trajectory and the (Adam) HGN predicted quantity over the rolled-out trajectory.}
    \label{tab:9-lj-adam}
    \scriptsize
    \begin{tabular}{llllll}
    \toprule
        & T=1 & T=25000 & T=50000 & T=74999 & T=99999 \\
    \midrule
    \(q\) MSE & 3.089e-13 &  5.043e-03  & 1.554e-02  & 6.052e-02  & 1.279e-01 \\
    \(q\) rel. \(l^2\) & 3.863e-07 &  4.964e-02 &  8.568e-02 &  1.711e-01 &  2.522e-01 \\
    True \(\mathcal{H}\) &
    -1.233e+01 & -1.233e+01 & -1.233e+01 & -1.233e+01 & -1.233e+01 \\
    Model \(\widehat{\mathcal{H}}\) & -1.233e+01 &  -1.233e+01 & -1.233e+01 & -1.233e+01 & -1.233e+01 \\
    \bottomrule
    \end{tabular}
\end{table}

\subsection{Benchmarking against SOTA architectures}

\subsection{Robustness against noise}\label{appendix:sec:noise-study}

To evaluate our model's robustness against additive noise, we add Gaussian noise with different standard deviations (\(\sigma\)) to the ground truth positions and momenta before including them in the training set. We experimented with 5 nodes in 3D on the training set with 1000 samples in phase space using 5-fold cross-validation, and on a test set of 1000 samples in phase space. To stabilize the results, we repeated this experiment 5 times with different seeds and report the average results in \Cref{tab:noise-study}.

\begin{table}
    \centering
    \caption{Results of training with noisy data are displayed. Relative \(l^2\) and mean-squared errors are displayed.}
    \label{tab:noise-study}
    \scriptsize
    \begin{tabular}{lllllllll}
        \toprule
        \(\sigma\) & 0 & 1e-5 & 1e-4 & 1e-3 & 1e-2 & 1e-1 & 1 & 2 \\
        \midrule
        CV MSE & 6.18e-3 & 6.59e-3 & 5.35e-3 & 7.84e-3 & 6.45e-3 & 3.38e-2 & 2.78 & 1.11e+1 \\
        CV rel. \(l^2\) & 4.03e-2 & 4.07e-2 & 3.72e-2 & 4.29e-2 & 4.05e-2 & 9.61e-2 & 6.64e-1 & 8.78e-1 \\
        Test MSE & 9.52e-3 & 9.77e-3 & 9.12e-3 & 1.04e-2 & 9.12e-3 & 9.67e-3 & 1.01e-1 & 3.94e-1 \\
        Test rel. \(l^2\) & 4.96e-2 & 4.92e-2 & 4.83e-2 & 5.09e-2 & 4.80e-2 & 4.94e-2 & 1.65e-1 & 3.27e-1 \\
        \bottomrule
    \end{tabular}
\end{table}

The columns in \Cref{tab:noise-study} list the standard deviation \(\sigma\) used in the experiments. The results indicate the robustness of our method to Gaussian noise added to the state. For more realistic scenarios, the model can further be improved by modeling uncertainties and assessing sensitivities for noise in a real-data setting.

\subsection{Batch-wise training}\label{appendix:sec:batch-wise}

To demonstrate batching, we prepared an example with Lennard-Jones potential with 4 particles in 2D trained with both ELM, SWIM, and their batch-wise versions when solving the linear problem (\Cref{eq:fully-linear-system}). Batching is performed by sub-sampling the training data set (5000 states in phase space) and averaging the resulting last-layer coefficients. \Cref{tab:batch-wise-training} lists the results, where the columns indicate which random feature method is used, rows indicate cross-validation (5-fold) errors with train set size 5000 and test errors averaged over 5 different seeds of test size 5000.

\begin{table}
    \centering
    \caption{Batch-wise training results compared to direct least squares solutions are displayed with training time in seconds and memory usage in GiB. Relative \(l^2\) and mean-squared errors are displayed.}
    \scriptsize
    \begin{tabular}{lllll}
        \toprule
         & ELM & ELM (batched) & SWIM & SWIM (batched) \\
        \midrule
        CV MSE & 9.01 & 2.15e+03 & 3.84e-1 & 1.08 \\
        CV rel. \(l^2\) & 2.59e-1 & 2.54 & 4.65e-2 & 7.89e-2 \\
        Test MSE & 9.06 & 1.98e+3 & 3.71e-1 & 1.09 \\
        Test rel. \(l^2\) & 2.6e-1 & 2.49 & 4.61e-2 & 8.02e-2 \\
        Training time & 3.96 & 4.48 & 3.88 & 4.42 \\
        Memory usage & 2.4 & 1.1 & 2.4 & 1.1 \\
        \bottomrule
    \end{tabular}
    \label{tab:batch-wise-training}
\end{table}

We note that the memory required for batch-wise training can be further improved with additional tuning, for example, by explicitly freeing the GPU memory, we could tune the memory requirement to be as minimal as possible (0.04 GiB). In this case, however, the training time also increases to around \(5.7\) seconds. Different tuning strategies (e.g., for lower memory, for quicker runtime) are therefore important to consider when comparing different training strategies.

We believe that with an established linear solver like LSQR \citep{paige-1982-lsqr}, LSMR \citep{fong-2011-lsmr}, LSRN \citep{mahoney-2014-lsrn}, one can further study the batch-wise training of our model for even larger systems, tuned for specific needs such as low memory or fast training.

\subsection{Benchmarking different random features}\label{appendix:sec:different-rf}

We also experimented with random Fourier features (RFF) \citep{rahimi-2007random-fourier-features} by setting
\[ W_{ij} \sim \mathcal{N}(0, \frac{1}{\sigma^{\texttt{RFF}}}), \,\,\,\, b_i \sim \texttt{Uniform}(0, 2\pi), \,\,\,\, z = \sqrt{\frac{2}{\texttt{\#features}}}\cos(W\tran x + b), \]

where \(z\) is the random features and \(\texttt{\#features}\) is the size of \(z\).

\begin{table}
    \centering
    \caption{Results of training with noisy data using RFF are displayed. Relative \(l^2\) and mean-squared errors are displayed.}
    \label{tab:noise-study-with-rff}
    \scriptsize
    \begin{tabular}{lllllllll}
        \toprule
        \(\sigma\) & 0 & 1e-5 & 1e-4 & 1e-3 & 1e-2 & 1e-1 & 1 & 2 \\
        \midrule
        CV MSE & 3.02 & 3.03 & 3.05 & 3.04 & 3.01 & 3.09 & 5.59 & 1.4e+1 \\
        CV rel. \(l^2\) & 9.1e-1 & 9.14e-1 & 9.15e-1 & 9.13e-1 & 9.1e-1 & 9.17e-1 & 9.41e-1 & 9.87e-1 \\
        Test MSE & 3.43 & 3.61 & 3.44 & 3.47 & 3.48 & 3.39 & 3.62 & 3.87 \\
        Test rel. \(l^2\) & 9.74e-1 & 1 & 9.77e-1 & 9.82e-1 & 9.84e-1 & 9.71e-1 & 1 & 1.04 \\
        \bottomrule
    \end{tabular}
\end{table}
\begin{table}
    \centering
    \caption{Results of training with noisy data using ELM are displayed. Relative \(l^2\) and mean-squared errors are displayed.}
    \label{tab:noise-study-with-elm}
    \scriptsize
    \begin{tabular}{lllllllll}
        \toprule
        \(\sigma\) & 0 & 1e-5 & 1e-4 & 1e-3 & 1e-2 & 1e-1 & 1 & 2 \\
        \midrule
        CV MSE & 3e-1 & 3.2 & 2.44e-1 & 6.78e-1 & 2.4e-1 & 4.62e-1 & 8.75 & 2.32e+2 \\
        CV rel. \(l^2\) & 2.54e-1 & 5.22e-1 & 2.39e-1 & 3.24e-1 & 2.31e-1 & 2.87e-1 & 8.73e-1 & 1.82 \\
        Test MSE & 4.7e-1 & 2.89e-1 & 2.63e-1 & 3.65e-1 & 3.04e-1 & 3.06e-1 & 5.91e+1 & 2.42e+6 \\
        Test rel. \(l^2\) & 3.08e-1 & 2.76e-1 & 2.66e-1 & 3.02e-1 & 2.76e-1 & 2.76e-1 & 1.4 & 1.65e+2 \\
        \bottomrule
    \end{tabular}
\end{table}
We have extended the noise-scale experiment in \Cref{appendix:sec:noise-study} with \(\sigma^{\texttt{RFF}} = 1\) and list the results in table \Cref{tab:noise-study-with-rff}. Additionally, we run the same experiment using ELM by setting
\[ W_{ij} \sim \mathcal{N}(0, 1), \,\,\,\, b_i \sim \texttt{Uniform}(-\pi, \pi), \]
and list the results in \Cref{tab:noise-study-with-elm}

We note that data-agnostic methods perform better when tuned slightly towards the problem. To demonstrate this, we experimented on the same system but without noise. In \Cref{tab:fourier-results} we list relative \(l^2\) errors for RFF; the parameters \(\sigma^{\texttt{RFF}}_1\) and \(\sigma^{\texttt{RFF}}_2\) are node/edge encoder and message encoder RFF parameters, respectively. We note better approximations with larger sigmas (lower standard deviation), similar to what we have observed with ELM. This is an important point, highlighting the value of data-agnostic methods in certain cases, especially when their additional tunable parameters are set appropriately. In our main paper experiments, we mainly chose the best-performing random feature method that requires no extra tuning, enabling fast training while maintaining accuracy comparable to gradient-descent-based approaches. However, in real-world scenarios, this fast training could be leveraged to further tune the hyperparameters of RFF or ELM and select the best-performing configuration.
\begin{table}[h]
    \centering
    \caption{RFF results are displayed where \(\sigma^{\text{RFF}}_{1}\) is the node and edge encoder RFF parameter, and \(\sigma^{\text{RFF}}_2\) is the message encoder RFF parameter. Relative \(l^2\) errors are displayed.}
    \label{tab:fourier-results}
    \scriptsize
    \begin{tabular}{llll}
        \toprule
        & \(\sigma^{\texttt{RFF}}_2 = 0.1\) & \(\sigma^{\texttt{RFF}}_{2} = 1\) & \(\sigma^{\texttt{RFF}}_{2} = 10\) \\
        \midrule
        \(\sigma^{\texttt{RFF}}_1 = 0.1\) & 1.02 & 1.01 & 1.01 \\
        \(\sigma^{\texttt{RFF}}_1 = 1\) & 1.03 & 9.4e-1 & 6.52e-1 \\
        \(\sigma^{\texttt{RFF}}_1 = 10\) & 5.34e-1 & 7.04e-2 & 4.76e-2 \\
        \bottomrule
    \end{tabular}
\end{table}

\end{document}